\documentclass[10pt,journal,compsoc]{IEEEtran}
\usepackage{url}
\usepackage{footnote}
\usepackage{amsthm,amsmath,amssymb,amsfonts}
\usepackage{color}
\usepackage{graphicx}
\usepackage{caption}
\usepackage{booktabs}
\usepackage{algorithm}\usepackage{algpseudocode}
\usepackage{float}

\ifCLASSOPTIONcompsoc
  \usepackage[nocompress]{cite}
\else
  \usepackage{cite}
\fi
\ifCLASSINFOpdf
\else
\fi
\begin{document}
%
\title{Differentiable Self-Adaptive Learning Rate}
%
%
%
%

\author{Bozhou~Chen,
        Hongzhi~Wang*~\IEEEmembership{Member,~IEEE,}
        Chenmin~Ba
\IEEEcompsocitemizethanks{\IEEEcompsocthanksitem All authors are with the Department of Computer Science and Technology, Harbin Institute of Technology, Harbin, China.
\protect\\
E-mails: \{bozhouchen, wangzh\}@hit.edu.cn, 21S003042@stu.hit.edu.cn, sxh@hit.edu.cn}
\thanks{Manuscript received None; revised None.}
}

%
%

\markboth{IEEE TRANSACTIONS ON PATTERN ANALYSIS AND MACHINE INTELLIGENCE}%
{}
%



\IEEEtitleabstractindextext{%
\begin{abstract}
Learning rate adaptation is a popular topic in machine learning.
Gradient Descent trains neural nerwork with a fixed learning rate. Learning rate adaptation is proposed to accelerate the training process through adjusting the step size in the training session. Famous works include Momentum, Adam and Hypergradient.
Hypergradient is the most special one. Hypergradient achieved adaptation by calculating the derivative of learning rate with respect to cost function and utilizing gradient descent for learning rate.
However, Hypergradient is still not perfect. In practice, Hypergradient fail to decrease training loss after learning rate adaptation with a large probability. Apart from that, evidence has been found that Hypergradient are not suitable for dealing with large datesets in the form of minibatch training. Most unfortunately, Hypergradient always fails to get a good accuracy on the validation dataset although it could reduce training loss to a very tiny value.
To solve Hypergradient's problems, we propose a novel adaptation algorithm, where learning rate is parameter specific and internal structured. We conduct extensive experiments on multiple network models and datasets compared with various benchmark optimizers. It is shown that our algorithm can achieve faster and higher qualified convergence than those state-of-art optimizers.
\end{abstract}

}

\maketitle


%

\IEEEraisesectionheading{\section{Introduction}\label{sec: intro}}

\IEEEPARstart{L}{earning} rate is the core of an optimizer and a key hyper-parameter in artificial neural networks. Related researches can be classified as three categories, including basic gradient descent, adaptation algorithms and hypergradient based algorithms.

The early GD(\underline{g}radient \underline{d}escent) algorithm~\cite{gradientdescent} fixes the learning rate in the whole training session and utilizes the first-order derivative of parameters. Three implementations are SGD(\underline{s}tochastic \underline{g}radient \underline{d}escent)~\cite{bottou2012stochastic}, minibatch gradient descent and batch gradient descent.
GD is able to fit a neural network on a dataset as long as with sufficient time. However, trade-off between convergence speed and model performance always exists and they cannot be achieved in the same time. Cause a large learning rate could accelerate training but could not reach the best performance~\cite{intability}. While a small learning rate is just the opposite. Apart from that, minibatch training and batch training are essentially non-convex optimization, where GD usually stucks into a local minima.

Adaptation methods are taken to avoid GD's fakes.
Momentum~\cite{momentum} takes the exponential moving average of historical gradient as step size. History gradient can help run out of the local minima. However, the trade-off between convergence speed and performance still exists.
AdaGrad~\cite{adagrad} subtly scales the step size by multiplying learning rate with the reciprocal of two-norm of historical gradients to have step size reduce along with training session. While AdaGrad is interfered a lot by historical gradient. Because step size keeps decreasing along with training session resulting that it converges very sluggishly in the later stage. Actually, step size in AdaGrad decreases purposelessly.
RMSProp~\cite{rmsprop} replaces two-norm in AdaGrad with exponential moving average of historical gradients to reduce the interference, which can be regarded as an extension of GD and AdaGrad.
Adam~\cite{adam} combines the best properties of Momentum and RMSProp, so that it can solve both the two fakes of GD surprisingly. However, Adam is not sensitive enough, especially when step size are needed to increase. Additionally, evidence shows that Adam always performs unsatisfactory in the late stage of a training session. When fitting a large datasets with minibatch training, a popular proposal is Momentum and change learning rate with a scheduler manually but not Adam to get stronger model.

HD(\underline{h}ypergradient \underline{d}escent) implements adaptation by calculating the derivative of learning rate with respect to cost function and utilizing gradient descent for learning rate. So HD knows exactly the need of cost funtion for learning rate, i.e., when to increase and when to decrease. However, case where HD misunderstands usually exists, i.e., training loss increases unexpectedly after HD's adaptation. Additionally, evidence can be found to show that HD are not suitable for minibatch training tasks. In other word, it's a hard problem for HD to process large datasets. Although HD gained a good training loss on published datasets, the performance on the validation dataset are nearly of no practical value.

In this paper, we propose a novel optimizer DSA(\underline{d}ifferentiable \underline{s}elf-\underline{a}daptive learning rate), which is an extension and improvement of HD and SGD. The core of DSA is the implementation of learning rate. Actually, the derivative of learning rate in HD is computed by an inefficient approximation. In DSA, the approximation to gradient is replaced with an effective detection to decrease HD's misunderstanding. Besides, many other tricks are adopted to enhance the proposed algorithm. For example, learning rate has been internal structured and is specific for each parameter.

Contributions of DSA are summarized as follows.
\begin{list}{\labelitemi}{\leftmargin=1em}\itemsep 0pt \parskip 0pt
    \item Proposed DSA could miss up HD's inherent shortcoming and we give out a complete program to deal with large datasets.
    \item DSA can be applied to different occasions, optimizing a wide range of network models and solving various datasets.
    \item We conduct extensive experiments on multiple neural network models and published datasets compared with various state-of-art optimizers. Experimental results have demonstrated the absolute advantage of DSA in speed, performance and sensitivity.
\end{list} 

In the remaining of this paper,  Section~\ref{sec: method} describes the proposed method. Experiments are conducted in Section~\ref{sec: exp}. We draw the conclusions in Section~\ref{sec: conclusion} and overview related work in Section~\ref{sec: related}.
\section{Method \label{sec: method}}
In this section, we will illustrate DSA(\underline{d}ifferentiable \underline{s}elf-\underline{a}daptive) algorithm in detail. Firstly, the basic algorithm HD(\underline{H}ypergradient \underline{D}escent) will be introduced as a foreshadowing for DSA in Section~\ref{sec: hd} and analysed about the practical pitfalls in Section~\ref{sec: short}. Finally, we propose our algorithm DSA in Section~\ref{sec: dsa}.
\subsection{Hypergradient Descent \label{sec: hd}}
This section is dedicated to reviewing Hypergradient Descent~\cite{hypergradient}. Basically, the gradient of learning rate with respect to cost function is gained through the parameter updated from the last iteration. The updating method of Gradient Descent is as
\begin{equation}
\label{equ: gd}
W_t = W_{t-1} - \alpha * \nabla{f(W_{t-1})},
\end{equation}
where $\alpha$ is the learning rate and $\nabla{f(W_{t-1})}$ is the gradient of previous parameter $W_{t-1}$.
Obviously, $W_t$ can be reagarded as a function of $\alpha$. HD makes the assumption that the optimal value of $\alpha$ does not change much between two consecutive iterations so that it can use the update rule for the previous step to optimize $\alpha$ in the current one. In other words, HD uses ${\partial f(W_{t-1})}/{\partial\alpha}$ to approximate ${\partial f(W_t)}/{\partial\alpha}$. For this, we will compute ${\partial f(W_{t-1})}/{\partial\alpha}$. Noting that $W_{t-1} = W_{t-2} - \alpha * \nabla{f(W_{t-2})}$, and applying the chain rule, we can get
\begin{align}
\label{equ: hd}
\begin{split}
\frac{\partial f(W_{t-1})}{\partial\alpha}
&= \nabla{f(W_{t-1})} \cdot \frac{\partial(W_{t-2} - \alpha\nabla{f(W_{t-2})})}{\partial\alpha}\\
&= \nabla{f(W_{t-1})} \cdot (-\nabla{f(W_{t-2})}),  
\end{split}
\end{align}
which allows us to compute the needed hypergradient with a simple dot product and the memory cost of only one extra copy of the original gradient. With this hypergradient, we construct a update rule for the learning rate as
\begin{align}
\label{equ: hd_lr}
\begin{split}
\alpha_t
&= \alpha_{t-1} - \beta\frac{\partial f(W_{t-1})}{\partial \alpha}\\
&= \alpha_{t-1} + \beta\nabla{f(W_{t-1})} \cdot \nabla{f(W_{t-2})}
\end{split}
\end{align}
introducing $\beta$ as $\alpha$'s step size. Considering the sequence $\alpha_t$, Formula~\ref{equ: gd} can be modified to
\begin{equation}
\label{equ: hd_gd}
W_t = W_{t-1} - \alpha_t * \nabla{f(W_{t-1})} .
\end{equation}
\subsection{Analyses about Hypergradient Descent \label{sec: short}}
\begin{figure}[h]
    \begin{minipage}{0.5\linewidth}
        \centering
        \includegraphics[width=\linewidth]{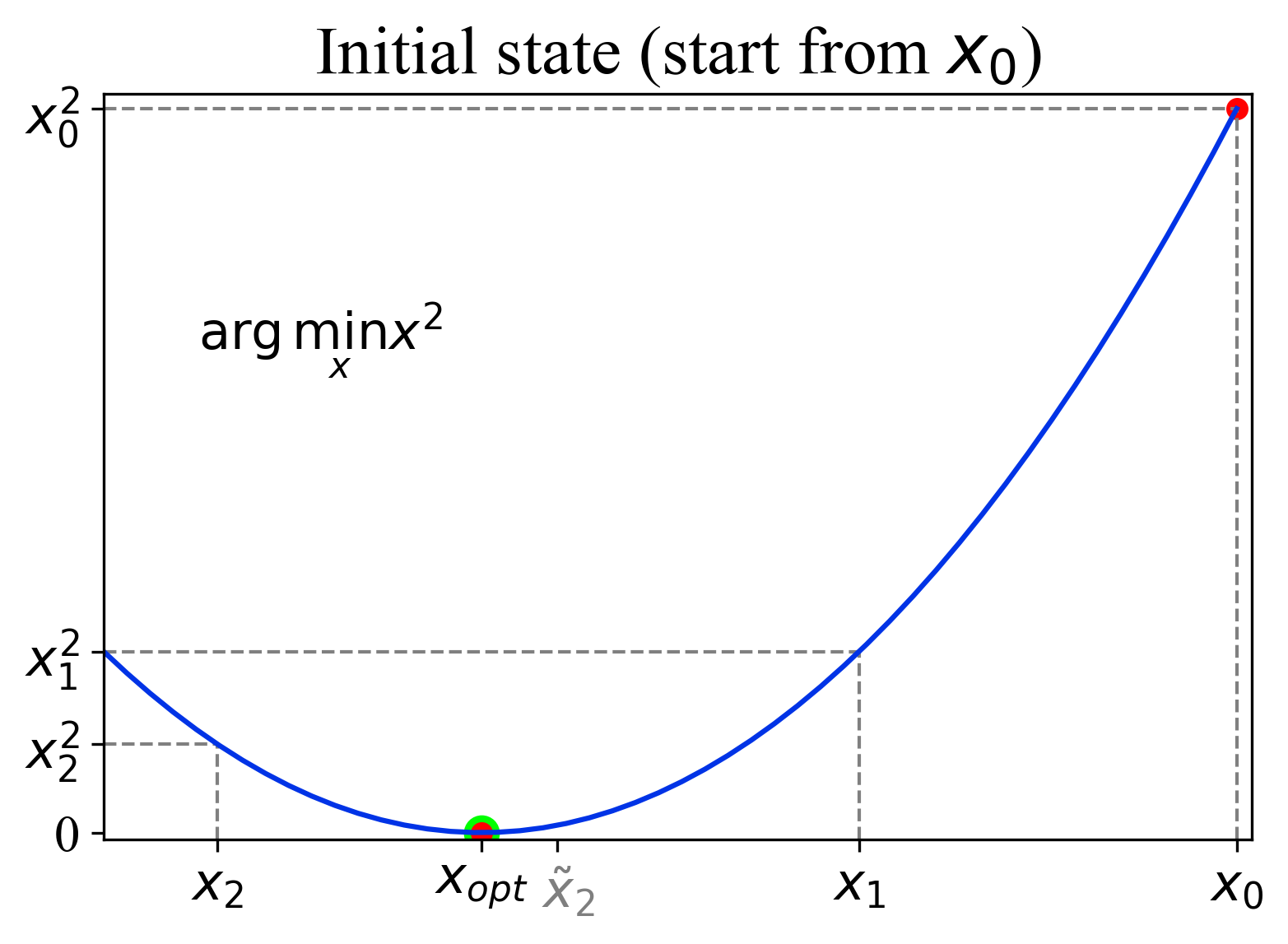}
    \end{minipage}
    \begin{minipage}{0.5\linewidth}
        \centering
        \includegraphics[width=\linewidth]{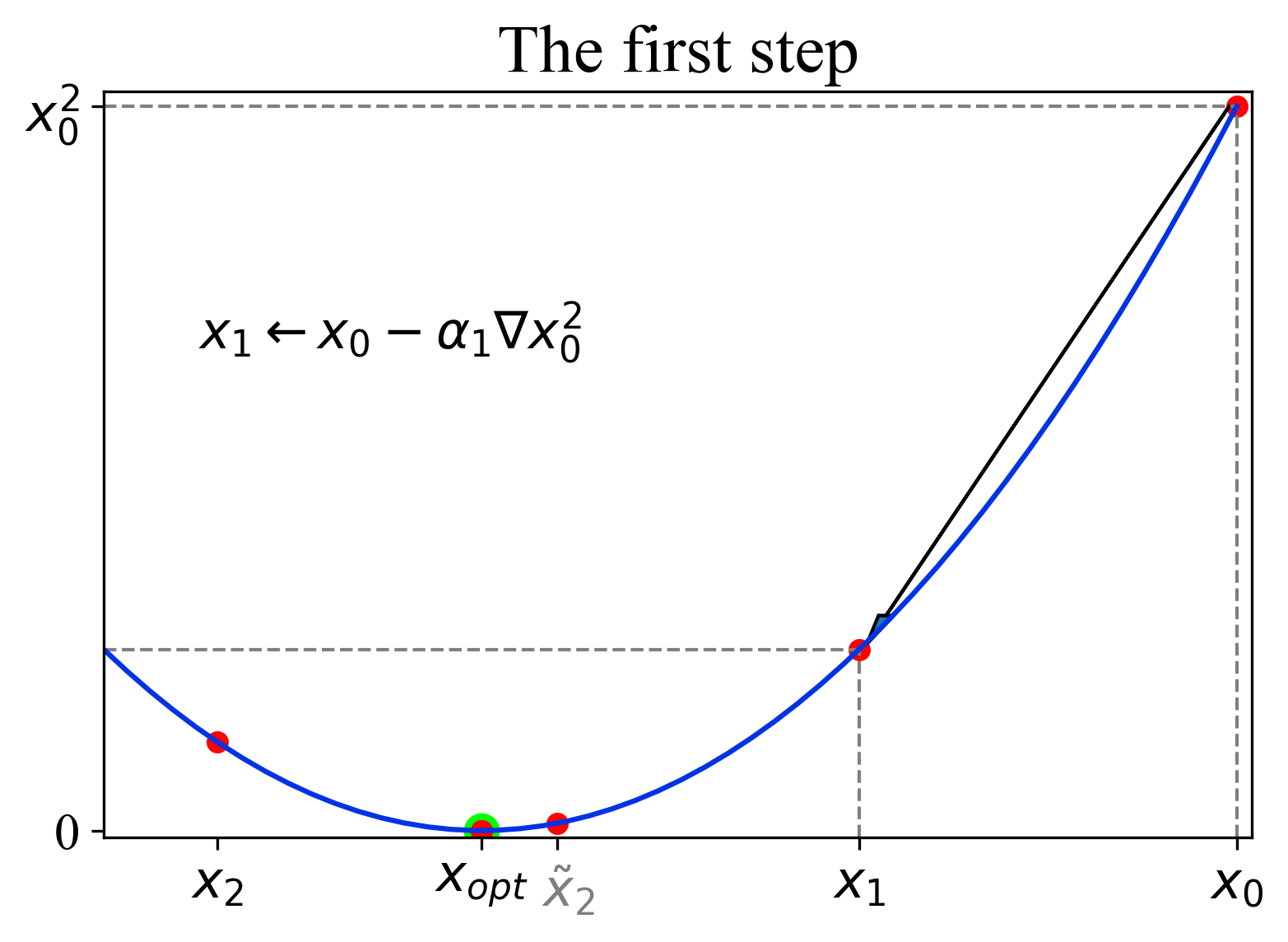}
    \end{minipage}
    \begin{minipage}{0.5\linewidth}
        \centering
        \includegraphics[width=\linewidth]{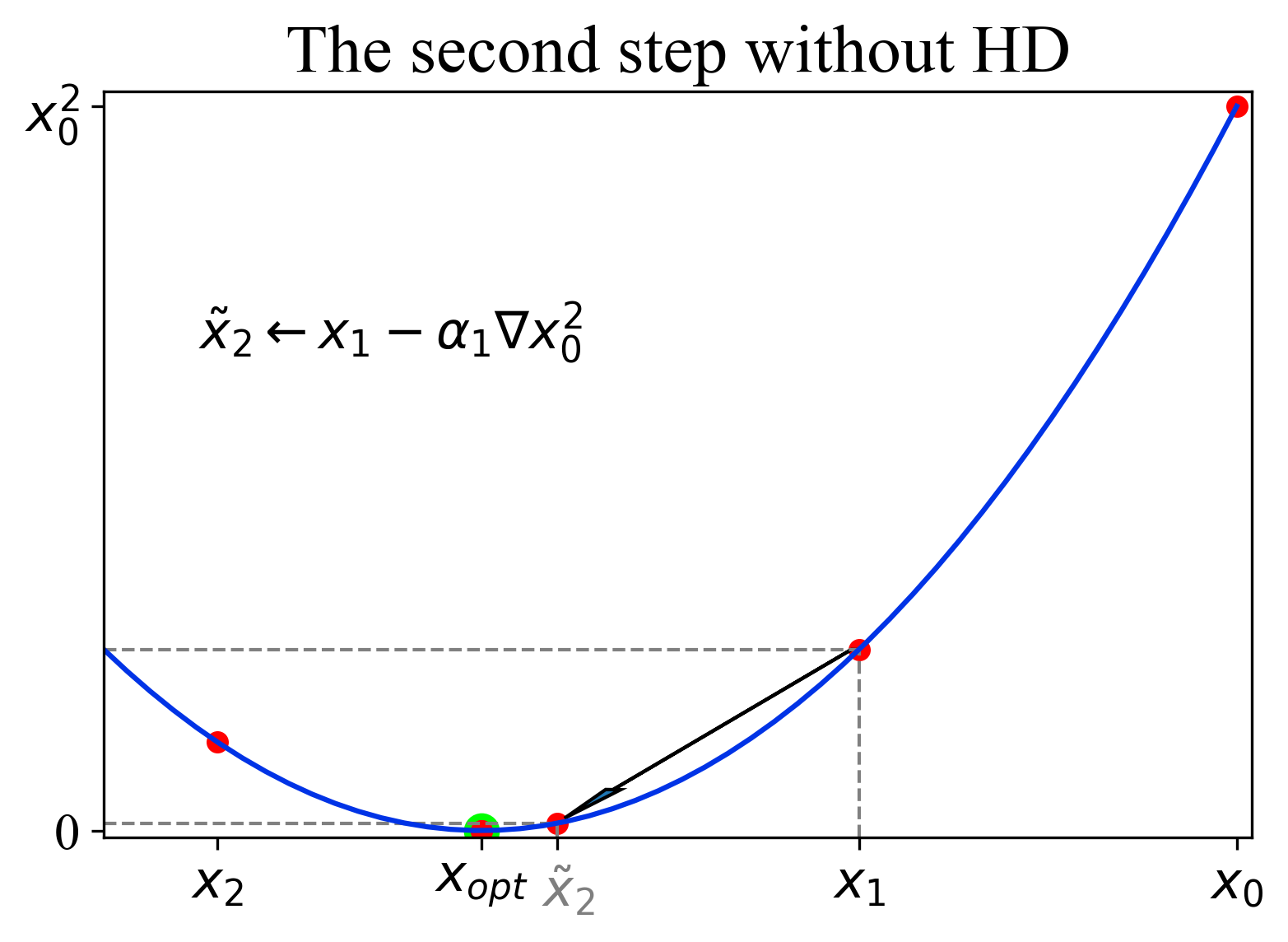}
    \end{minipage}
    \begin{minipage}{0.5\linewidth}
        \centering
        \includegraphics[width=\linewidth]{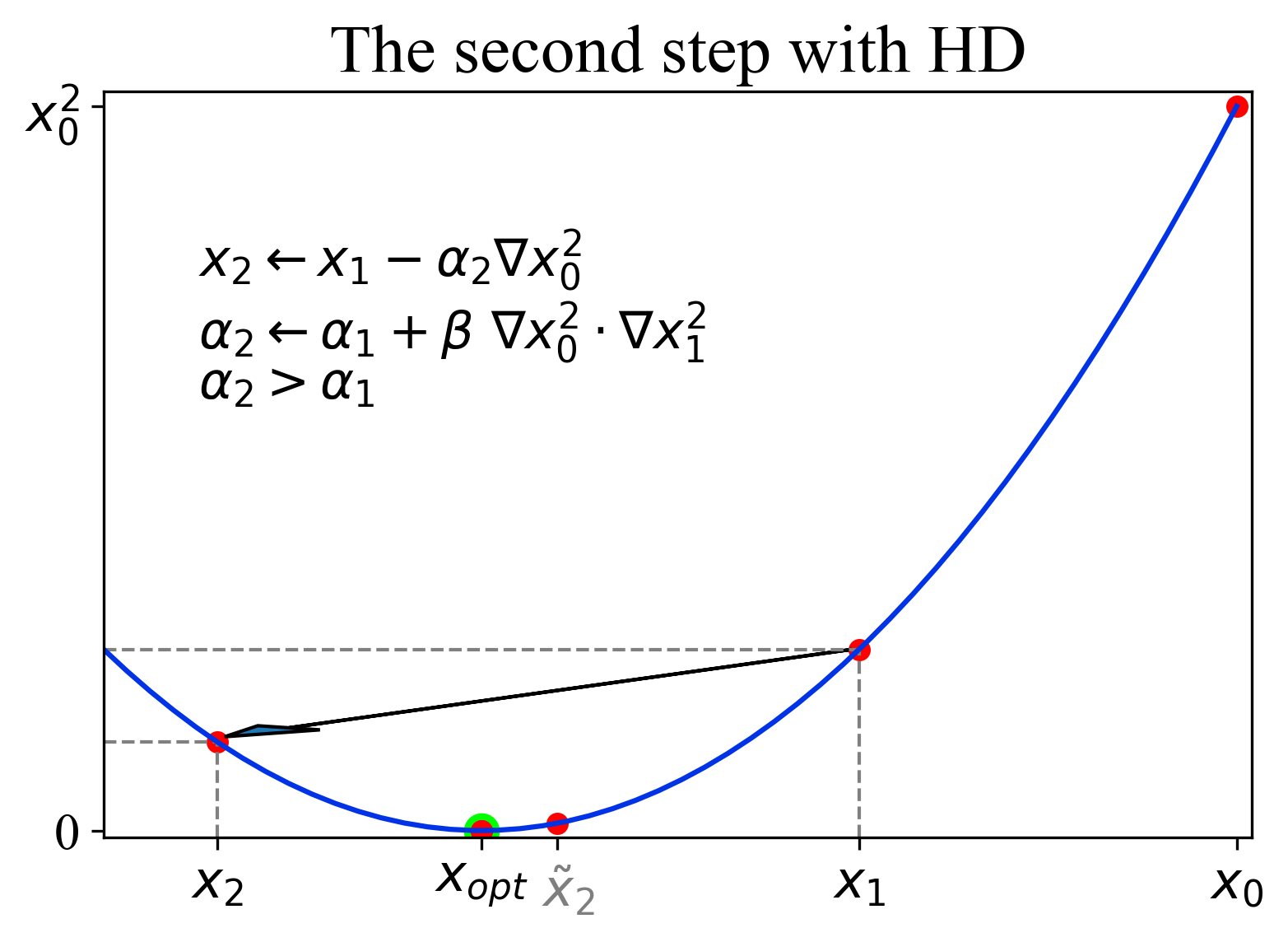}
    \end{minipage}
    \caption{Miss case using HD}
    \label{fig: hd_miss}
\end{figure}

In this section, we will take more in-depth analysis for Hypergradient Descent, including the sensitivity of HD, why HD may misunderstand the cost function and other problems mentioned in Section~\ref{sec: intro}.
\\\textbf{Sensitivity of HD}\\
\begin{figure}[h]
    \begin{minipage}{0.5\linewidth}
        \centering
        \includegraphics[width=\linewidth]{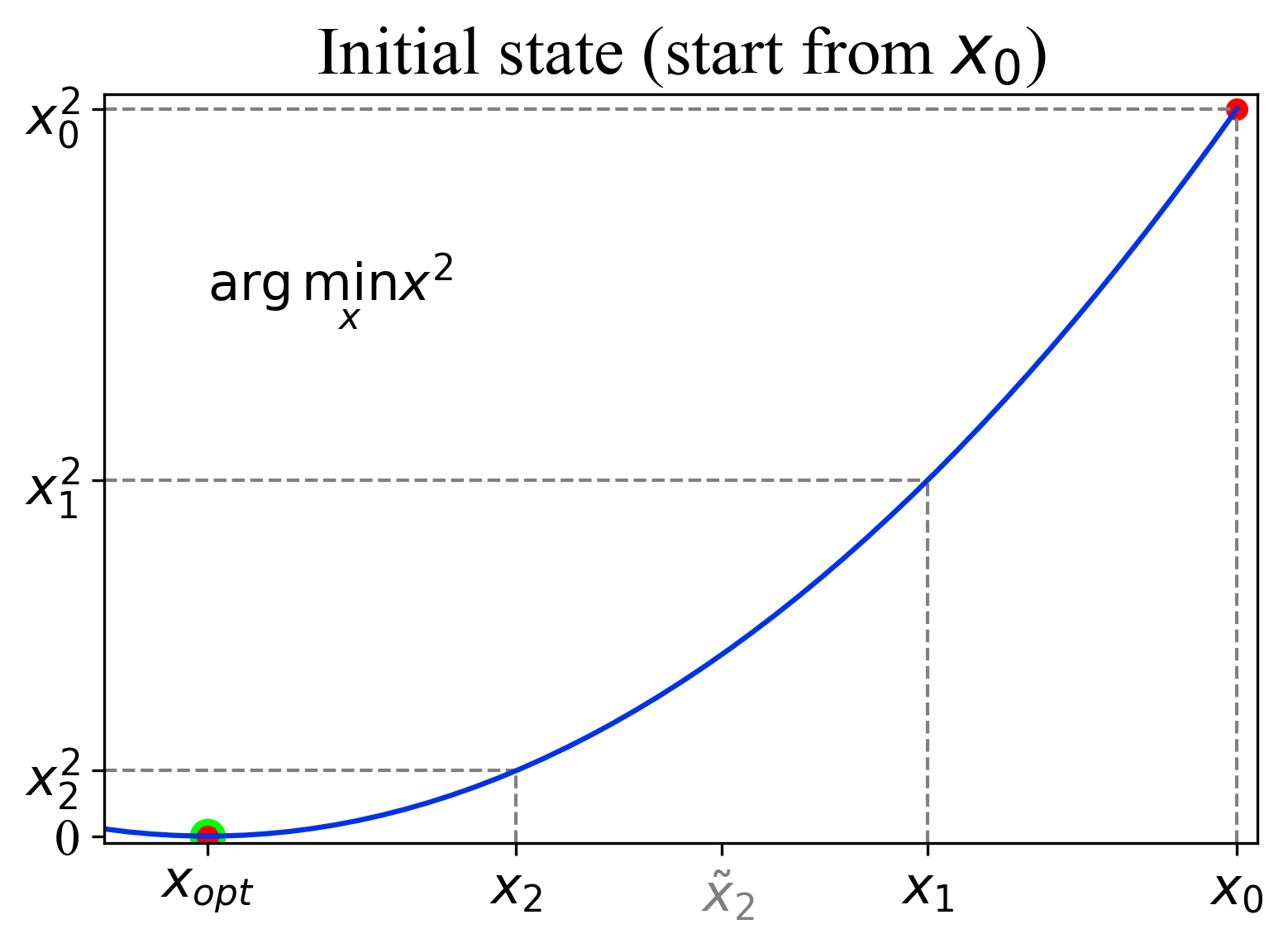}
    \end{minipage}
    \begin{minipage}{0.5\linewidth}
        \centering
        \includegraphics[width=\linewidth]{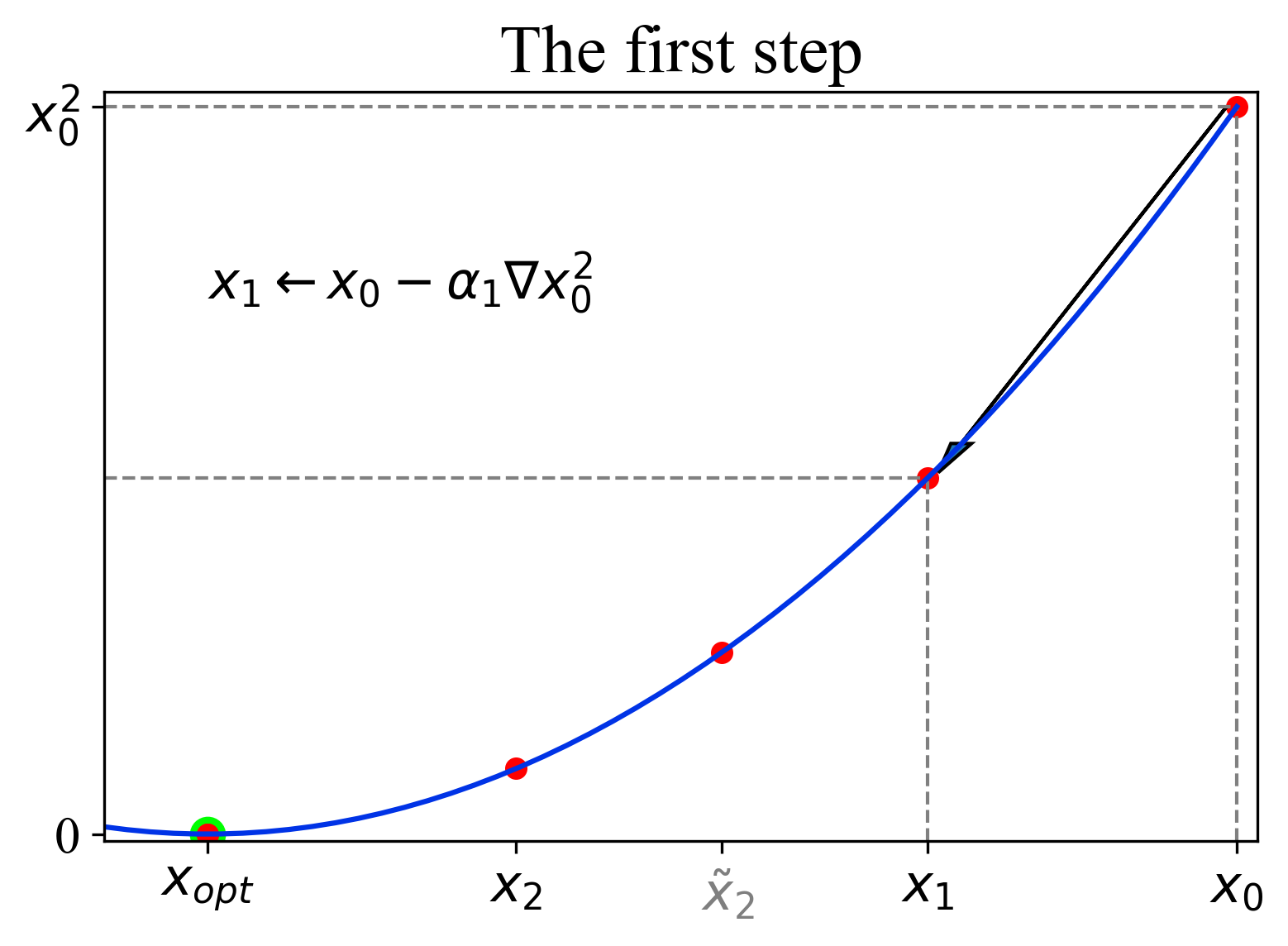}
    \end{minipage}
    \begin{minipage}{0.5\linewidth}
        \centering
        \includegraphics[width=\linewidth]{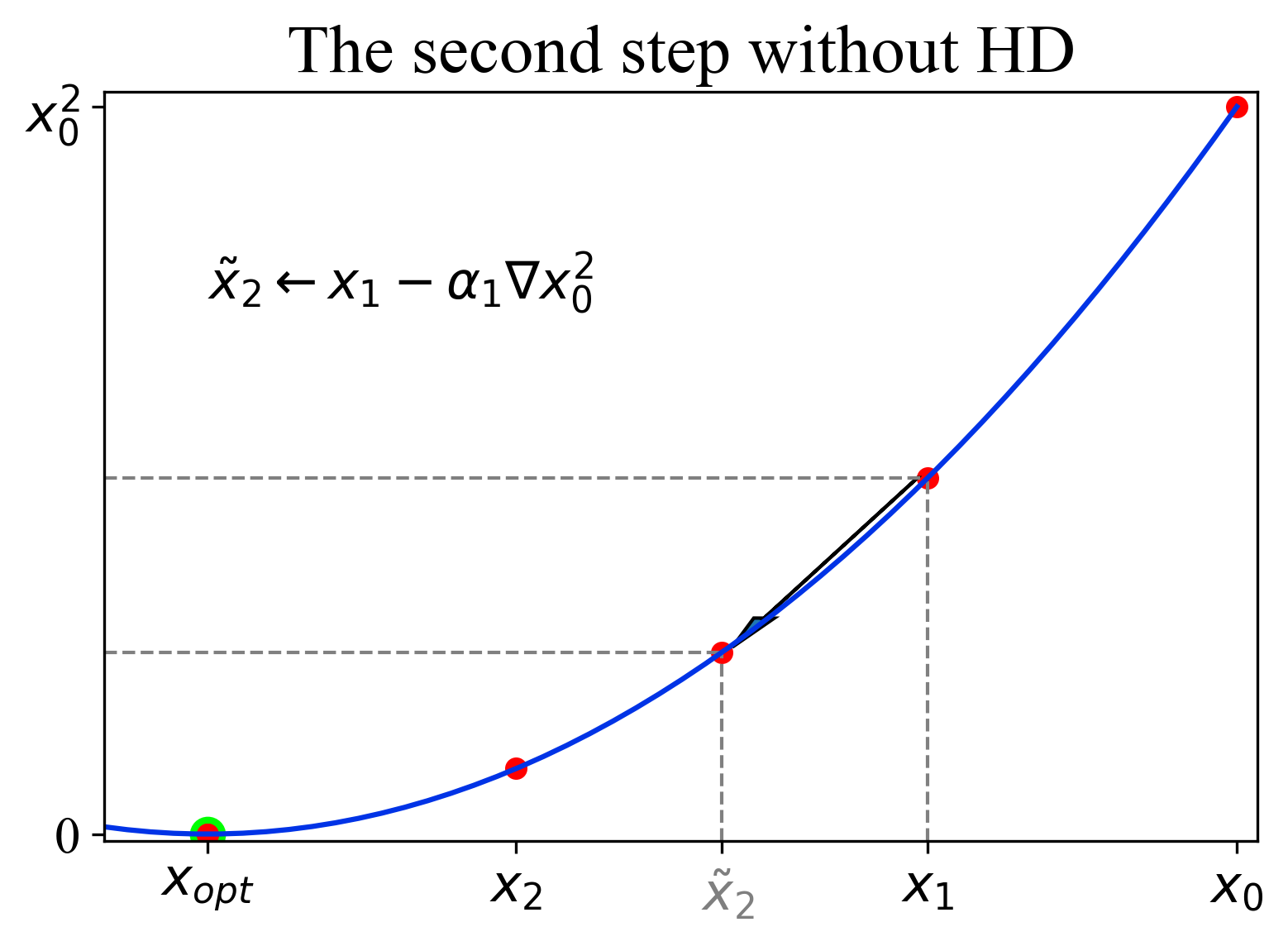}
    \end{minipage}
    \begin{minipage}{0.5\linewidth}
        \centering
        \includegraphics[width=\linewidth]{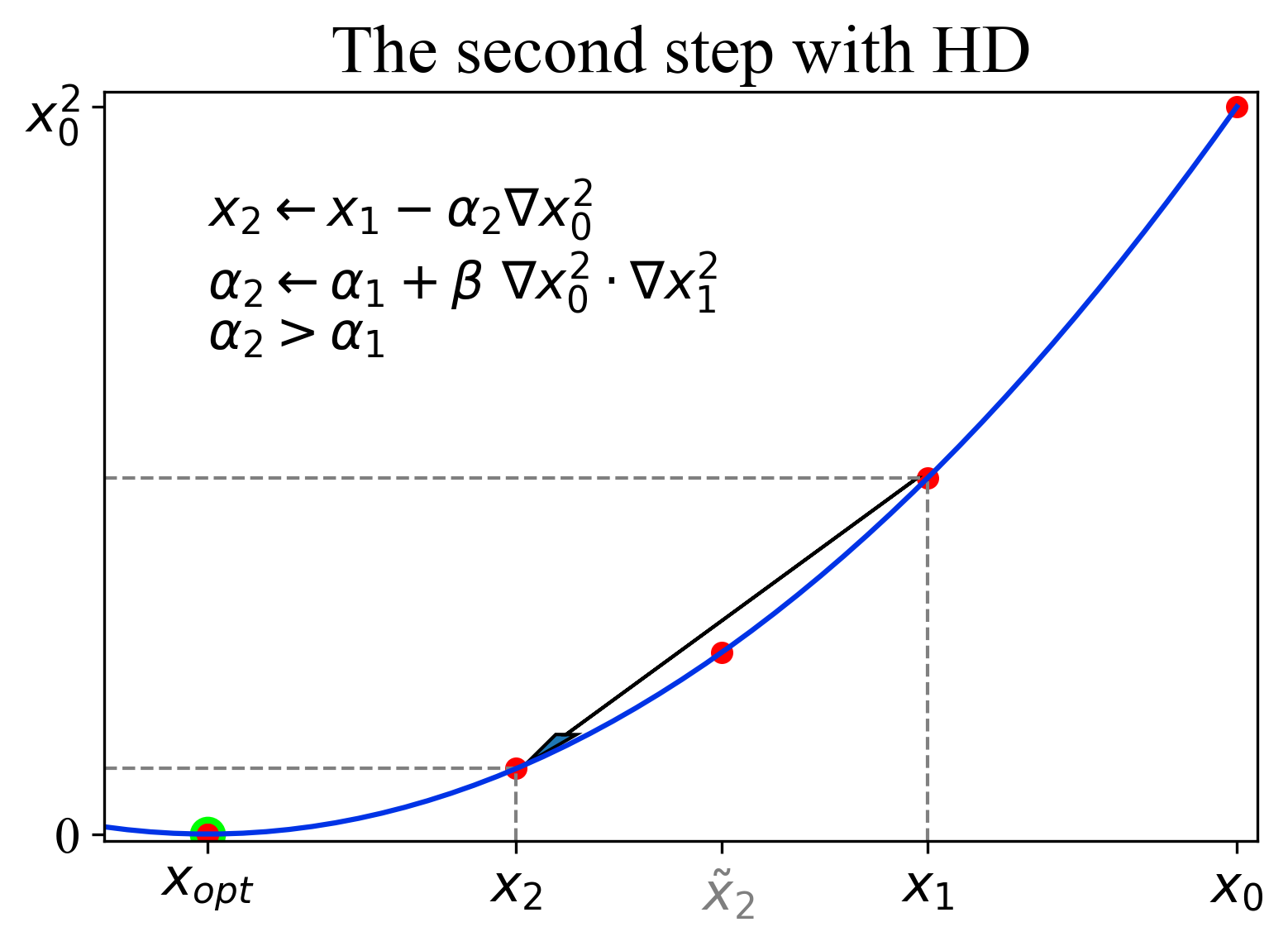}
    \end{minipage}
    \caption{Minimization of $x^2$ using HD}
    \label{fig: hd_sensitivity}
\end{figure}
HD is more sensitive than other traditional algorithms because learning rate in HD can increase according to occasions but not just decrease. Take a simple convex optimization $\arg\min_x{x^2}$ as example which is visualized in Fig.~\ref{fig: hd_sensitivity}. Optimization starts from $x_0$, and $x$ steps to $x_1$ with the initial learning rate $\alpha_1$. Suppose $x_1$ is still far from the extreme point, i.e., $|x_1-x_{opt}| \gg \alpha_1$. In the second iteration, the step size of learning rate is $\beta\nabla{x_1^2}\cdot\nabla{x_0^2} > 0$. Therefore, learning rate will increase and the step size of $x$ will be larger than that of occasions without HD.
Actually, we could find a more intuitive explanation for this behavior of HD. In each current iteration, if the step direction indicated by parameter's gradient is the same to that of last iteration, it can be understood as the current position is on a smooth descending surface, where speeding up is a better policy to approaching the extreme point.
\\\textbf{Shortcoming of HD}\\
In the last paragraph, we learned why HD can control learning rate sensitively. However, the explanation is also the reason why HD usually misunderstands the cost function. In the example of last paragraph, we have a key assumption that $x_1$ is still far from the extreme point, which ensures that increasing the learning rate will not cause the extreme point to be crossed after one step. While unfortunately, occasions like Fig.~\ref{fig: hd_miss} always exist. As shown in the figure, $x_1$ is already very close to the extreme opint after the first iteration. While according to HD's updating rule for learning rate, step size increases to a larger value resulting the miss of HD, i.e., $x_2^2 > \tilde{x}_2^2$. 
We have taken a small experiment to calculate the frequency of HD's misunderstanding. From the second iteration, we compare $L = f(W_{t-1} - \alpha_{t-1}\nabla{f(W_{t-1})})$ and $L_{adapt} = f(W_{t-1} - \alpha_t\nabla{f(W_{t-1})})$, where $f(W)$ is the cost function or the loss function, the smaller the better. $L$ is the loss if we don't apply HD to take adaptation for learning rate, while $L_{adapt}$ is the opposite. Obviously, $L_{adapt} \leq L$ is as our expectation. If $L_{adapt} > L$, we think one misunderstanding happens and count as a miss. When the training session is over, we can calculate the misunderstanding frequency as $M/(T-1)$, where $M$ is the count of miss and $T$ is the total iterations. We apply HD to train a simple MLP on four different fearture datasets and record the average miss rate in a complete training session as shown in TABLE~\ref{tab: hd_miss_rate}. We can see that nearly one-fourth adaptation is a failure when dealing with a somewhat complex dataset.
\begin{table}[h]
    \caption{Miss rate of HD on four different feature datasets}
    \label{tab: hd_miss_rate}
    \centering
    \begin{tabular}{lrrrr}
    \toprule
    \textbf{DataSets}&IRIS&WINE&CAR&AGARICUS\\
    \textbf{MissRate}&0.057&\bf 0.228&0.061&0.029\\
    \hline
    \end{tabular}
\end{table}
\begin{figure}[h]
    \begin{minipage}{0.5\linewidth}
        \centering
        \includegraphics[width=\linewidth]{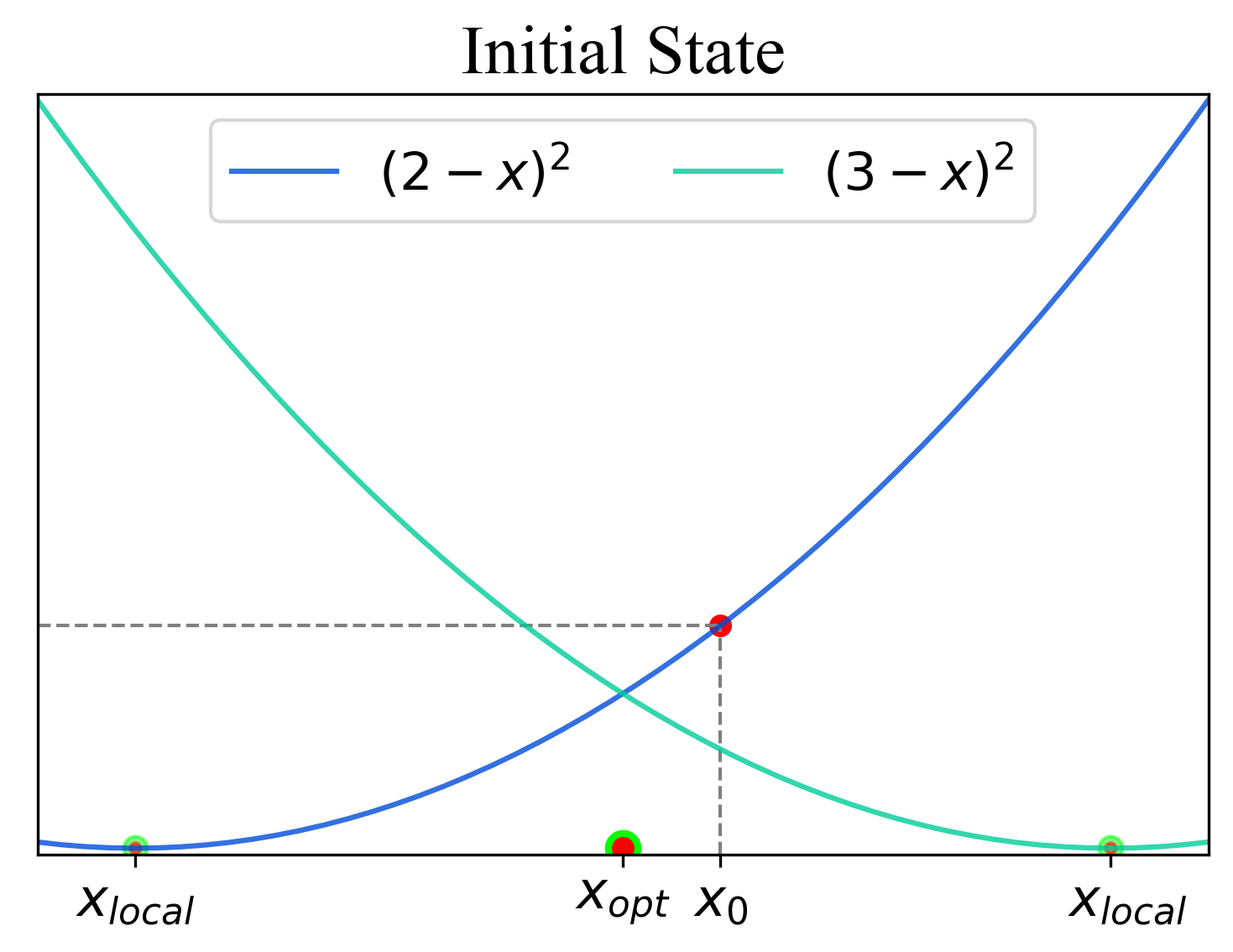}
    \end{minipage}
    \begin{minipage}{0.5\linewidth}
        \centering
        \includegraphics[width=\linewidth]{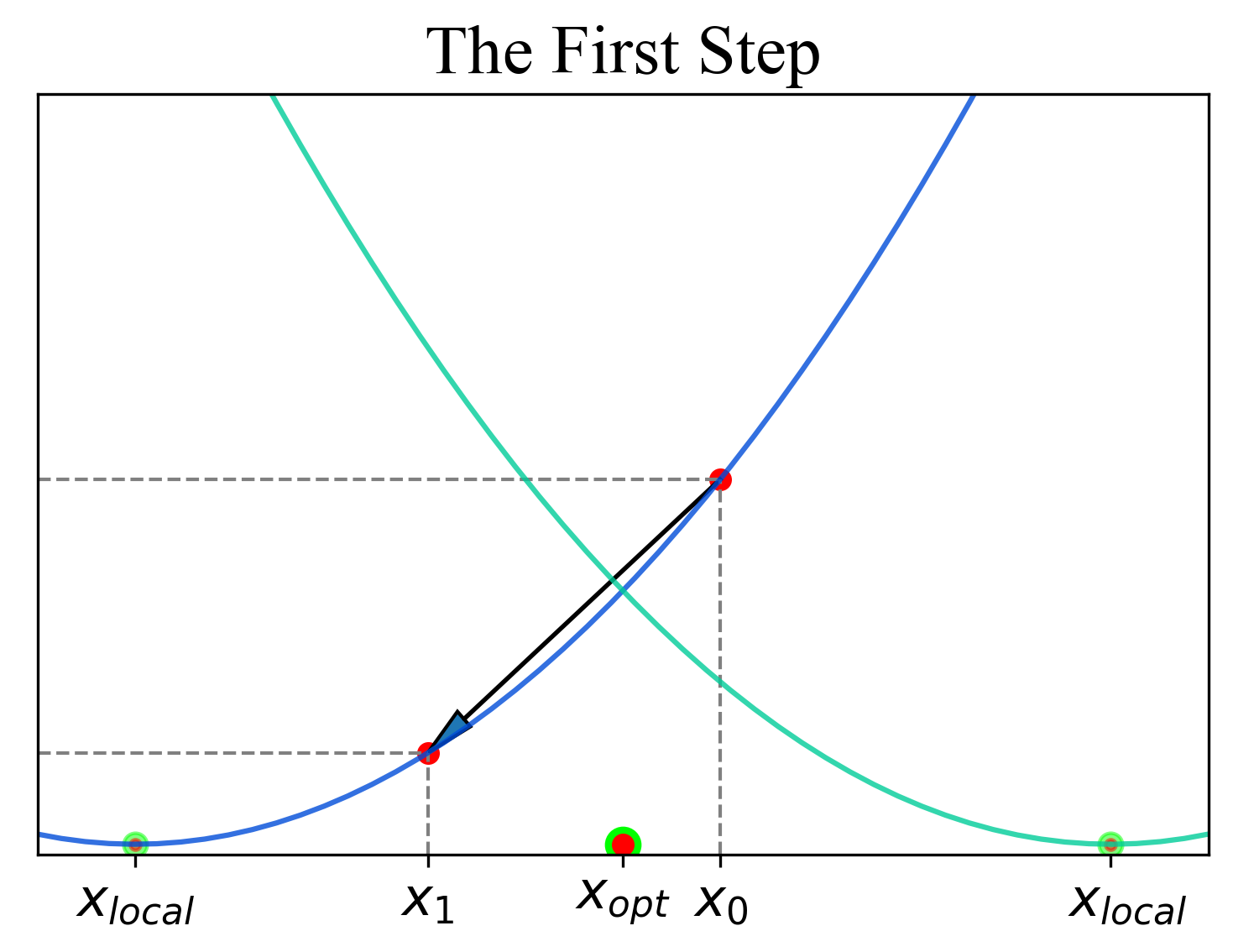}
    \end{minipage}
    \begin{minipage}{0.5\linewidth}
        \centering
        \includegraphics[width=\linewidth]{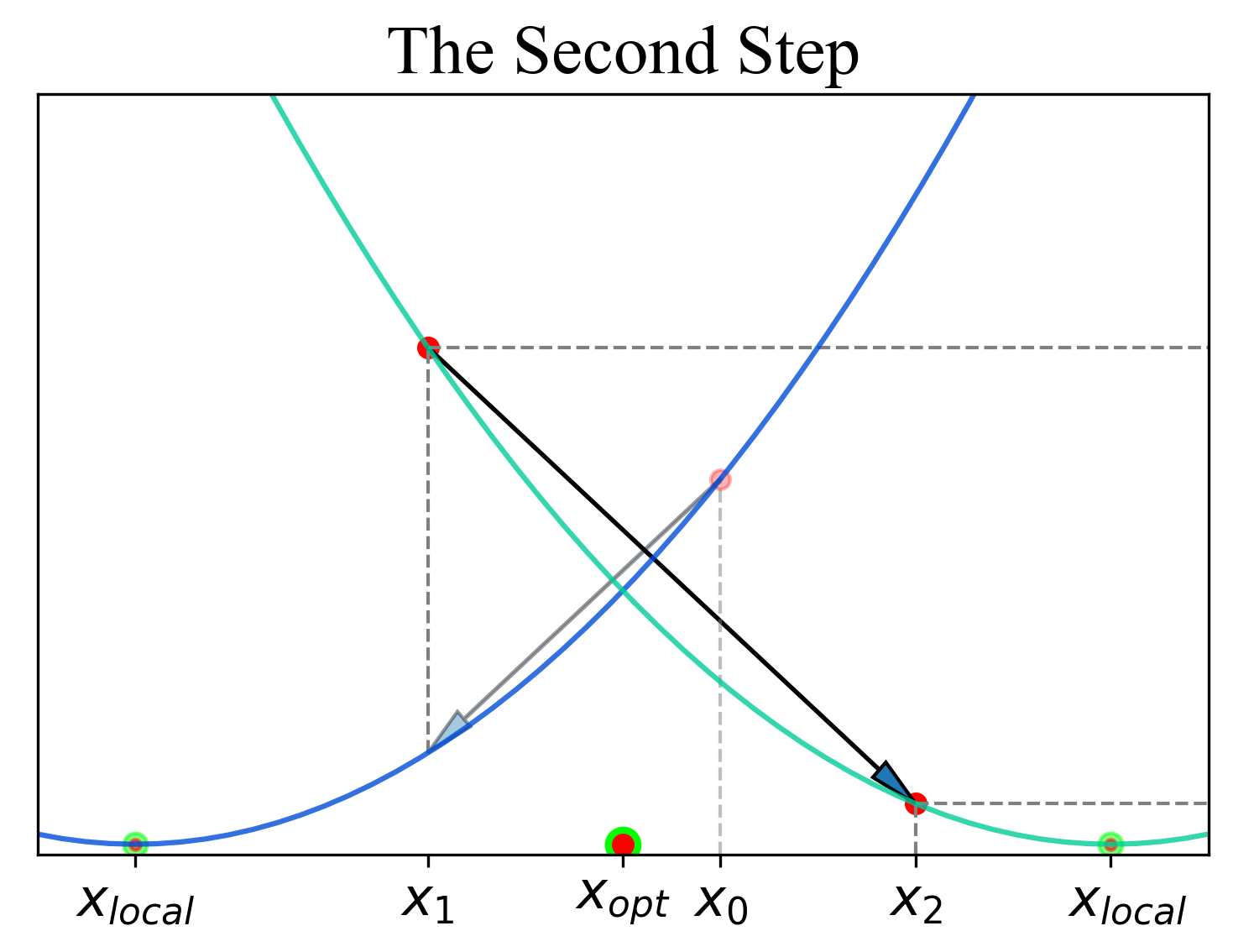}
    \end{minipage}
    \begin{minipage}{0.5\linewidth}
        \centering
        \includegraphics[width=\linewidth]{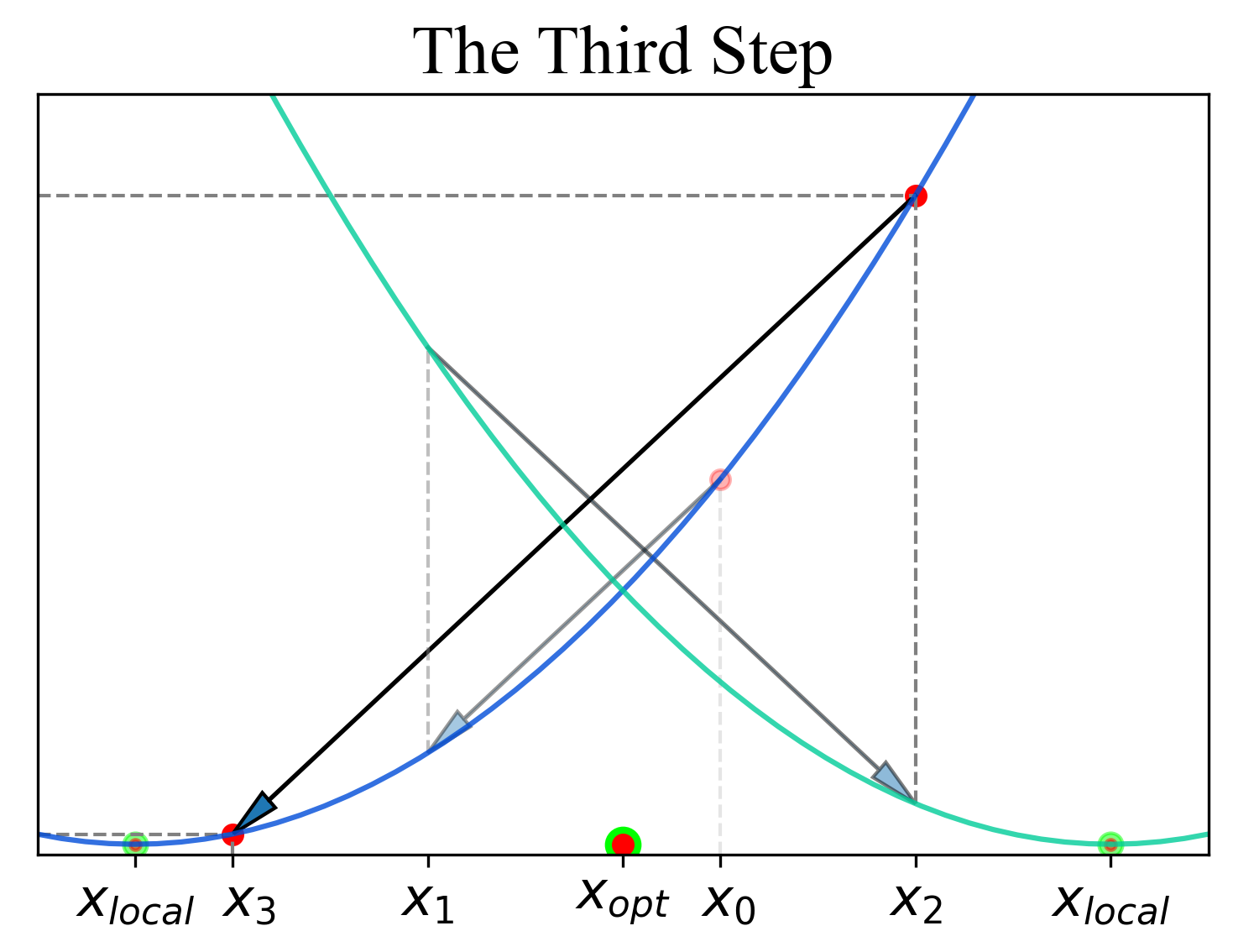}
    \end{minipage}
    \caption{Minibatch training using HD}
    \label{fig: hd_mini}
\end{figure}

When dealing with minibatch training tasks, the miss rate of HD will be larger. Because the assumption that using ${\partial f(W_{t-1})}/{\partial\alpha}$ to approximate ${\partial f(W_t)}/{\partial\alpha}$ requires to remain unchanged between two consecutive iterations. But cost function is different between two consecutive iterations, cause training samples are changing with iterations in minibatch training. In each current iteration, if the step direction of parameter on current distribution is the same to that on the last iteration's distribution, learning rate will increase. This is absolutely ridiculous.
We can use an effective detection to solve this problem, which will be discussed in Section~\ref{sec: dsa}.
Here, in order to make readers better understand why hypergradient based learning rate adaptation method is not suitable for minibatch training, we first make an assumption. Assuming that every time HD reaches a new surface, it can decide whether learning rate should increase or decrease at the current position immediately, regardless of the change of cost function. That is, if the current position is far from the extreme point, the learning rate increases, and vice versa.
Then let's take a simple regression task as evidence to illustrate the idea. The target is to fit two points $(1,2), (1,3)$ with line $y = w\cdot x$, i.e., $\min\sum{(y_i - w\cdot x_i) ^ 2}$. We fit the line in the form of minibatch, i.e., the first batch contains one point $(1,2)$ and the second contains another point. Iterations are visualized in Fig.~\ref{fig: hd_mini}, noting that we substitute parameter $w$ with $x$ in figures. Obviously, the optimal resolvement is $x_{opt}$, However, HD will never achieve that point. Because in each iteration, HD try to increase step size to make the current cost function as smaller as possible. As a result, $|x_0-x_{opt}|<|x_1-x_{opt}|<|x_2-x_{opt}|<|x_3-x_{opt}|$ and HD gets into a trap. Learning rate increases blindly, $x$ is more and more close to the local minima after each iteration but not the global optimal.

\begin{figure}[h]
    \begin{minipage}{0.5\linewidth}
        \centering
        \includegraphics[width=\linewidth]{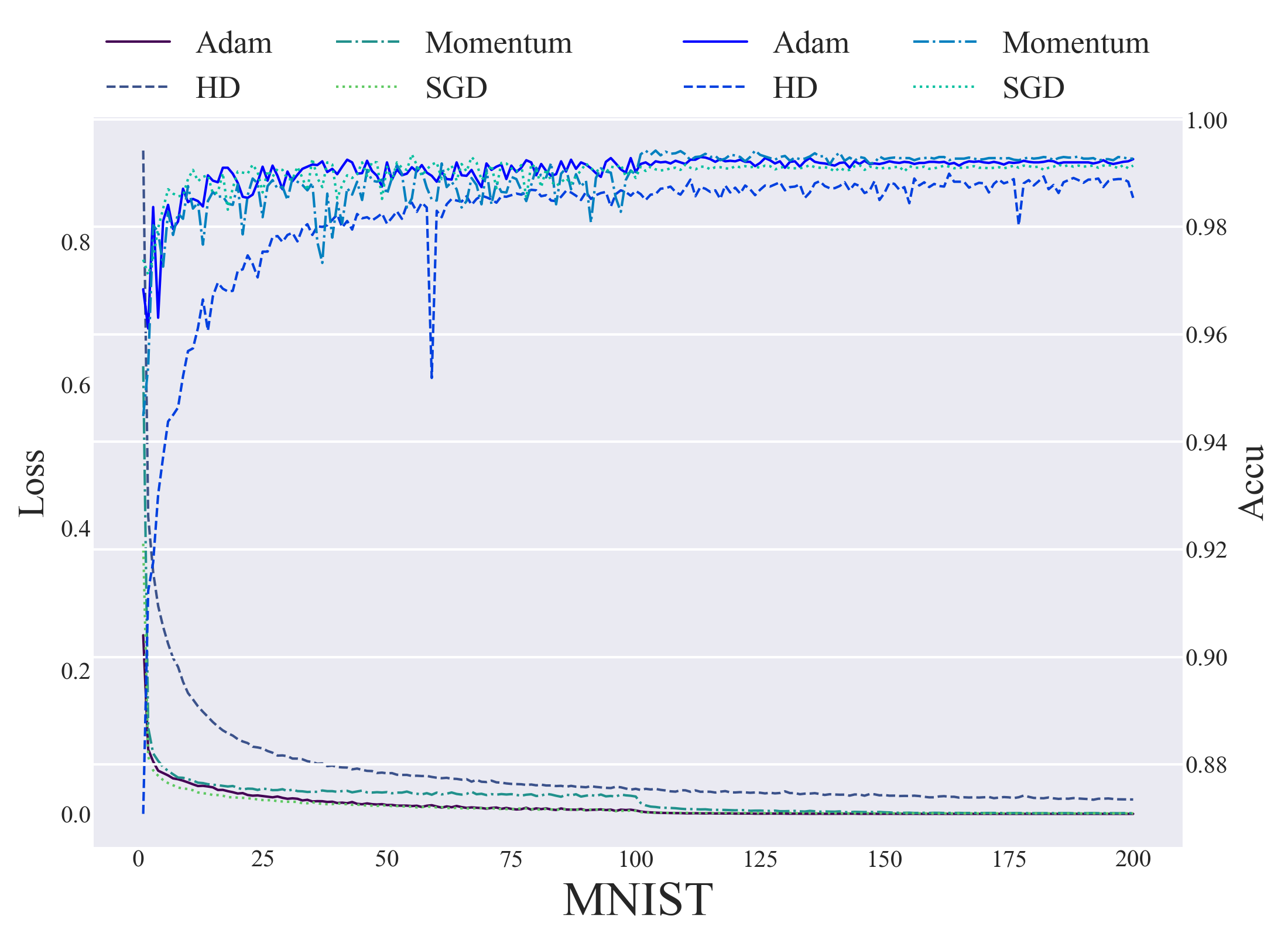}
    \end{minipage}
    \begin{minipage}{0.5\linewidth}
        \centering
        \includegraphics[width=\linewidth]{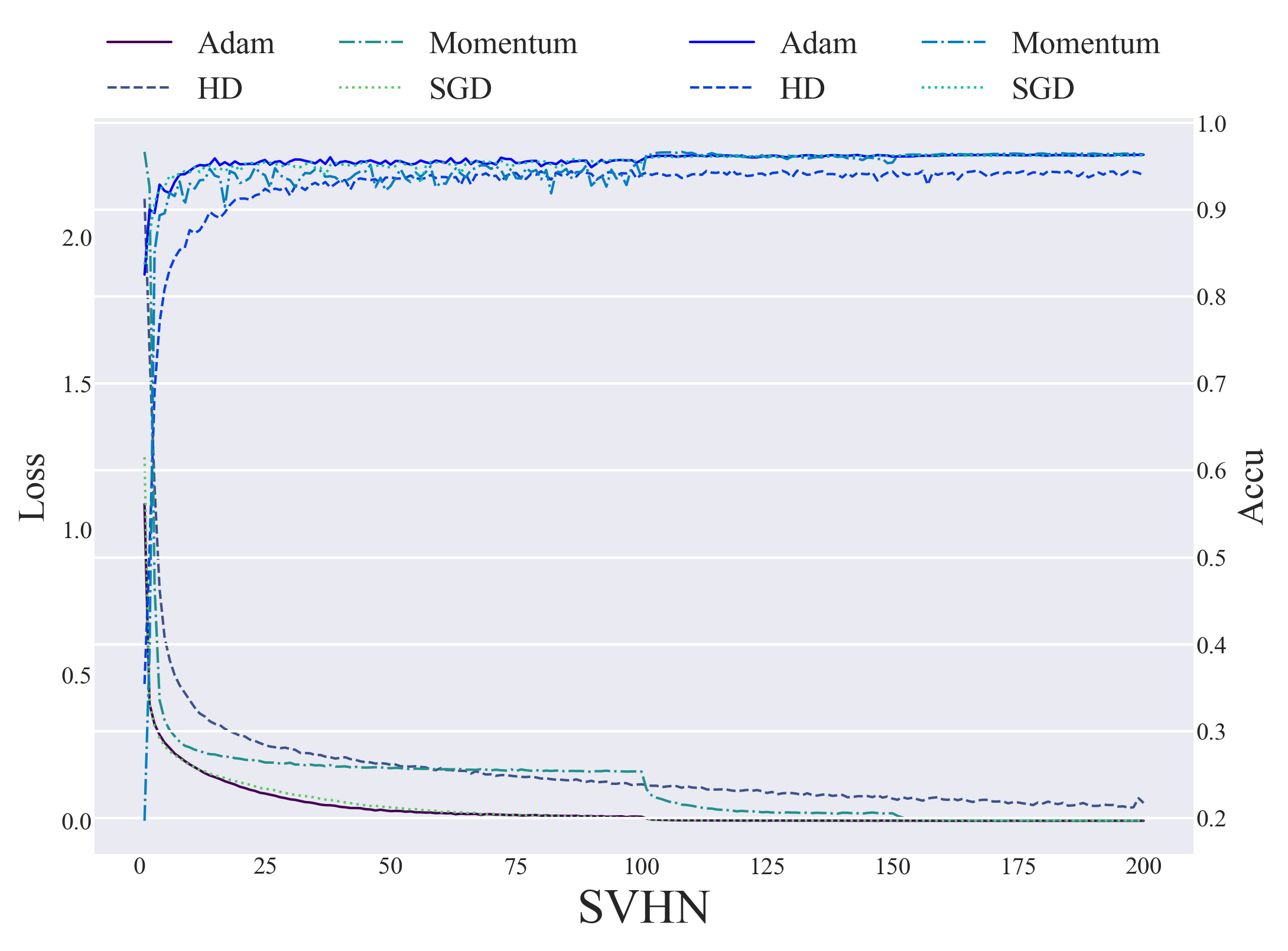}
    \end{minipage}
    \begin{minipage}{0.5\linewidth}
        \centering
        \includegraphics[width=\linewidth]{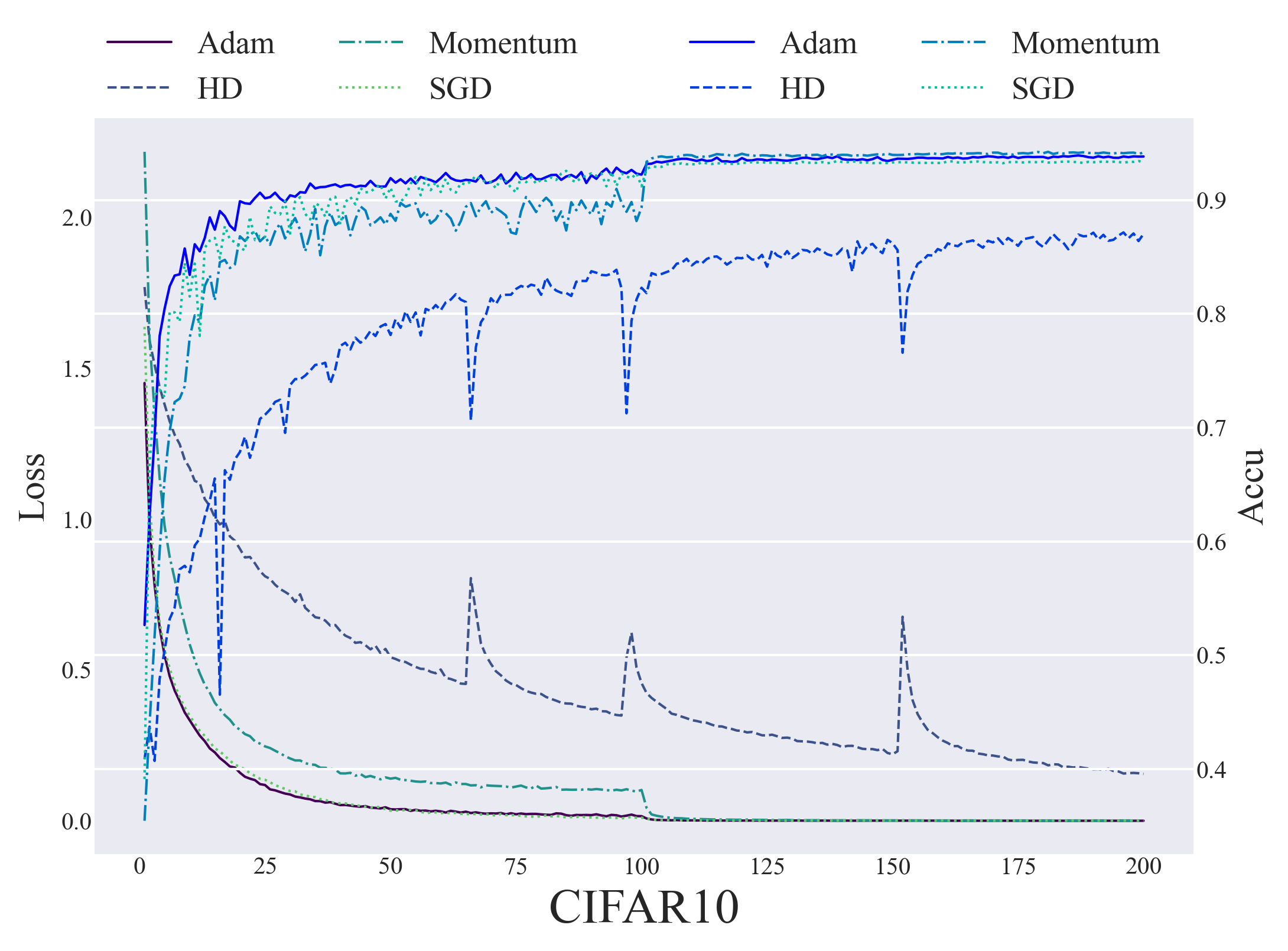}
    \end{minipage}
    \begin{minipage}{0.5\linewidth}
        \centering
        \includegraphics[width=\linewidth]{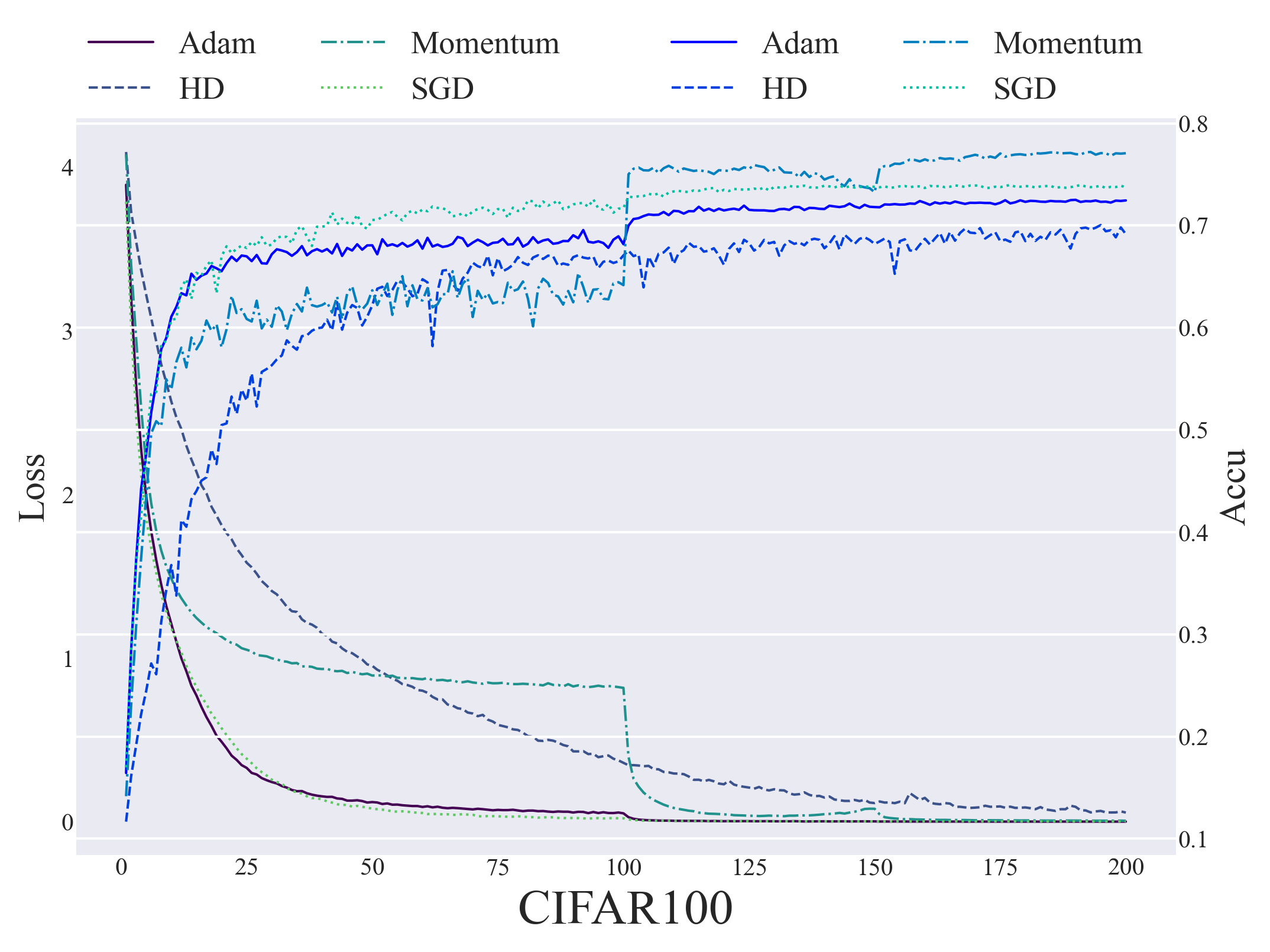}
    \end{minipage}
    \caption{Training loss and validation accuracy of HD with minibatch training}
    \label{fig: hd_mini_loss}
\end{figure}

Although HD could decrease the training loss to a quite low scale, the performance of the trained model is actually ordinary. TABLE~\ref{tab: hd_miss_rate_minibatch} tells us the miss rate when processing large datasets in minibatch training, where nearly half of learning rate's adaptation is a miss. Fig.~\ref{fig: hd_mini_loss} indicates that although HD may get a good train loss such as SVHN, the trained neural network still fail to get a good accuracy on validation dataset.
\begin{table}[h]
    \caption{Miss rate of HD with minibatch}
    \label{tab: hd_miss_rate_minibatch}
    \centering
    \begin{tabular}{lrrrr}
    \toprule
    \textbf{DataSets}&MNIST&SVHN&CIFAR10&CIFAR100\\
    \textbf{MissRate}&0.4909&0.4946&0.4970&0.4977\\
    \hline
    \end{tabular}
\end{table}
\subsection{Differentiable Self-Adaptive Learning Rate \label{sec: dsa}}
In this section, we will propose our algorithm DSA while trying to solve HD's pitfalls. Totally, Section~\ref{sec: detect} aims to solve HD's misunderstanding and Section~\ref{sec: dsa_mini} is against the problem in minibatch training. In Section~\ref{sec: trick}, techniques to further improve DSA will be illustrated in detail. Finally, pseudo code of DSA will be drawn in Section~\ref{sec: pseudo}.
\subsubsection{Detection rather than Experience\label{sec: detect}}
As talked in Section~\ref{sec: hd}, HD approximates ${\partial f(W_t)}/{\partial\alpha}$ with ${\partial f(W_{t-1})}/{\partial\alpha}$. This approximation takes use of the experience in the last iteration to decide the change for learning rate. While DSA takes detection for the future distribution rather than empirical adaptation to avoid HD's misunderstanding. In DSA, we introduce an internal variable $\widetilde{W}$
\begin{align}
\label{equ: dsa_wilde_w}
\widetilde{W}
&= W_{t-1} - \alpha * \nabla{f(W_{t-1})},
\end{align}
which is a function of $\alpha$. Nextly, we try to minimize $f(\widetilde{W}(\alpha))$ by optimizing $\alpha$ through
\begin{align}
\label{equ: dsa_partial_alpha}
\begin{split}
\frac{\partial f(\widetilde{W})}{\partial\alpha}
&= \nabla{f(\widetilde{W})} \cdot \frac{\partial(W_{t-1} - \alpha\nabla{f(W_{t-1})})}{\partial\alpha}\\
&= \nabla{f(\widetilde{W})} \cdot (-\nabla{f(W_{t-1})}).
\end{split}
\end{align}
Therefore, update rule for learning rate will be modified to
\begin{align}
\label{equ: dsa_alpha_rule}
\begin{split}
\alpha_t
&= \alpha_{t-1} - \beta\frac{\partial f(\widetilde{W})}{\partial \alpha}\\
&= \alpha_{t-1} + \beta\nabla{f(\widetilde{W})} \cdot \nabla{f(W_{t-1})}.
\end{split}
\end{align}
While the parameter's step stays the same
\begin{align}
\label{equ: dsa_parameter_rule}
\begin{split}
W_t = W_{t-1} - \alpha_t * \nabla{f(W_{t-1})}.
\end{split}
\end{align}
The cost is $f(\widetilde{W}(\alpha_{t}))$ after $t$-th iteration.
If we don't apply DSA adaptation for learning rate, the cost will be $f(\widetilde{W}(\alpha_{t-1}))$. And it's surely that $f(\widetilde{W}(\alpha_{t}))$ is smaller than $f(\widetilde{W}(\alpha_{t-1}))$ because DSA adaptation for learning rate is essentially a gradient descent for $f(\widetilde{W}(\alpha))$ with respect to $\alpha$.

\begin{figure}[h]
    \begin{minipage}{0.5\linewidth}
        \centering
        \includegraphics[width=\linewidth]{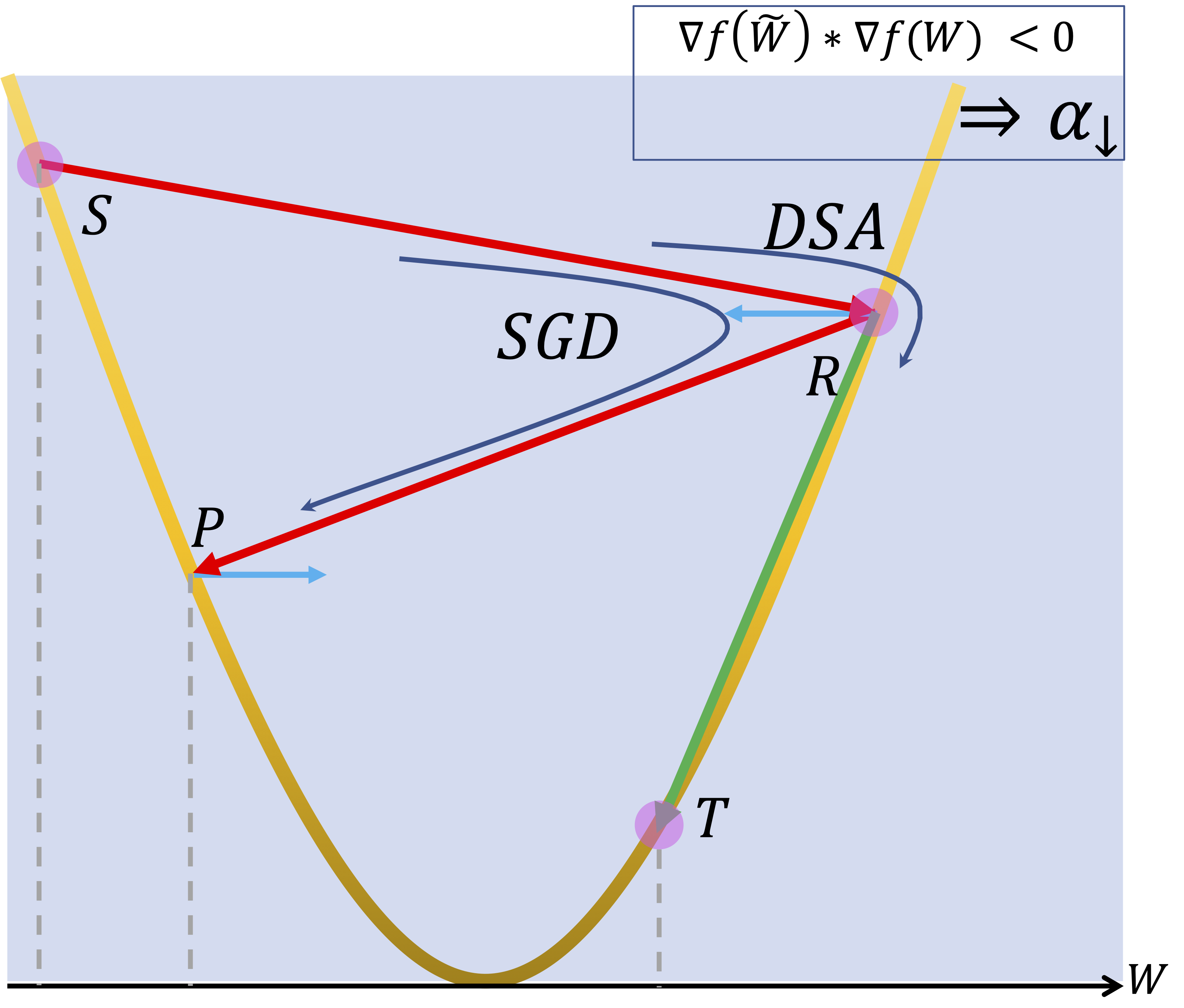}
        \centerline{(a)}
    \end{minipage}
    \begin{minipage}{0.475\linewidth}
        \centering
        \includegraphics[width=\linewidth]{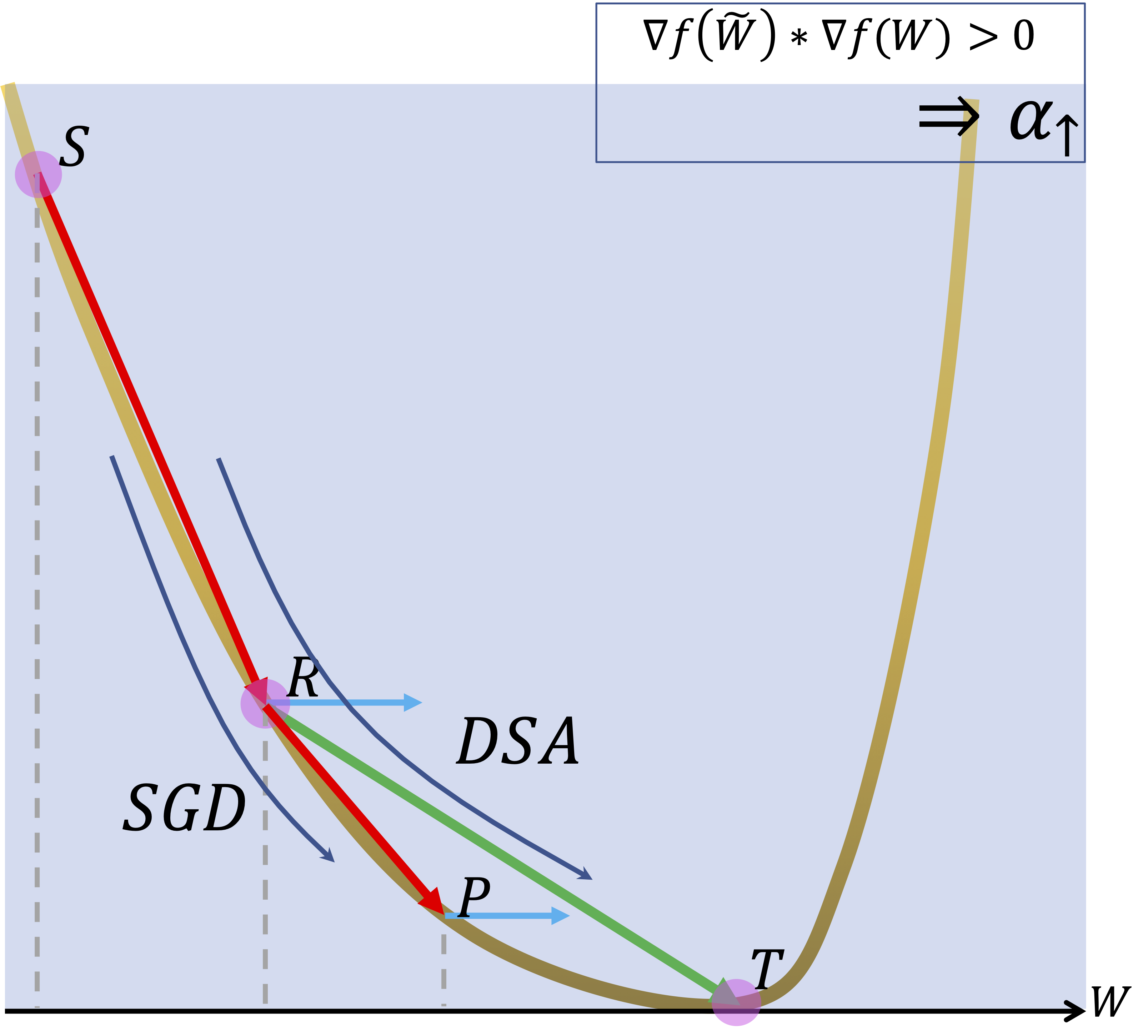}
        \centerline{(b)}
    \end{minipage}
    \caption{Update rule for learning rate in DSA}
    \label{fig: dsa}
\end{figure}
The most important difference from HD is the detection with current learning rate, i.e., $\widetilde{W}$. If $\widetilde{W}$ crosses the extreme point, the learning rate should decrease. Otherwise, the learning rate will increase. In other words, HD makes decision according to the history while DSA according to the future. We visualize the update rule of learning rate in Fig.~\ref{fig: dsa}. Parameter $W$ steps from the start point S, reaching and taking detection at R. P is the detected point with $\alpha_{t-1}$, i.e., the position of $\widetilde{W}$. If gradient at P has the same direction to R, i.e., $\nabla f(\widetilde{W})\cdot\nabla f(W) > 0$, learning rate will increase. Otherwise it decreases. Obviously, DSA can avoid HD's misunderstanding perfectly.
\subsubsection{Dealing with Minibatch Training\label{sec: dsa_mini}}
Another pitfall of HD is the ability of processing large datasets using minibatch training. Whlie DSA will get into the trap more easily than HD as shown in Fig.~\ref{fig: hd_mini}. Because DSA can fit a convex function far more quickly than HD and other optimizers. In other word, the ideal occasion for DSA is that only one cost function exists in the training, i.e., batch training. Fortunately, we can use DSA to train a neural network in the form of batch training, where the network is pretrained with Momentum optimizer on minibatch data. Evidence told us this kind of further training using DSA could always enhance the network to a stronger one and we will show this in the experiment.
\subsubsection{Further Improvements for DSA\label{sec: trick}}
In this part, we will give three techniques to make DSA more sensitive, efficient and safe. Skills include \textit{parameter specific learning rate}, \textit{intenal structured learning rate} and \textit{step size equal learning rate}.
\\\textbf{Parameter Specific Learning Rate}\\
In normal machine learning algorithm, learning rate is always a scalar, while parameters' requirements for learning rate is various on the scale. Therefore, we try to make it a vector and specific for every parameter.
The proposed paper of HD has denoted that \textit{It is straightforward to generalize the introduced method to the case where $\alpha$ is a vector of per-parameter learning rates}~\cite{hypergradient}.
However, truth obeys wishes. Actually, the gradient for $\alpha$ is a result of two vectors' inner product
\begin{equation}
\label{equ: dsa_alpha_gradient_basic}
\nabla{f(\widetilde{W})} \cdot \nabla{f(W_{t-1})}=\sum_{i}{\nabla{f(\widetilde{W})}_i\nabla{f(W_{t-1})}_i}.
\end{equation}
If we directly generalize the update rule to the form of parameter specific, it would be
\begin{equation}
\label{equ: dsa_parameter_specific}
\nabla{f(\alpha)_i}=-\nabla{f(\widetilde{W})}_i\nabla{f(W_{t-1})}_i.
\end{equation}
However, the famous problem grad loss will be more significant in formula~\ref{equ: dsa_parameter_specific}. The scale of $\nabla{f(\widetilde{W})}_i\nabla{f(W_{t-1})}_i$ is always near $10^{-6}$ and even smaller, which makes learning rate adaptation stay in name only. Given that, we use learning rate's gradient as an indicator. That is, $\alpha$ will increase a step of $\beta$ when the gradient is positive. The update rule of learning rate turns into
\begin{align}
\label{equ: dsa_alpha_rule_direction}
\begin{split}
\alpha_i^{(t)}=\alpha_i^{(t-1)} + \beta\frac{\nabla{f(\widetilde{W})}_i\nabla{f(W^{(t-1)})}_i}{|\nabla{f(\widetilde{W})}_i\nabla{f(W^{(t-1)})}_i| + \epsilon},
\end{split}
\end{align}
where $\nabla{f(\widetilde{W})}_i\nabla{f(W_{t-1})}_i/(|\nabla{f(\widetilde{W})}_i\nabla{f(W_{t-1})}_i| + \epsilon) \in \{-1,1,0\}$ and $\epsilon$ is a infinitesimal.
\\\textbf{Internal Structured Learning Rate}\\
The formula~\ref{equ: dsa_alpha_rule_direction} is not safe, because $\alpha$ is possible to be negtive or oversized. Here we endow learning rate an internal structure $\sigma(\alpha) * \gamma$, where $\gamma=0.1$ and $\sigma$ is sigmoid activation
\begin{equation*}
\label{equ: sigmoid}
\sigma(\alpha) = \frac{1}{1 + \exp^{-\alpha}}.
\end{equation*}
As a result, learning rate $\sigma(\alpha) * \gamma$ is in the range of $(0,0.1)$. Straightforward to derive that the update rule for $\alpha$ is still formula~\ref{equ: dsa_alpha_rule_direction}. And the step of $W$ will be
\begin{align}
\label{equ: dsa_parameter_rule_final}
\begin{split}
W^{(t)} = W^{(t-1)} - \frac{\gamma}{1 + \exp\{-\alpha^{(t)}\}} * \nabla{f(W^{(t-1)})}.
\end{split}
\end{align}
\\\textbf{Step Size Equal Learning Rate}\\
Sometimes, grad loss of parameter is also prominent, such as the usage of sigmoid activation. Similar with learning rate, we can use $\nabla f(W^{(t-1)})/(|\nabla f(W^{(t-1)})| + \epsilon)$ to indicate the step direction of $W$. As a result, the step size of $W$ is determind by learning rate completely, i.e., step size equals learning rate.
\begin{equation}
\label{equ: dsa_parameter_direction}
W^{(t)} = W^{(t-1)} - \frac{\gamma}{1 + \exp\{-\alpha^{(t)}\}} * \frac{\nabla f(W^{(t-1)})}{|\nabla f(W^{(t-1)})| + \epsilon}
\end{equation}
This technique is proprietary for DSA, because only DSA could adapt learning rate with enough sensitivity.
\subsubsection{Pseudo Code of DSA\label{sec: pseudo}}
Pseudo code of DSA is in Algorithm~\ref{algo: dsa}. Line~2 is detection of parameters. Line~4 adapts learning rate according to formula~\ref{equ: dsa_alpha_rule_direction}. Line~5 recomputes values of parameter. Note that formula~\ref{equ: dsa_parameter_direction} is an optional technique.
\begin{algorithm}[h]
\centering
\caption{Differentiable Self-Adaptive Learning Rate}
\label{algo: dsa}
\begin{algorithmic}[1]
\Require cost function $f(W)$, learning rate $\gamma\sigma(\alpha)$, $\alpha$'s step size $\beta$, total iterations $T$, infinitesimal $\epsilon$
\Ensure trained parameters $W^{(T)}$
\For {$t:1\mapsto T$}
    \State $\widetilde{W} = W^{(t-1)} - \gamma\sigma(\alpha^{(t-1)}) * \nabla{f(W^{(t-1)})}$
    \State \scriptsize{$\Delta \alpha^{(t-1)} = \beta{(\nabla{f(\widetilde{W})}\nabla{f(W^{(t-1)})})}/{(|\nabla{f(\widetilde{W})}\nabla{f(W^{(t-1)})}| + \epsilon)}$}\normalsize
    \State $\alpha^{(t)}=\alpha^{(t-1)} + \Delta \alpha^{(t-1)}$
    \State $W^{(t)} = W^{(t-1)} - \gamma\sigma(\alpha^{(t)}) * \nabla{f(W^{(t-1)})}$
\EndFor
\end{algorithmic}
\end{algorithm}
\section{Experiment\label{sec: exp}}
To verify the performance of proposed approaches, we conduct extensive experiments. In this section, we first introduce the basic settings necessary for experiments in Section~\ref{sec: exp setting}. Then we will show the results and take analyses in Section~\ref{sec: exp res}. Nextly, two case studies will be conducted in Section~\ref{sec: case study}. Finally, ablation experiment and sensitivity analysis are taken in Section~\ref{sec: sensitivity}.
\begin{table}[h]
\begin{minipage}{\linewidth}
\caption{Large dataset information}
\label{tab: data_meta_large}
\centering
\begin{tabular}{lrrr}
    \textbf{Dataset} & \#\textbf{Train}/ \#\textbf{Test}&\#\textbf{Attributes}&\#\textbf{Class} \\
    \toprule
    MNIST &60,000 /10,000 & 1*28*28& 10\\
    SVHN &73,257 /26,032 &3*32*32 & 10 \\
    CIFAR10 &50,000 /10,000 &3*32*32 &10 \\
    CIFAR100 &50,000 /10,000 &3*32*32 &100 \\
\end{tabular}
\end{minipage}
\begin{minipage}{\linewidth}
\caption{Small dataset information}
\label{tab: data_meta_small}
\centering
\begin{tabular}{lrrr}
    \textbf{Dataset} & \#\textbf{Train}/ \#\textbf{Test}&\#\textbf{Attributes}&\#\textbf{Class} \\
    \toprule
    IRIS & 120/ 30& 4 & 3\\
    WINE & 142/ 36& 13&3\\
    CAR & 1,382/ 346& 6&4 \\
    AGARICUS &6,499 / 1,625&116 & 2\\
\end{tabular}
\end{minipage}
\end{table}
\begin{table*}[h]
\begin{minipage}{0.5\linewidth}
    \caption{Train ResNet with CIFAR10}
    \label{tab: resnet cifar10}
    \centering
    \begin{tabular}{lrrrr}
            & ACCU & F1-SCORE & RECALL & PRECISION \\
        \toprule
        SGD & 93.29 & 86.23$\sim$97.12 & 86.98$\sim$96.54 & 85.50$\sim$97.70 \\
        Momentum$\star$ & 94.16 & 87.07$\sim$97.02 & 85.96$\sim$97.16 & 88.20$\sim$97.80 \\
        RMSPprop & 92.02 & 84.14$\sim$96.37 & 82.17$\sim$97.28 & 84.80$\sim$97.00 \\
        AdaDelta & 92.56 & 84.09$\sim$96.44 & 86.36$\sim$96.56 & 81.90$\sim$97.40 \\
        AdaGrad & 91.23 & 82.36$\sim$95.98 & 84.53$\sim$96.37 & 80.30$\sim$96.10 \\
        Adam & 93.82 & 87.35$\sim$96.86 & 88.75$\sim$96.62 & 86.00$\sim$97.10 \\
        HD & 87.03 & 76.04$\sim$93.75 & 75.00$\sim$95.35 & 77.10$\sim$93.10 \\
        DSA & \bf 94.60 & 88.55$\sim$97.50 & 89.32$\sim$97.31 & 87.80$\sim$97.70 \\
    \end{tabular}
\end{minipage}
\begin{minipage}{0.5\linewidth}
    \caption{Train ResNet with CIFAR100}
    \label{tab: resnet cifar100}
    \centering
    \begin{tabular}{lrrrr}
            & ACCU & F1-SCORE & RECALL & PRECISION \\
        \toprule
        SGD & 73.74 & 45.77$\sim$93.60 & 43.22$\sim$94.51 & 45.00$\sim$95.00 \\
        Momentum$\star$ & 77.08 & 54.00$\sim$94.42 & 54.00$\sim$95.88 & 52.00$\sim$96.00 \\
        RMSPprop & 65.56 & 34.62$\sim$90.36 & 37.50$\sim$91.75 & 27.00$\sim$94.00 \\
        AdaDelta & 71.48 & 36.48$\sim$91.71 & 49.15$\sim$90.53 & 29.00$\sim$95.00 \\
        AdaGrad & 68.87 & 41.05$\sim$91.00 & 40.62$\sim$91.00 & 39.00$\sim$93.00 \\
        Adam & 72.46 & 45.03$\sim$91.18 & 44.34$\sim$95.56 & 43.00$\sim$95.00 \\
        HD & 69.16 & 39.34$\sim$90.10 & 33.33$\sim$95.35 & 34.00$\sim$93.00 \\
        DSA & \bf 77.19 & 54.55$\sim$94.00 & 55.10$\sim$94.00 & 54.00$\sim$96.00 \\
    \end{tabular}
\end{minipage}
\end{table*}
\begin{table*}[h]
\begin{minipage}{0.5\linewidth}
\caption{Train ResNet with MNIST}
\label{tab: resnet mnist}
\centering
\begin{tabular}{lrrrr}
        & ACCU & F1-SCORE & RECALL & PRECISION \\
    \toprule
    SGD & 99.14 & 97.53$\sim$99.74 & 97.58$\sim$99.90 & 97.48$\sim$100.0 \\
    Momentum$\star$ & 99.27 & 97.97$\sim$99.82 & 97.97$\sim$99.90 & 97.97$\sim$100.0 \\
    RMSPprop & 98.92 & 97.04$\sim$99.78 & 97.09$\sim$99.80 & 97.00$\sim$99.91 \\
    AdaDelta & 99.18 & 97.83$\sim$99.65 & 97.59$\sim$99.80 & 97.87$\sim$100.0 \\
    AdaGrad & 98.99 & 97.29$\sim$99.69 & 97.29$\sim$99.80 & 97.29$\sim$100.0 \\
    Adam & 99.26 & 97.92$\sim$99.80 & 97.59$\sim$99.80 & 97.98$\sim$100.0 \\
    HD & 98.54 & 96.75$\sim$99.69 & 95.39$\sim$99.69 & 96.24$\sim$99.82 \\
    DSA & \bf 99.35 & 98.31$\sim$99.78 & 98.07$\sim$99.80 & 97.98$\sim$99.91 \\
\end{tabular}
\end{minipage}
\begin{minipage}{0.5\linewidth}
\caption{Train ResNet with SVHN}
\label{tab: resnet svhn}
\centering
\begin{tabular}{lrrrr}
        & ACCU & F1-SCORE & RECALL & PRECISION \\
    \toprule
    SGD & 96.23 & 94.69$\sim$97.17 & 94.63$\sim$97.08 & 94.66$\sim$97.30 \\
    Momentum$\star$ & 96.34 & 94.72$\sim$97.41 & 94.37$\sim$97.52 & 94.07$\sim$97.80 \\
    RMSPprop & 95.34 & 93.55$\sim$96.64 & 90.64$\sim$97.57 & 93.21$\sim$97.54 \\
    AdaDelta & 95.66 & 94.49$\sim$96.55 & 94.16$\sim$97.14 & 94.31$\sim$96.82 \\
    AdaGrad & 95.37 & 93.06$\sim$96.47 & 92.40$\sim$96.62 & 93.55$\sim$96.67 \\
    Adam & 96.30 & 94.60$\sim$97.07 & 94.46$\sim$97.21 & 94.52$\sim$97.53 \\
    HD & 93.93 & 90.83$\sim$95.62 & 88.79$\sim$96.99 & 89.14$\sim$96.65 \\
    DSA & \bf 96.64 & 95.19$\sim$97.84 & 94.17$\sim$97.97 & 94.66$\sim$97.70 \\
\end{tabular}
\end{minipage}
\end{table*}
\begin{table*}[h]
\begin{minipage}{0.5\linewidth}
\caption{Train DNN with MNIST}
\label{tab: dnn mnist}
\centering
\begin{tabular}{lrrrr}
        & ACCU & F1-SCORE & RECALL & PRECISION \\
    \toprule
    Adam & 98.56 & 96.80$\sim$99.56 & 96.73$\sim$99.73 & 96.75$\sim$99.69 \\
    Adamax$\star$ & 98.70 & 97.19$\sim$99.65 & 96.92$\sim$99.82 & 96.97$\sim$99.69 \\
    DSA & \bf 98.71 & 97.41$\sim$99.74 & 97.58$\sim$99.82 & 97.09$\sim$99.80 \\
\end{tabular}
\end{minipage}
\begin{minipage}{0.5\linewidth}
\caption{Train DNN with SVHN}
\label{tab: dnn svhn}
\centering
\begin{tabular}{lrrrr}
        & ACCU & F1-SCORE & RECALL & PRECISION \\
    \toprule
    Adam & 29.11 & 0.00$\sim$44.08 & 0.00$\sim$36.88 & 0.00$\sim$72.33 \\
    Adamax$\star$ & 89.77 & 84.87$\sim$93.57 & 83.66$\sim$93.14 & 82.29$\sim$94.55 \\
    DSA & \bf 90.72 & 86.89$\sim$93.58 & 85.50$\sim$94.02 & 85.12$\sim$95.12 \\
\end{tabular}
\end{minipage}
\end{table*}
\begin{table*}[h]
\begin{minipage}{0.5\linewidth}
\caption{Train FMP with MNIST}
\label{tab: fmp mnist}
\centering
\begin{tabular}{lrrrr}
        & ACCU & F1-SCORE & RECALL & PRECISION \\
    \toprule
    Adamax$\star$ & 99.26 & 98.36$\sim$99.69 & 98.26$\sim$99.79 & 97.76$\sim$99.90 \\
    DSA & \bf 99.30 & 98.41$\sim$99.74 & 98.08$\sim$99.82 & 98.54$\sim$99.80 \\
\end{tabular}
\end{minipage}
\begin{minipage}{0.5\linewidth}
\caption{Train FMP with SVHN}
\label{tab: fmp svhn}
\centering
\begin{tabular}{lrrrr}
        & ACCU & F1-SCORE & RECALL & PRECISION \\
    \toprule
    Adamax$\star$ & 96.04 & 93.50$\sim$97.04 & 91.95$\sim$97.13 & 92.30$\sim$97.54 \\
    DSA & \bf 96.07 & 93.80$\sim$97.13 & 91.88$\sim$96.99 & 92.30$\sim$97.45 \\
\end{tabular}
\end{minipage}
\end{table*}
\subsection{Experiment Setting\label{sec: exp setting}}
\begin{figure}[h]
    \centering
    \includegraphics[width=\linewidth]{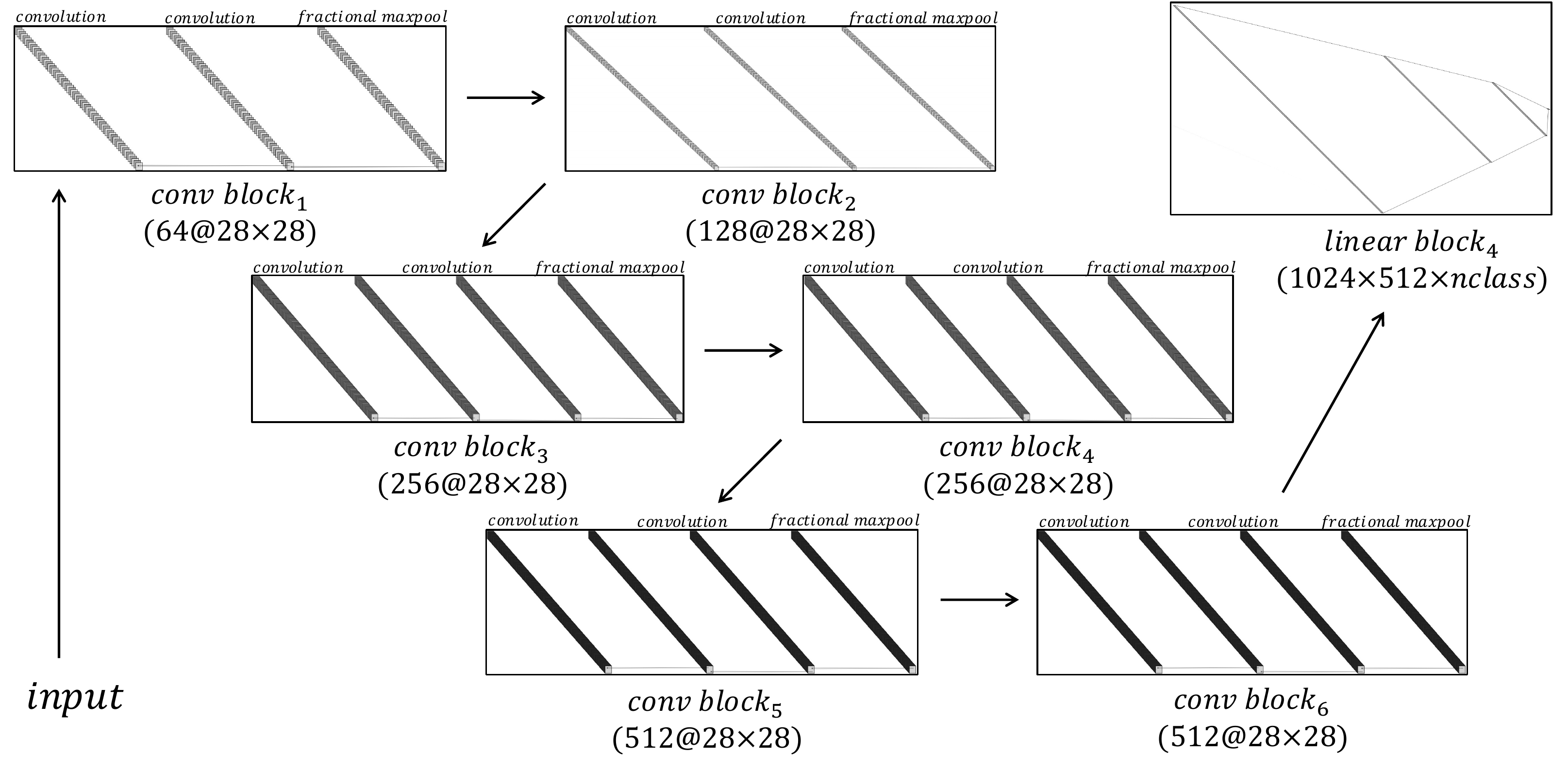}
    \caption{Structure of FMP.}
    \label{fig: fmp}
\end{figure}

\textbf{Neural Networks and Datasets}\\
We applied DSA to ResNet~\cite{resnet}, FMP~\cite{fmp}, DNN~\cite{dnn1,dnn2} and MLP~\cite{mlp}. Famous ResNet takes deep residual learning and here we apply ResNet-18. DNN is the common deep convolution neural network. The core of FMP is the fractional maxpool layer and it's visualized as Fig.~\ref{fig: fmp} supposing the input is a image of $28\times28$. FMP is composed of 6 convolution block and 1 linear block. Each convolution block ends up with a fractional maxpool. Each output of a convolution layer is processed by prelu activation.
On convolution neural network, we choose four published and well known validation datasets MNIST~\cite{mnist}, SVHN~\cite{svhn}, CIFAR10 and CIFAR100~\cite{cifar}. On MLP, we choose IRIS, WINE, CAR and AGARICUS\footnote{\url{https://archive.ics.uci.edu/ml/datasets}} as validation data set, which are distinguishable to different optimizers. The meta information of selected datasets is shown in TABLE~\ref{tab: data_meta_large} and TABLE~\ref{tab: data_meta_small}.
\\\textbf{Baseline}\\
Selected optimizers include SGD~\cite{gradientdescent}, Momentum~\cite{momentum}, AdaGrad~\cite{adagrad}, AdaDelta~\cite{adadelta}, RMSProp~\cite{rmsprop}, Adam~\cite{adam}, Adamax~\cite{adam}, AdamW~\cite{adamw} and Hypergradient Descent~\cite{hypergradient}. Note that, Hypergradient Descent has mutiple versions and we choose the one based on SGD here.
\\\textbf{Metrics}\\
We recorded accuracy, F1-score, recall and precision on validation dataset to measure the effect of different optimizers from different perspectives~\cite{scikit-learn}. Additionally, the train loss is recorded to demonstrate the convergence effect and miss rate is used to compare DSA and HD.
\\\textbf{Implementations}\\
The structure of FMP is visualized in Fig.~\ref{fig: fmp} and FMP is designed with reference to~\cite{3DSemanticSegmentationWithSubmanifoldSparseConvNet,SubmanifoldSparseConvNet}, which does very well on MNIST and SVHN. FMP has a huge mount of parameters, which is a challenge to DSA.
DNN is designed with reference to~\cite{dnn1,dnn2,dnn3} and we set channel size as a smaller value to simplify the model so that it can converge faster than the original.
MLP sequences 5 fully connected layers. The dimensions of each layer are: 32, 64, 256 and 128. The output of the first and third layers are processed using sigmoid activation. The output of the second and the fourth layers are processed by prelu activation~\cite{prelu}. The features output by the neural network are processed by log\_softmax activation and cross-entropy loss function.
The number of training epochs on large datasets is 200 and that on small datasets is commonly 1000. Actually, DSA will take one more loop than other adaptation optimizers in each iteration, so we always tend to use fewer epochs when applying DSA. DSA's experiments on large datasets is based on pretrained model, the pretraining is 180 iterations and the formal training of DSA is 10 epochs.
The batchsize of MNIST, SVHN, CIFAR10, CIFAR100 is 128, 64, 128, 32 respectively.
Large datasets are preprocessed with some efficient skills including random crop for images, random horizontal flip for images and normalization for each image.
When training convolution neural network with SGD, Momentum and Adam, we apply a learning rate scheduler where learning rate will be scaled $\gamma$ times than the current in the half and three quarters of the training session. Commonly, $\gamma = 0.1$.
Pretrained ResNet is obtained from the model trained by Momentum cooperated with learning rate scheduler. Pretraining stops at some iteration after the second learning rate scaling. While if neural network is DNN or FMP, we will apply Adamax as optimizer to pretrain the model. Because there are significant grad loss in DNN and huge amount of parameters in FMP, which are both hard problems for Momentum. If we use SGD or Momentum to pretrain DNN and FMP, ten thousands of iterations would be needed. Adamax is practically the most superior baseline here.
SGD's initial learning rate is 0.1. Momentum's initial learning rate is 0.1 and the momentum rate is 0.9. AdaGrad's initial learning rate is 0.01. RMSProp's initial learning rate is 0.01. Adam and AdamW's initial learning rate is 0.001. Adamax's initial learning rate is 0.002. HD's initial learning rate is 0.1 and step size of learning rate is 0.01. DSA's step size of $\alpha$ is 0.1 and initial learning rate is 0.001, i.e., initial $\alpha$ is -4.6. When dealing with large image datasets on pretrained model, the initial learning rate of DSA is $10^{-5}$ and the step size of $\alpha$ is 0.3.
Above settings for hyper-parameters refer to the proposal~\cite{NEURIPS2019_9015}.
Our experiments are conducted on GTX 3060Ti GPU for all the groups.
\begin{table*}[h]
\begin{minipage}{0.5\linewidth}
\caption{Train MLP with WINE (1000 Epochs)}
\label{tab: mlp wine}
\centering
\begin{tabular}{lrrrr}
        & ACCU & F1-SCORE & RECALL & PRECISION \\
    \toprule
    SGD & 44.44 & 0.0$\sim$61.54 & 0.0$\sim$44.44 & 0.0$\sim$100.0 \\
    Momentum & 44.44 & 0.0$\sim$61.54 & 0.0$\sim$44.44 & 0.0$\sim$100.0 \\
    RMSPprop & 98.28 & 97.30$\sim$100.0 & 96.88$\sim$100.0 & 94.74$\sim$100.0 \\
    AdaDelta & 86.21 & 55.56$\sim$97.30 & 50.00$\sim$100.0 & 62.50$\sim$94.74 \\
    AdaGrad & \bf100.0 & 100.0$\sim$100.0 & 100.0$\sim$100.0 & 100.0$\sim$100.0 \\
    Adam & 97.22 & 92.31$\sim$100.0 & 85.71$\sim$100.0 & 93.75$\sim$100.0 \\
    AdamW & \bf100.0 & 100.0$\sim$100.0 & 100.0$\sim$100.0 & 100.0$\sim$100.0 \\
    Adamax & 98.28 & 94.12$\sim$100.0 & 88.89$\sim$100.0 & 96.77$\sim$100.0 \\
    HD & 75.00 & 0.0$\sim$88.00 & 0.0$\sim$100.0 & 0.0$\sim$100.0 \\
    DSA & \bf100.0 & 100.0$\sim$100.0 & 100.0$\sim$100.0 & 100.0$\sim$100.0 \\
\end{tabular}
\end{minipage}
\begin{minipage}{0.5\linewidth}
\caption{Train MLP with CAR (500 Epochs)}
\label{tab: mlp car}
\centering
\begin{tabular}{lrrrr}
        & ACCU & F1-SCORE & RECALL & PRECISION \\
    \toprule
    SGD & 69.36 & 0.0$\sim$81.91 & 0.0$\sim$69.36 & 0.0$\sim$100.0 \\
    Momentum & 93.06 & 62.50$\sim$97.49 & 61.54$\sim$97.90 & 58.82$\sim$97.08 \\
    RMSPprop & 90.75 & 44.44$\sim$96.60 & 42.11$\sim$98.70 & 35.29$\sim$94.58 \\
    AdaDelta & 88.44 & 0.0$\sim$97.26 & 0.0$\sim$98.30 & 0.0$\sim$96.25 \\
    AdaGrad & 91.62 & 60.00$\sim$96.48 & 53.85$\sim$95.88 & 52.94$\sim$97.08 \\
    Adam & 99.42 & 98.75$\sim$100.0 & 97.53$\sim$100.0 & 99.17$\sim$100.0 \\
    AdamW & 99.42 & 94.74$\sim$99.79 & 94.44$\sim$100.0 & 90.00$\sim$100.0 \\
    Adamax & 97.11 & 75.00$\sim$99.58 & 64.29$\sim$100.0 & 70.59$\sim$99.17 \\
    HD & 69.36 & 0.0$\sim$81.91 & 0.0$\sim$69.36 & 0.0$\sim$100.0 \\
    DSA & \bf100.0 & 100.0$\sim$100.0 & 100.0$\sim$100.0 & 100.0$\sim$100.0 \\
\end{tabular}
\end{minipage}
\end{table*}
\begin{table*}[h]
\begin{minipage}{0.5\linewidth}
\caption{Train MLP with IRIS (30 Epochs)}
\label{tab: mlp iris}
\centering
\begin{tabular}{lrrrr}
        & ACCU & F1-SCORE & RECALL & PRECISION \\
    \toprule
    SGD & 43.33 & 0.0$\sim$60.47 & 0.0$\sim$43.33 & 0.0$\sim$100.0 \\
    Momentum & 36.67 & 0.0$\sim$53.66 & 0.0$\sim$36.67 & 0.0$\sim$100.0 \\
    RMSPprop & \bf100.0 & 100.0$\sim$100.0 & 100.0$\sim$100.0 & 100.0$\sim$100.0 \\
    AdaDelta & 56.67 & 0.0$\sim$100.0 & 0.0$\sim$100.0 & 0.0$\sim$100.0 \\
    AdaGrad & \bf100.0 & 100.0$\sim$100.0 & 100.0$\sim$100.0 & 100.0$\sim$100.0 \\
    Adam & 60.00 & 14.29$\sim$100.0 & 33.33$\sim$100.0 & 7.69$\sim$100.0 \\
    AdamW & 56.67 & 0.0$\sim$100.0 & 0.0$\sim$100.0 & 0.0$\sim$100.0 \\
    Adamax & 56.67 & 0.0$\sim$100.0 & 0.0$\sim$100.0 & 0.0$\sim$100.0 \\
    HD & 20.00 & 0.0$\sim$33.33 & 0.0$\sim$20.00 & 0.0$\sim$100.0 \\
    DSA & \bf100.0 & 100.0$\sim$100.0 & 100.0$\sim$100.0 & 100.0$\sim$100.0 \\
\end{tabular}
\end{minipage}
\begin{minipage}{0.5\linewidth}
\caption{Train MLP with AGAICUS (100 Epochs)}
\label{tab: mlp agaricus}
\centering
\begin{tabular}{lrrrr}
        & ACCU & F1-SCORE & RECALL & PRECISION \\
    \toprule
    SGD & 52.43 & 0.0$\sim$68.79 & 0.0$\sim$52.43 & 0.0$\sim$100.0 \\
    Momentum & 52.43 & 0.0$\sim$68.79 & 0.0$\sim$52.43 & 0.0$\sim$100.0 \\
    RMSPprop & 99.94 & 99.94$\sim$99.94 & 99.87$\sim$100.0 & 99.88$\sim$100.0 \\
    AdaDelta & 79.88 & 73.17$\sim$83.90 & 72.26$\sim$100.0 & 57.70$\sim$100.0 \\
    AdaGrad & \bf100.0 & 100.0$\sim$100.0 & 100.0$\sim$100.0 & 100.0$\sim$100.0 \\
    Adam & 99.94 & 99.94$\sim$99.94 & 99.87$\sim$100.0 & 99.88$\sim$100.0 \\
    AdamW & 99.94 & 99.94$\sim$99.94 & 99.87$\sim$100.0 & 99.88$\sim$100.0 \\
    Adamax & 99.88 & 99.87$\sim$99.88 & 99.87$\sim$99.88 & 99.87$\sim$99.88 \\
    HD & 52.43 & 0.0$\sim$68.79 & 0.0$\sim$52.43 & 0.0$\sim$100.0 \\
    DSA & \bf100.0 & 100.0$\sim$100.0 & 100.0$\sim$100.0 & 100.0$\sim$100.0 \\
\end{tabular}
\end{minipage}
\end{table*}
\begin{figure*}[h]
\centering
\includegraphics[width=\linewidth]{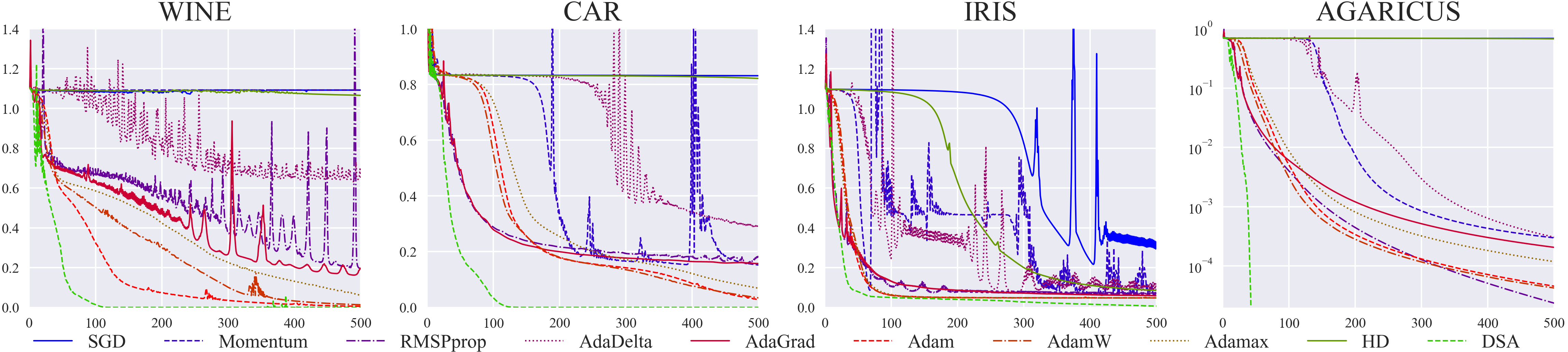}
\caption{Train loss on MLP}
\label{fig: mlp_loss}
\end{figure*}
\begin{figure*}[h]
\centering
\includegraphics[width=\linewidth]{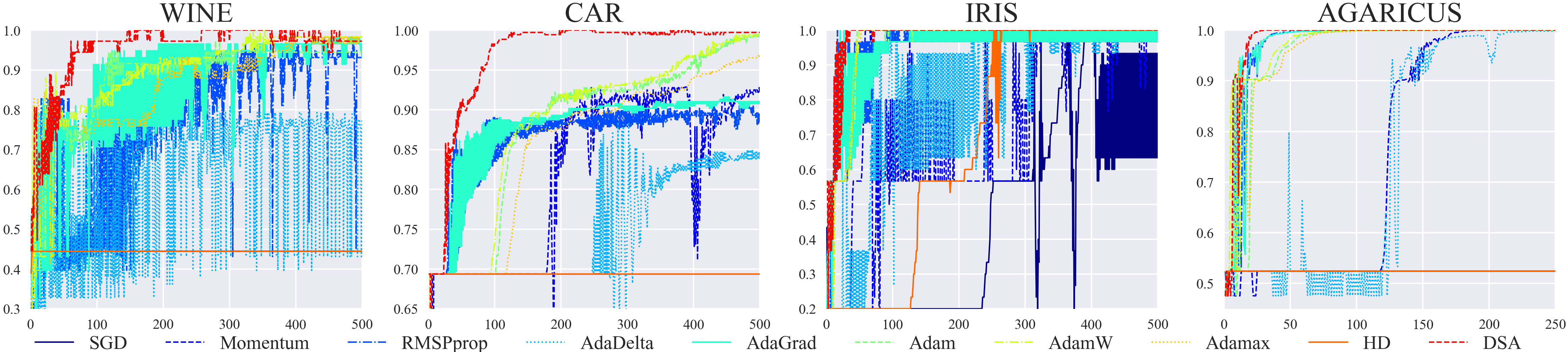}
\caption{Validation accuracy on MLP}
\label{fig: mlp_accu}
\end{figure*}
\begin{figure*}[h]
    \centering
    \includegraphics[width=\linewidth]{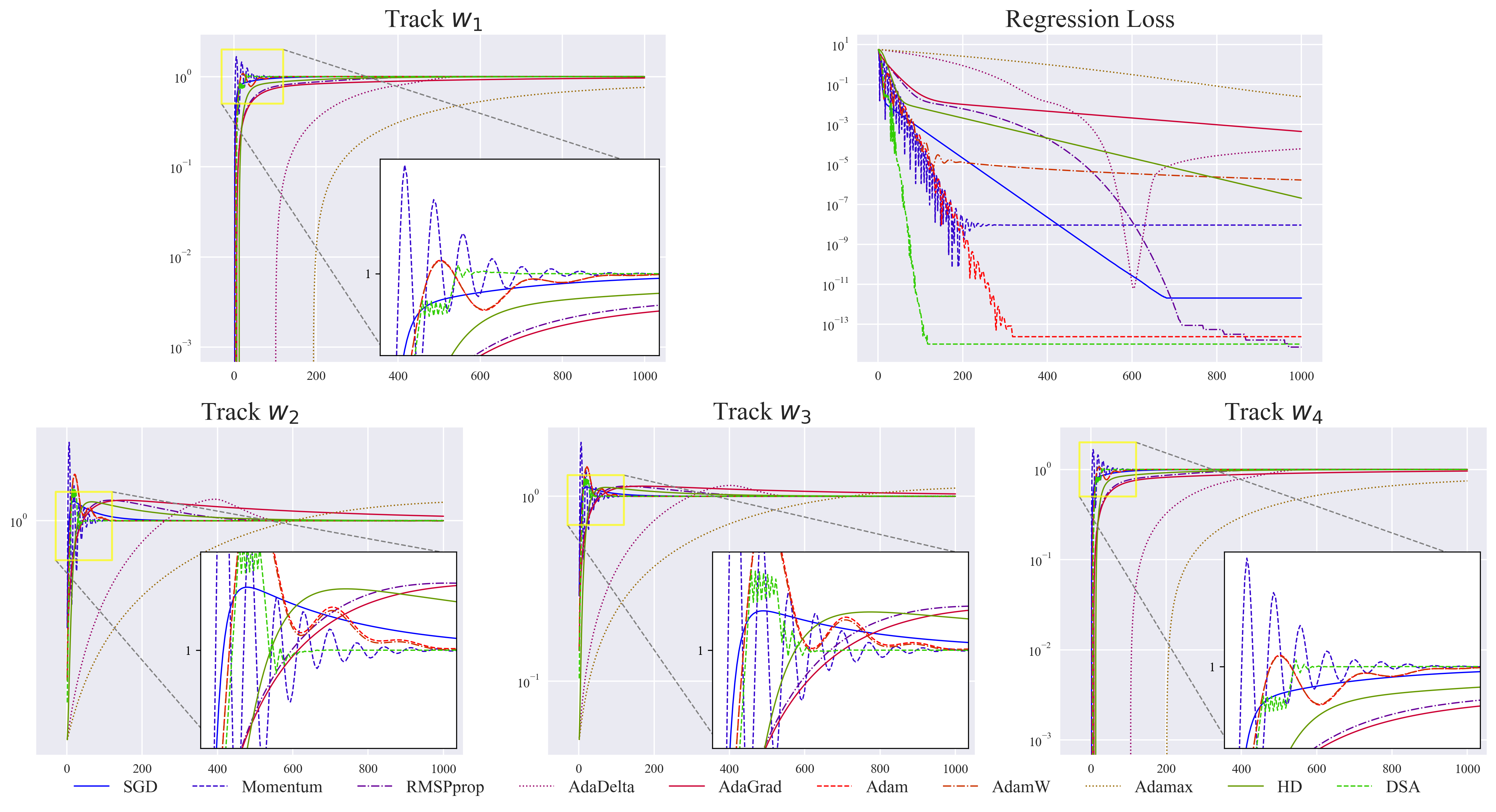}
    \caption{Regression loss and track of parameters}
    \label{fig: case_sum}
\end{figure*}
\begin{figure*}[h]
\begin{minipage}{\linewidth}
    \centering
    \begin{minipage}{0.246\linewidth}
        \includegraphics[width=\linewidth]{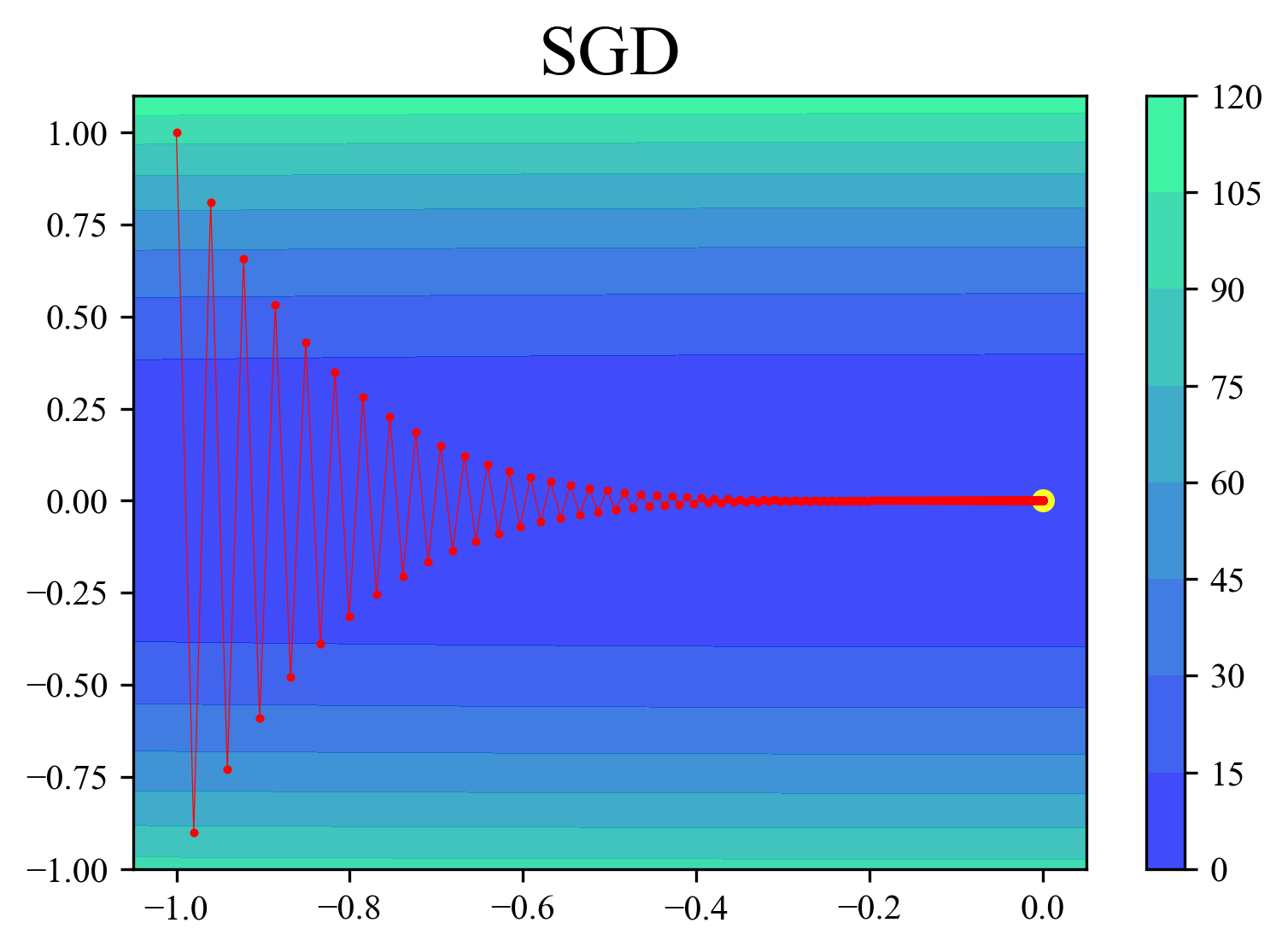}
    \end{minipage}
    \begin{minipage}{0.246\linewidth}
        \includegraphics[width=\linewidth]{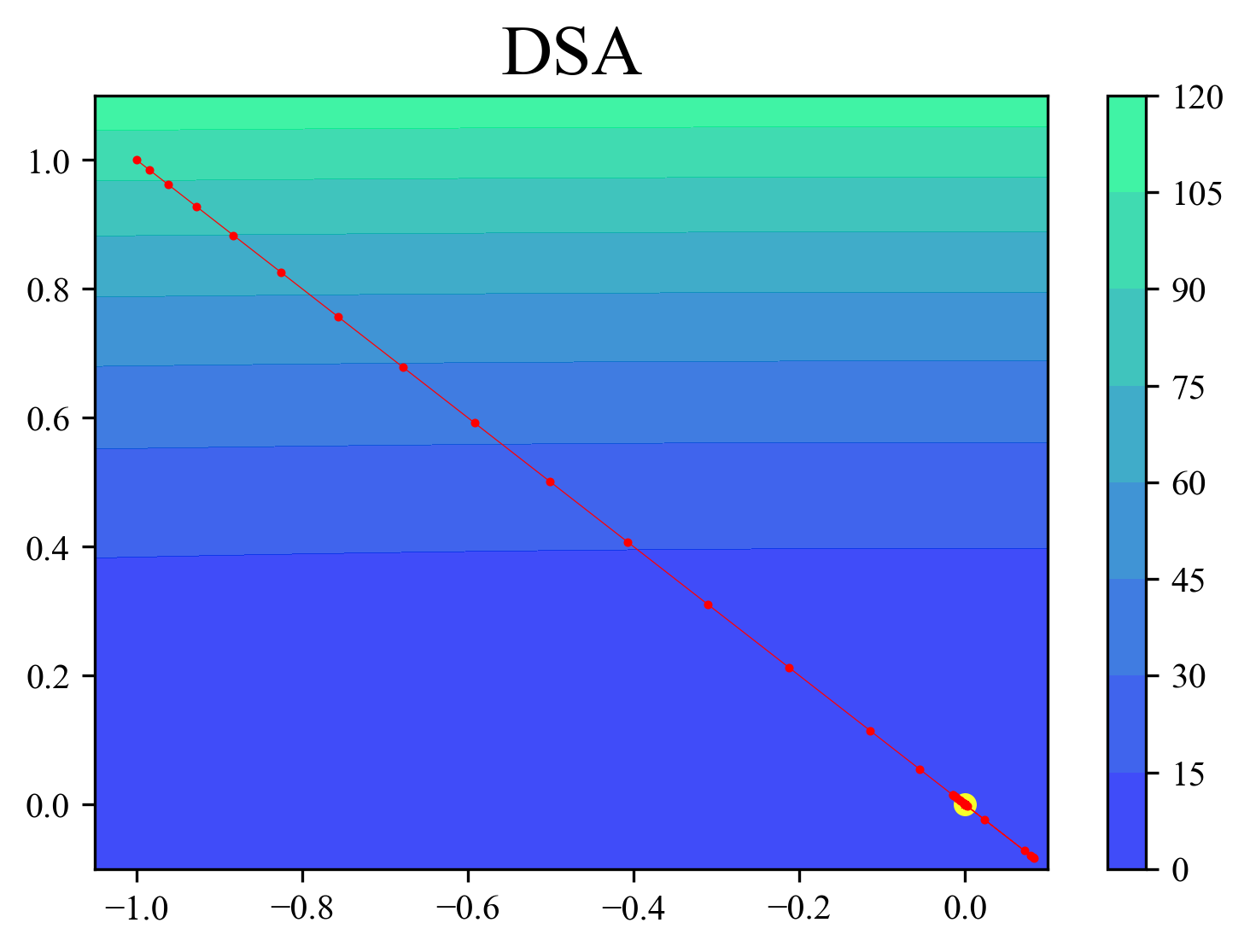}
    \end{minipage}
    \begin{minipage}{0.246\linewidth}
        \includegraphics[width=\linewidth]{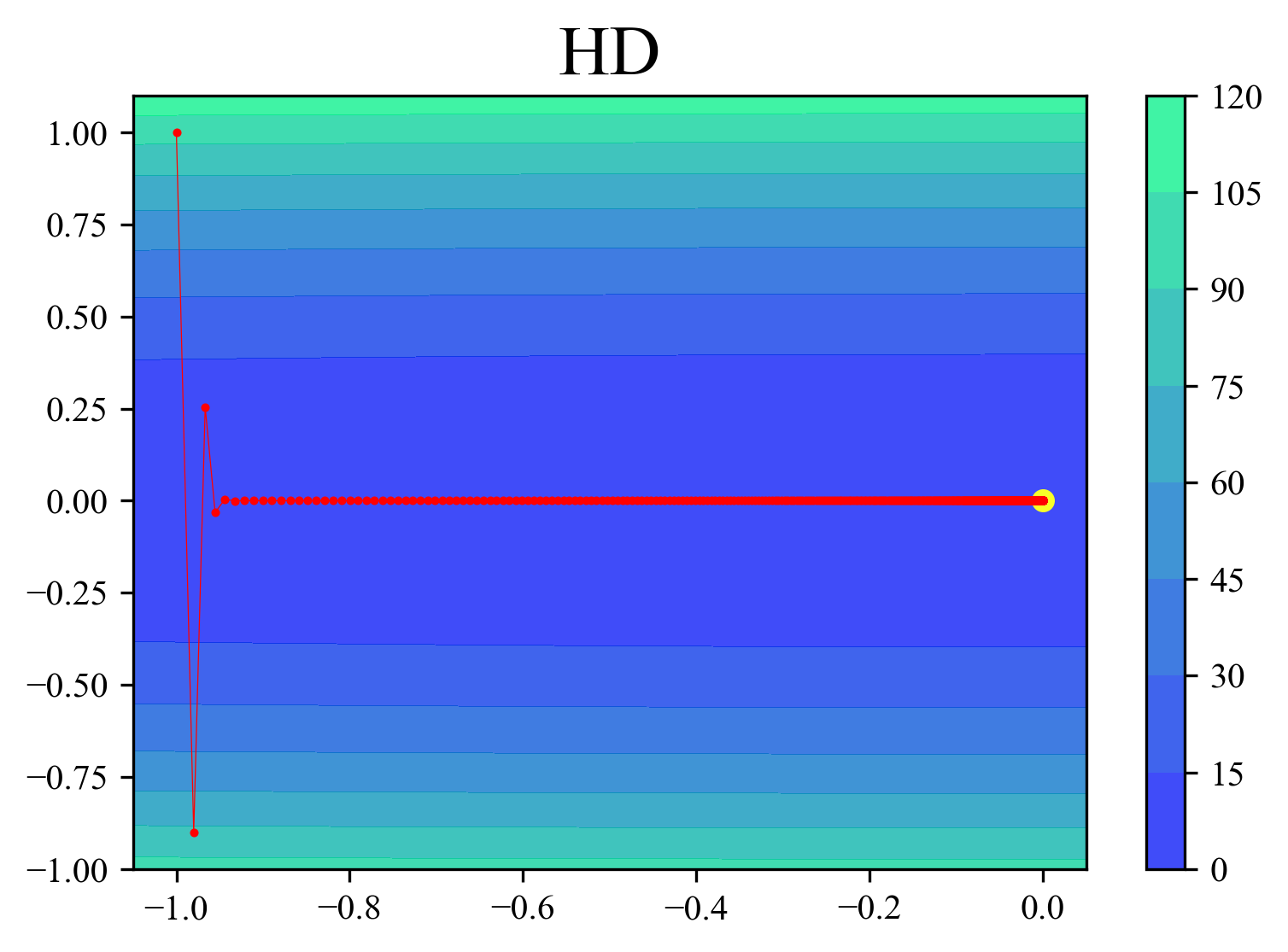}
    \end{minipage}
    \begin{minipage}{0.246\linewidth}
        \includegraphics[width=\linewidth]{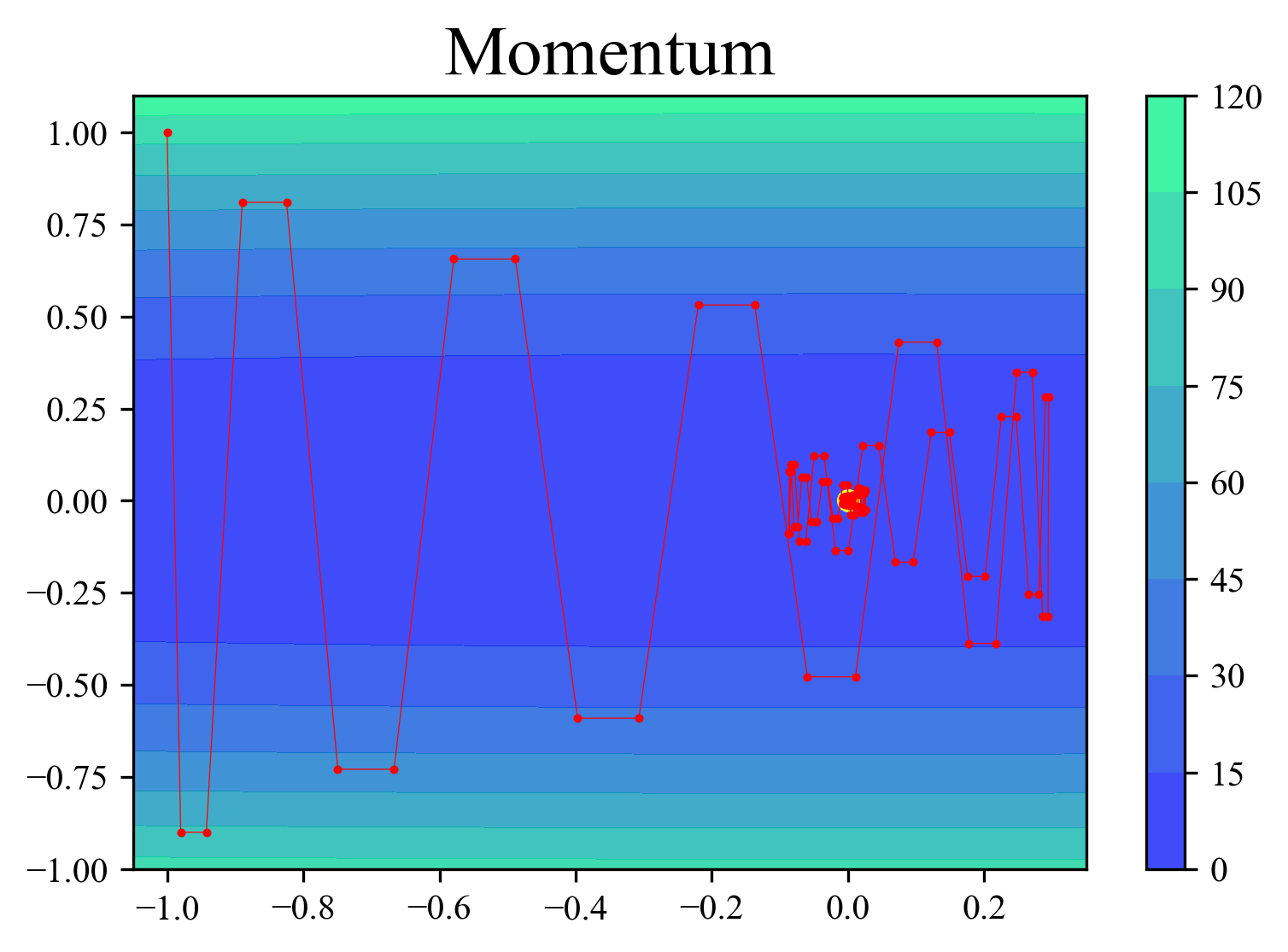}
    \end{minipage}
    \begin{minipage}{0.246\linewidth}
        \includegraphics[width=\linewidth]{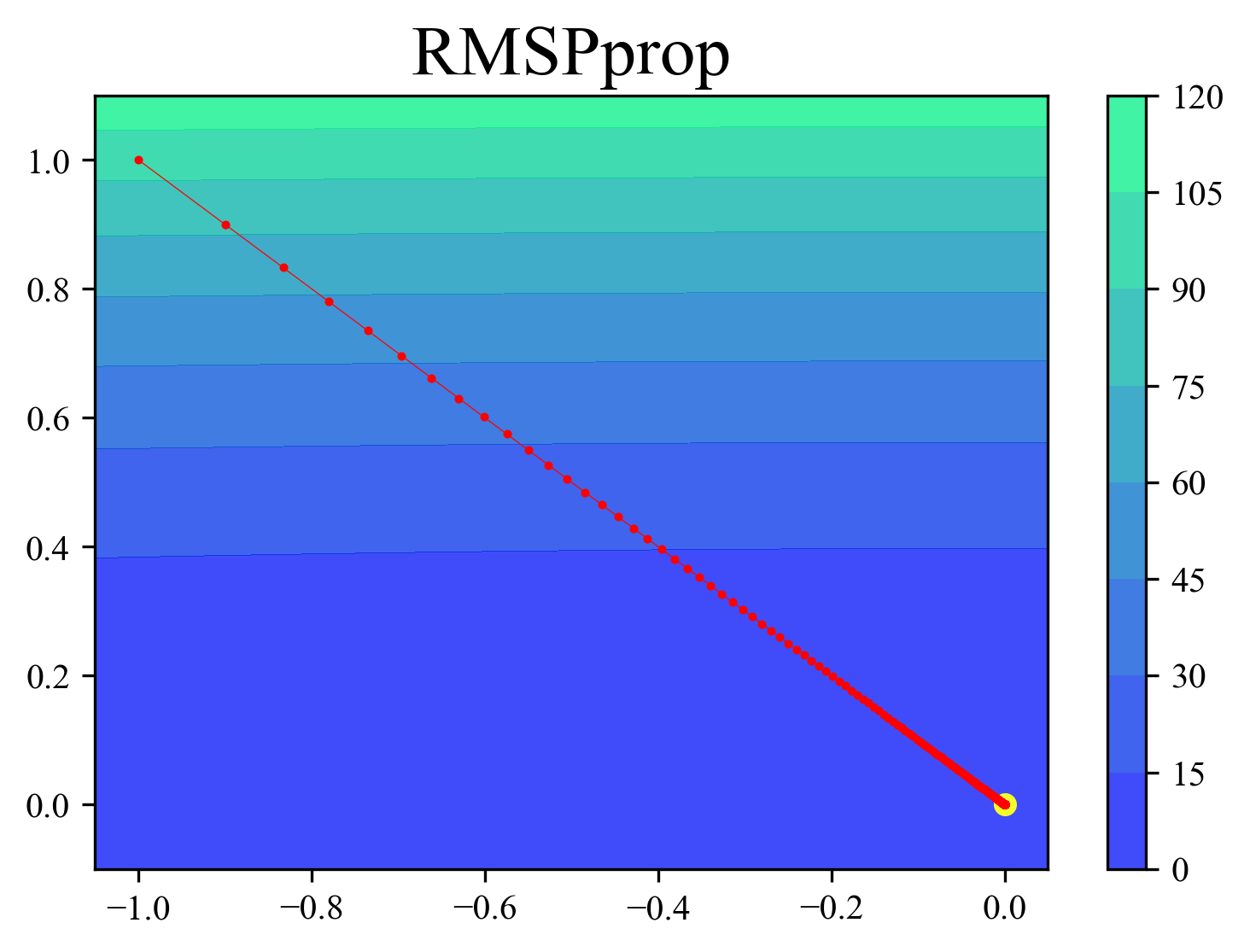}
    \end{minipage}
    \begin{minipage}{0.246\linewidth}
        \includegraphics[width=\linewidth]{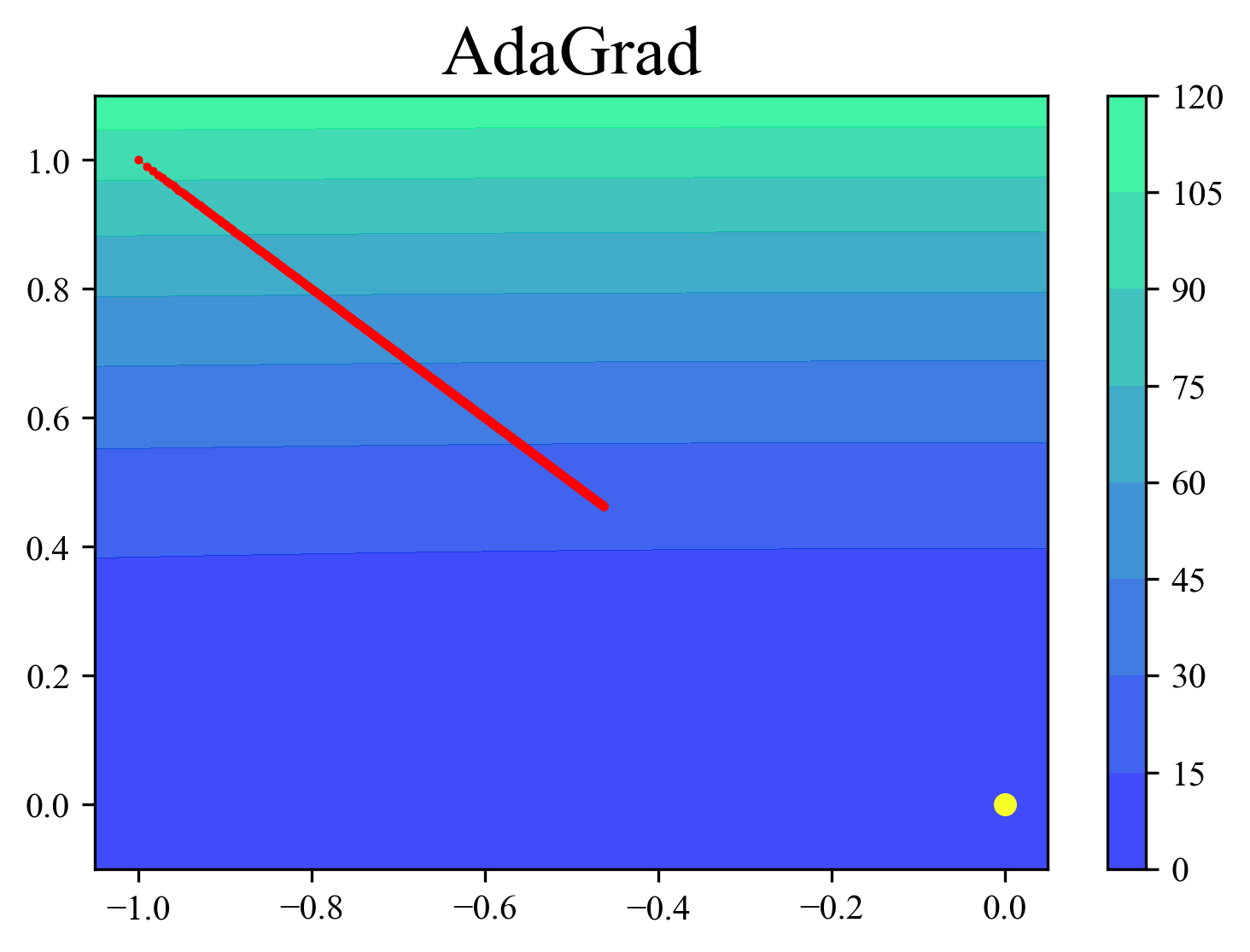}
    \end{minipage}
    \begin{minipage}{0.246\linewidth}
        \includegraphics[width=\linewidth]{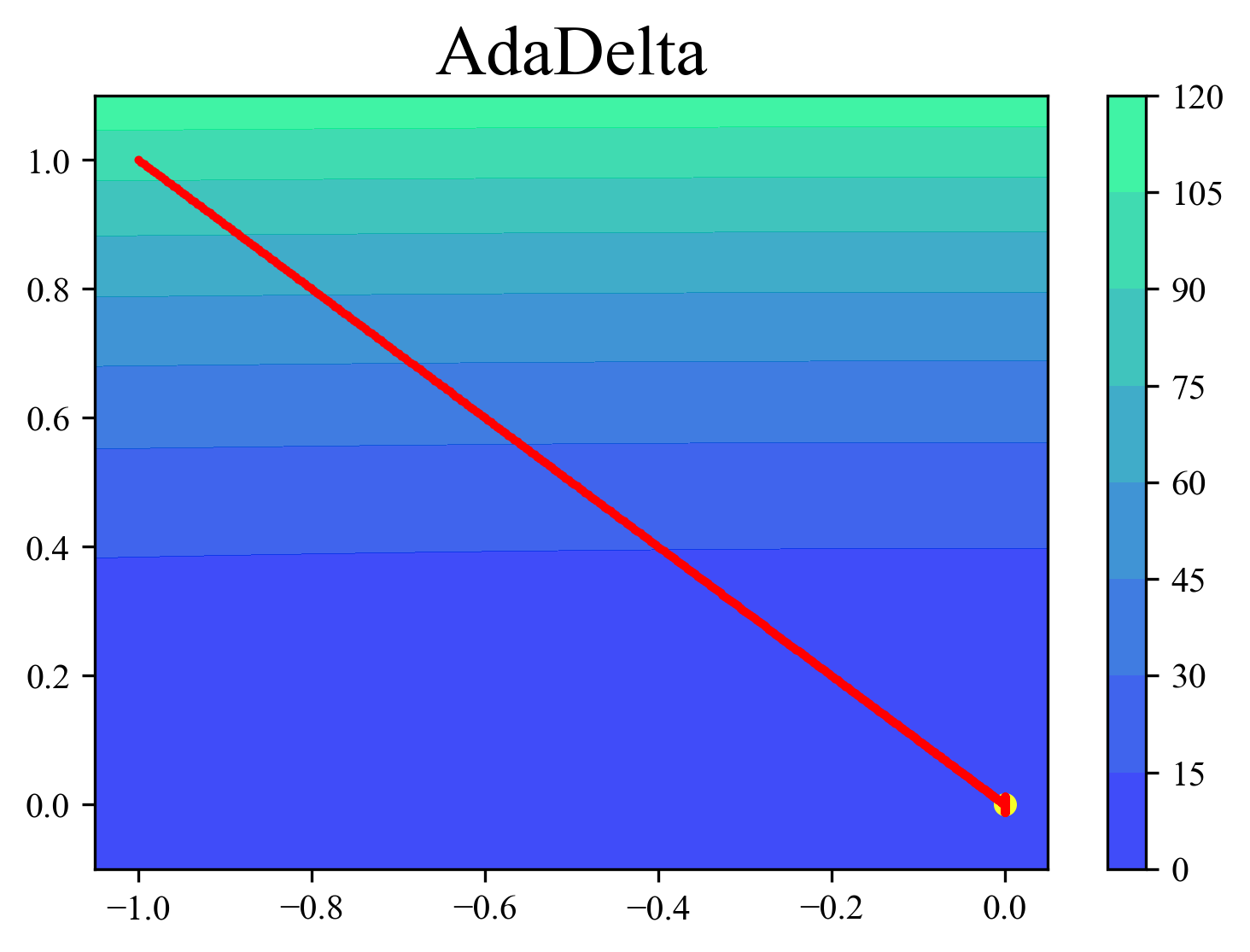}
    \end{minipage}
    \begin{minipage}{0.246\linewidth}
        \includegraphics[width=\linewidth]{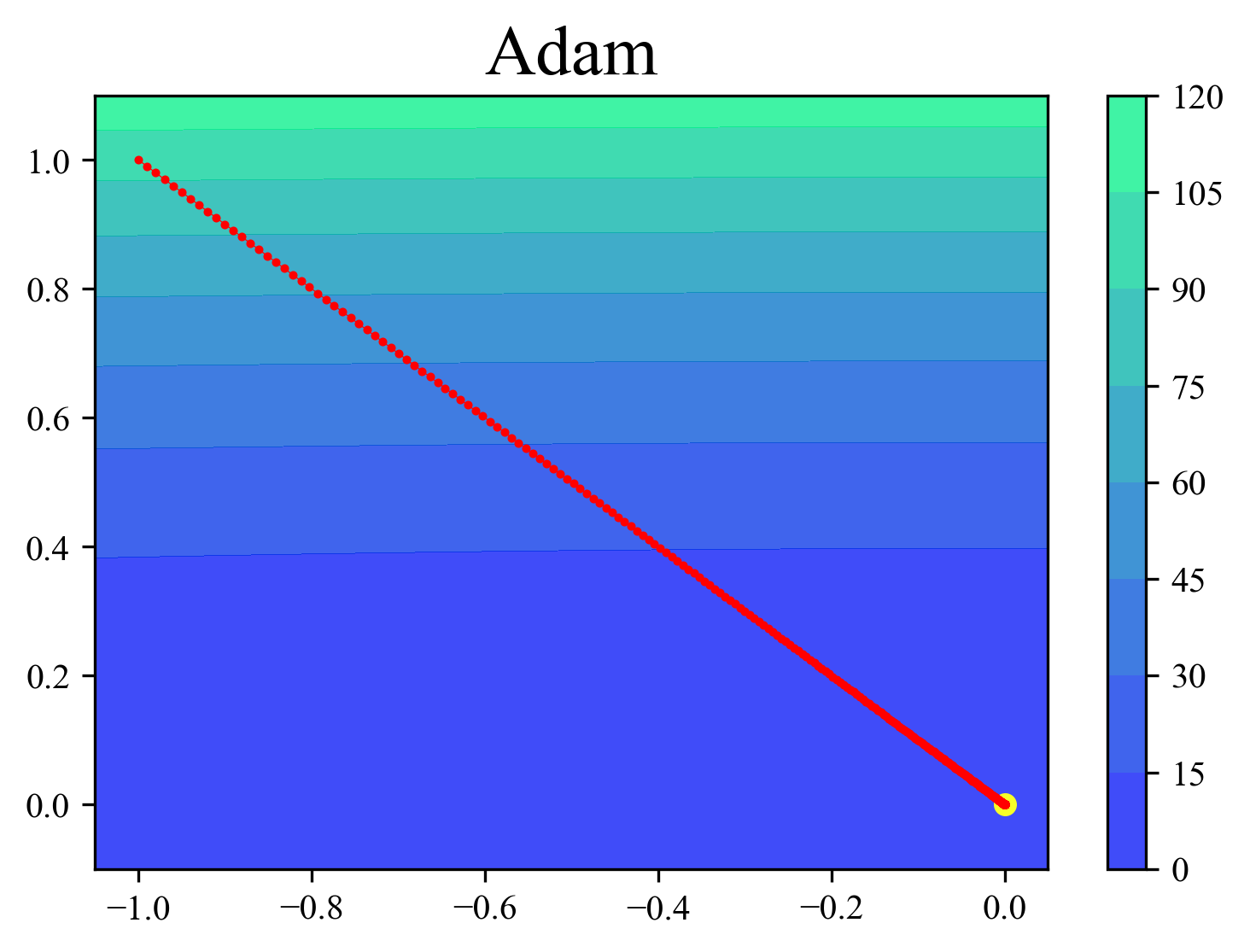}
    \end{minipage}
    \caption{Track visualization (start from (-1,1), $a = 1, b = 95$)}
    \label{fig: case_track_95}
\end{minipage}
\begin{minipage}{\linewidth}
    \centering
    \begin{minipage}{0.246\linewidth}
        \includegraphics[width=\linewidth]{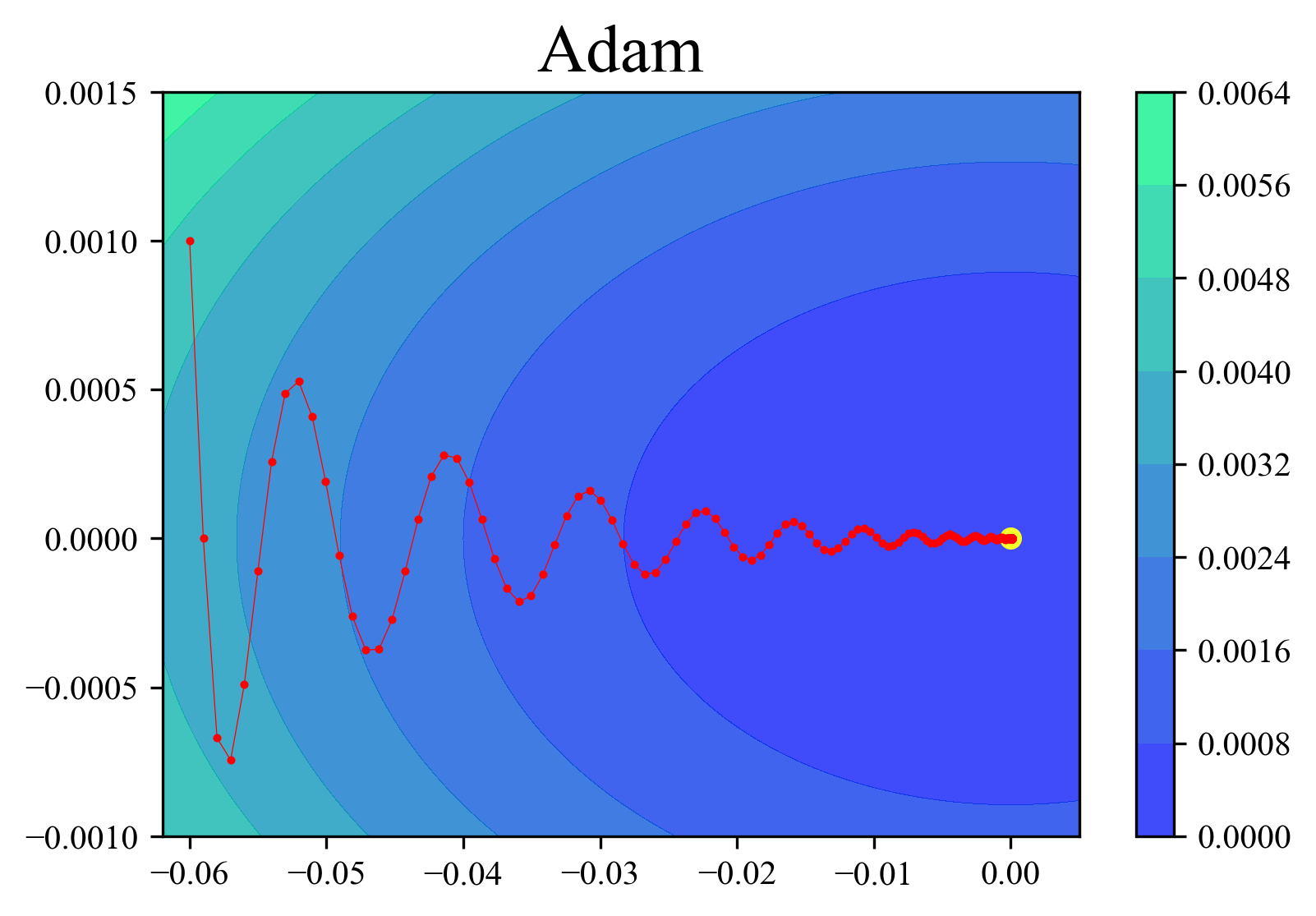}
    \end{minipage}
    \begin{minipage}{0.246\linewidth}
        \includegraphics[width=\linewidth]{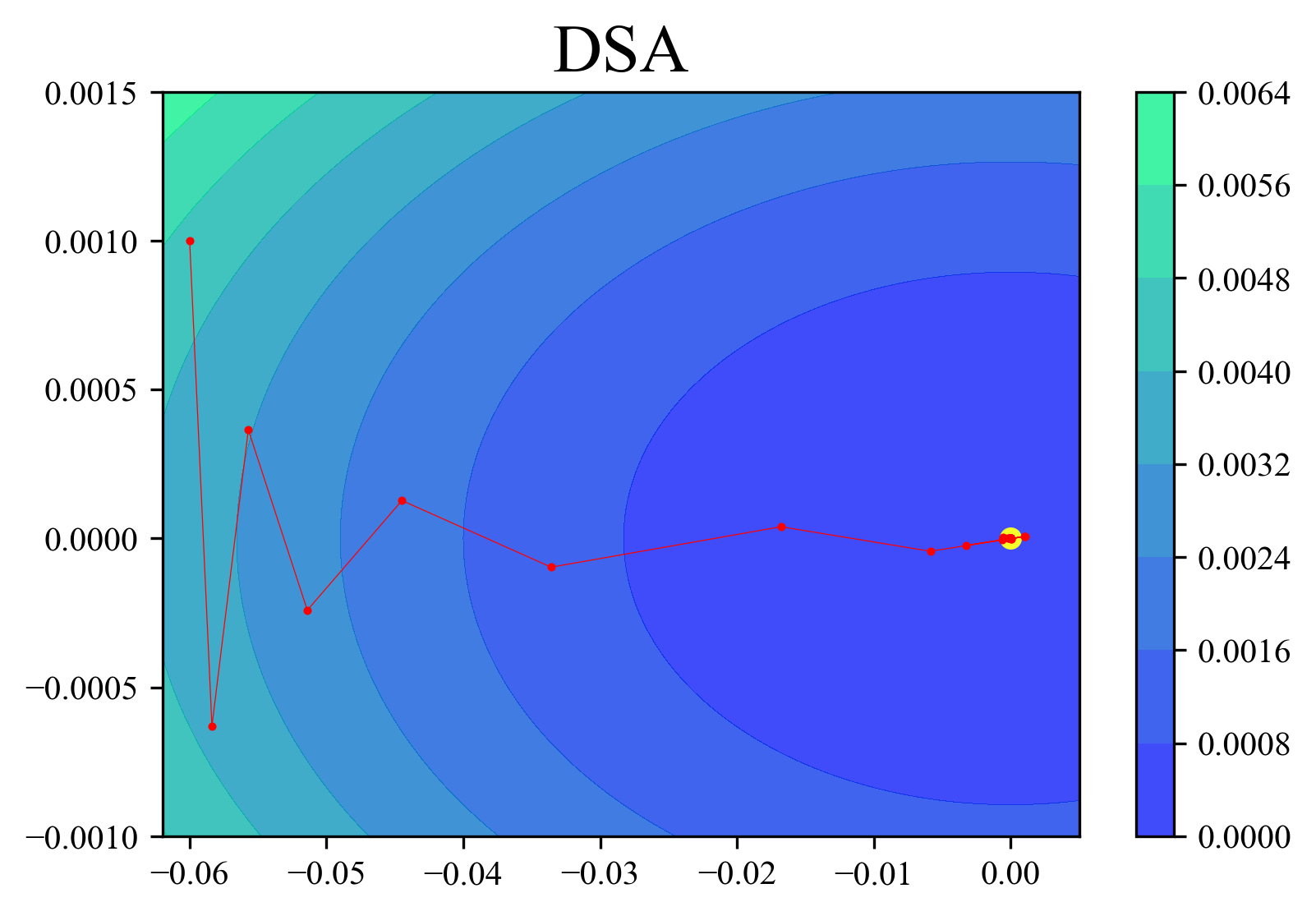}
    \end{minipage}
    \begin{minipage}{0.246\linewidth}
        \includegraphics[width=\linewidth]{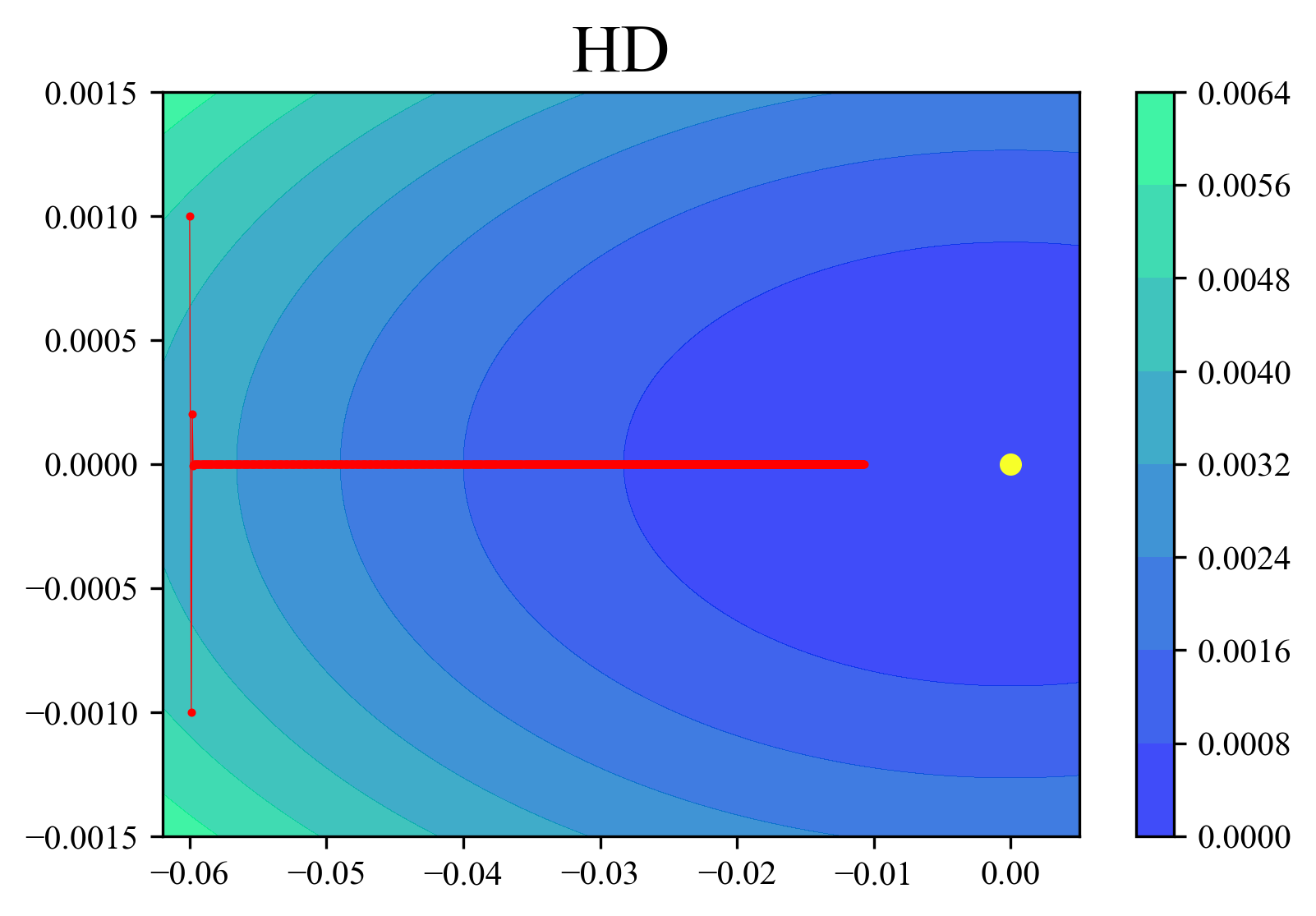}
    \end{minipage}
    \begin{minipage}{0.246\linewidth}
        \includegraphics[width=\linewidth]{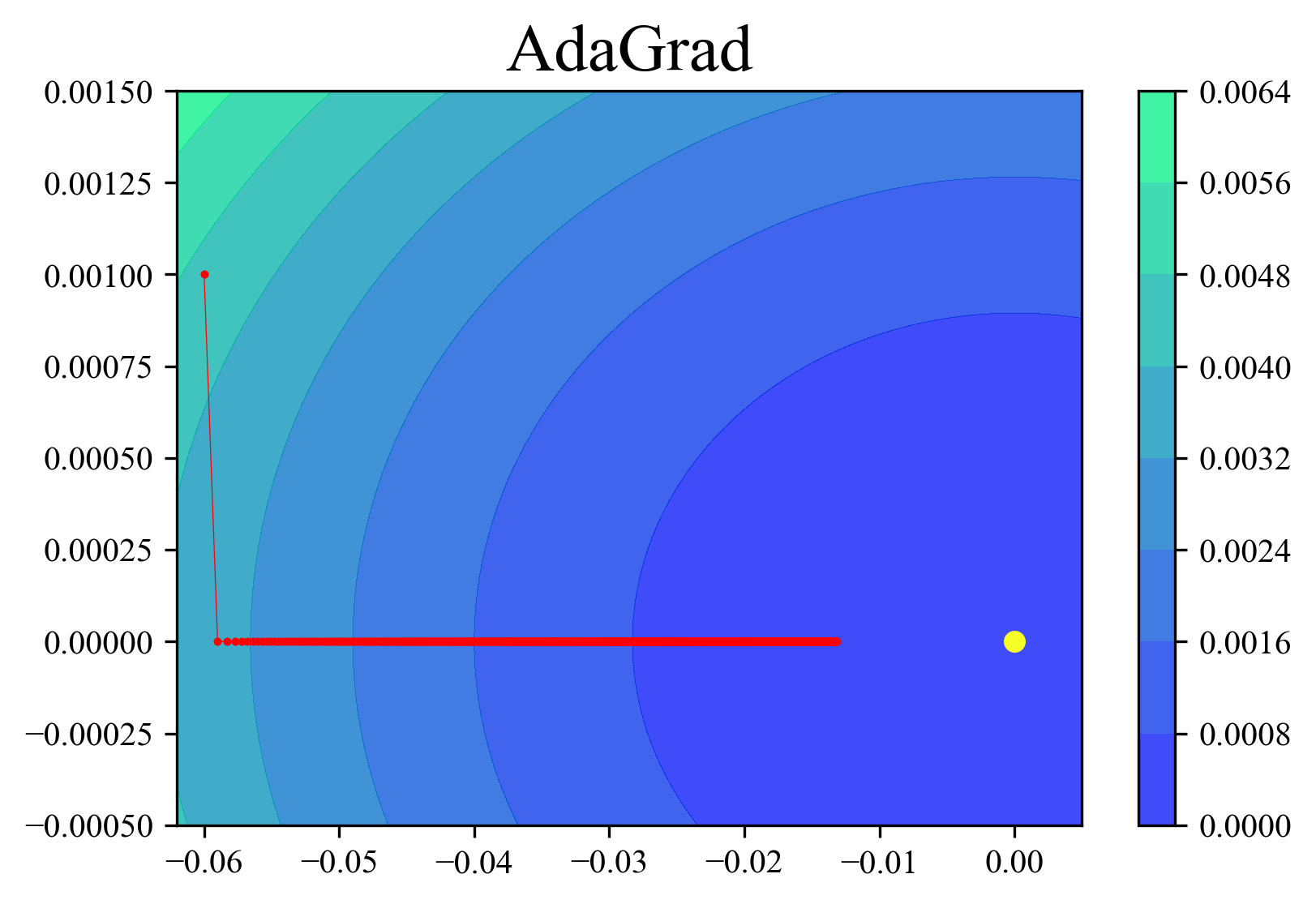}
    \end{minipage}
    \begin{minipage}{0.246\linewidth}
        \includegraphics[width=\linewidth]{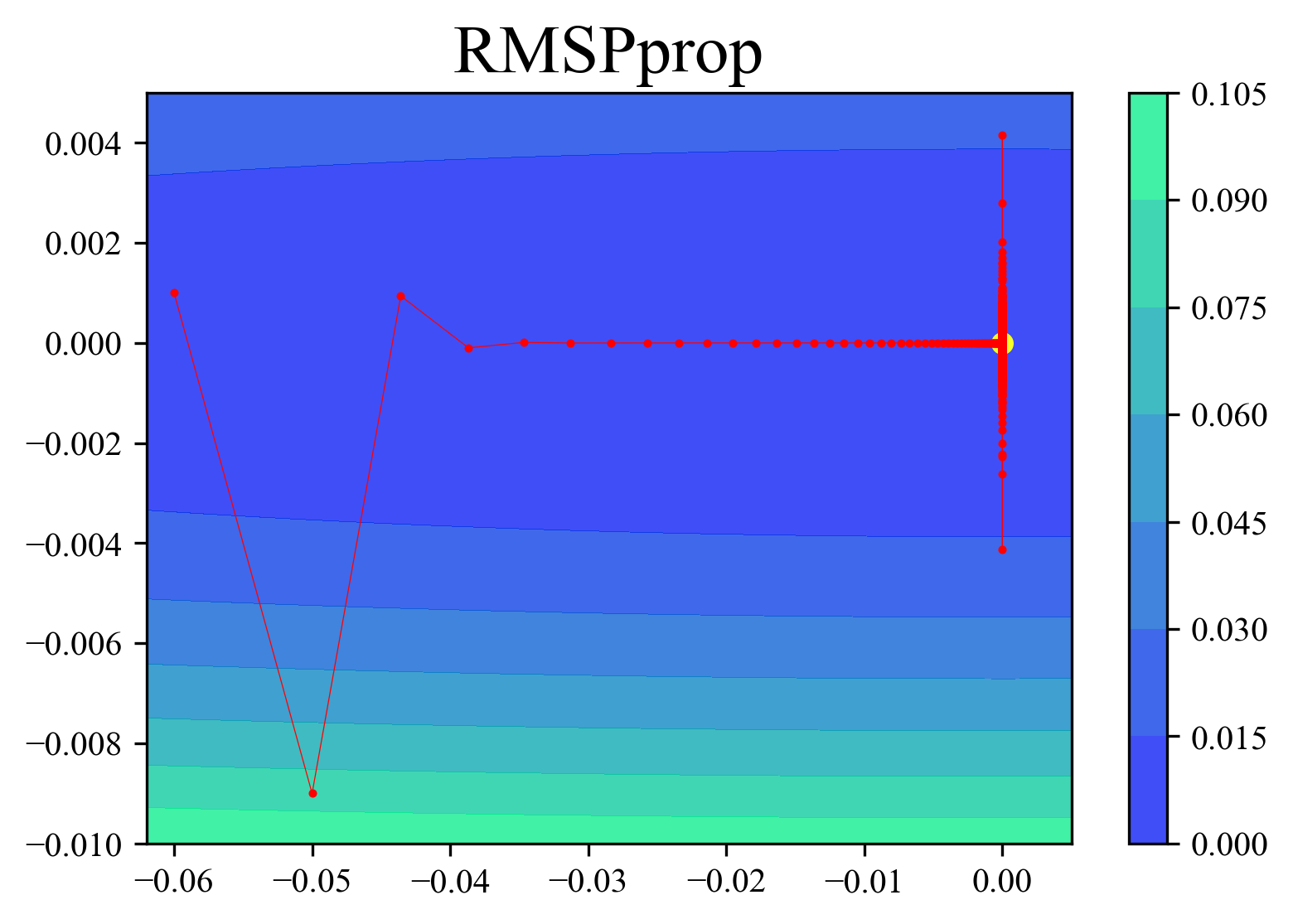}
    \end{minipage}
    \begin{minipage}{0.246\linewidth}
        \includegraphics[width=\linewidth]{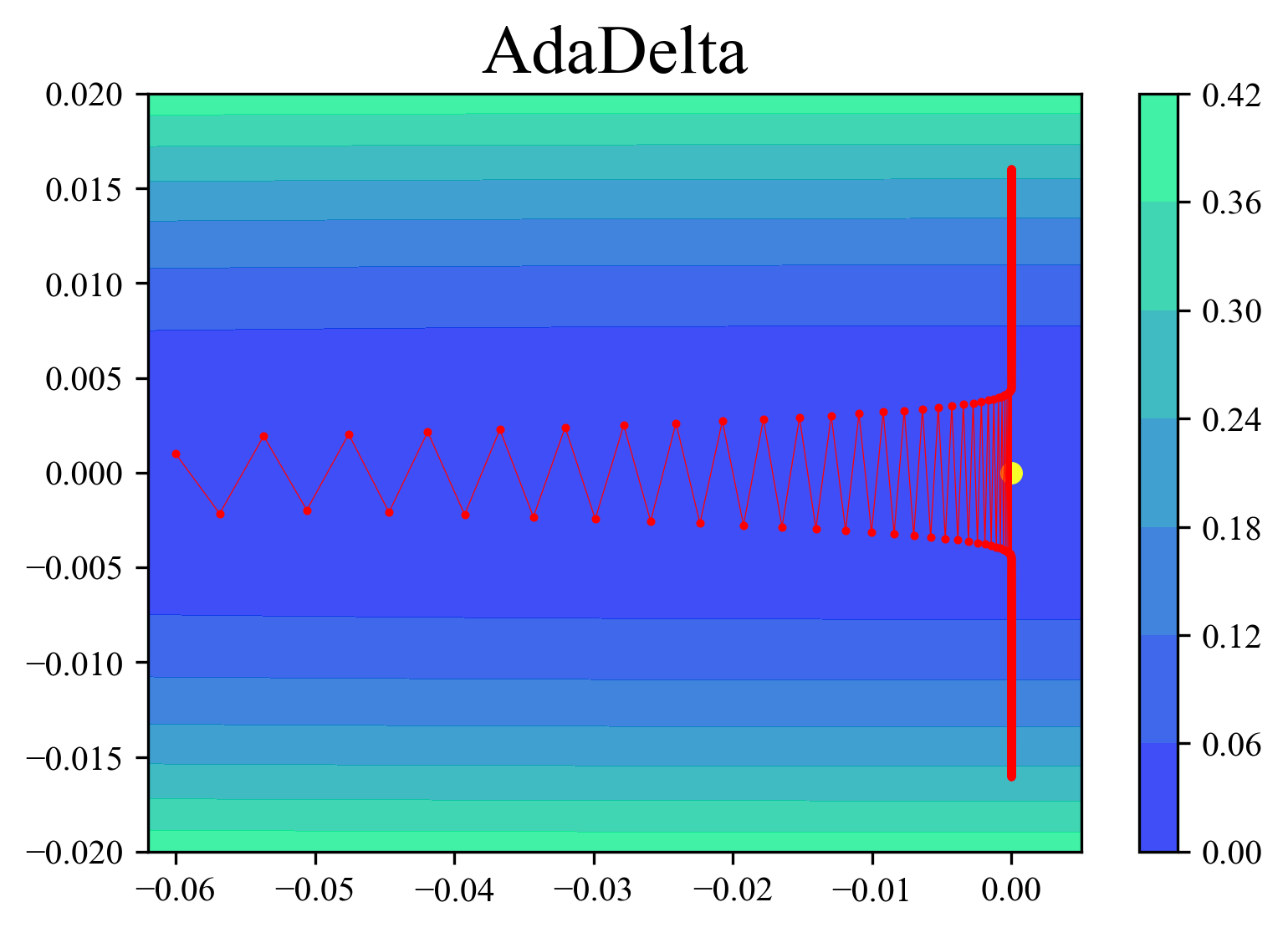}
    \end{minipage}
    \begin{minipage}{0.246\linewidth}
        \includegraphics[width=\linewidth]{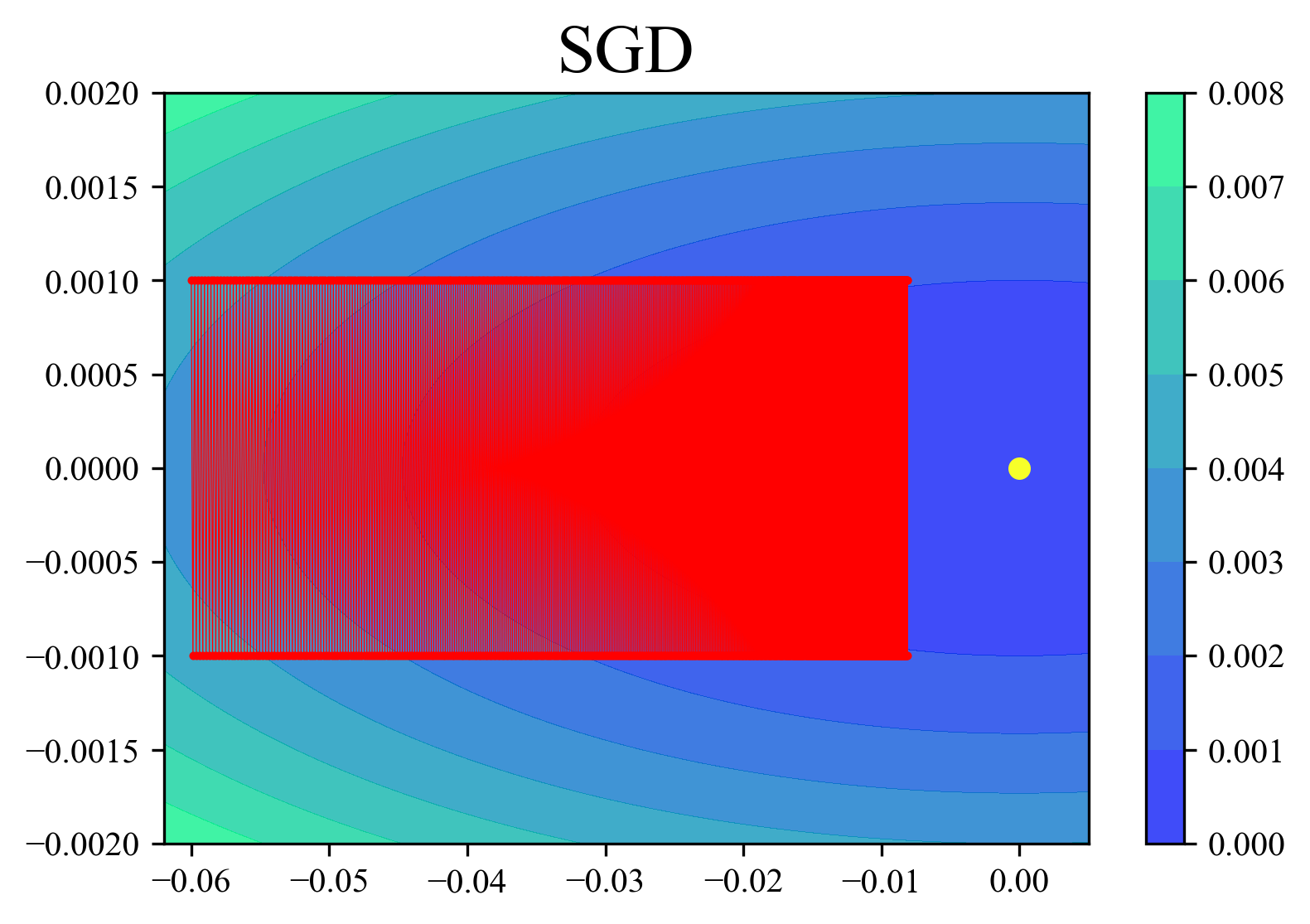}
    \end{minipage}
    \begin{minipage}{0.246\linewidth}
        \includegraphics[width=\linewidth]{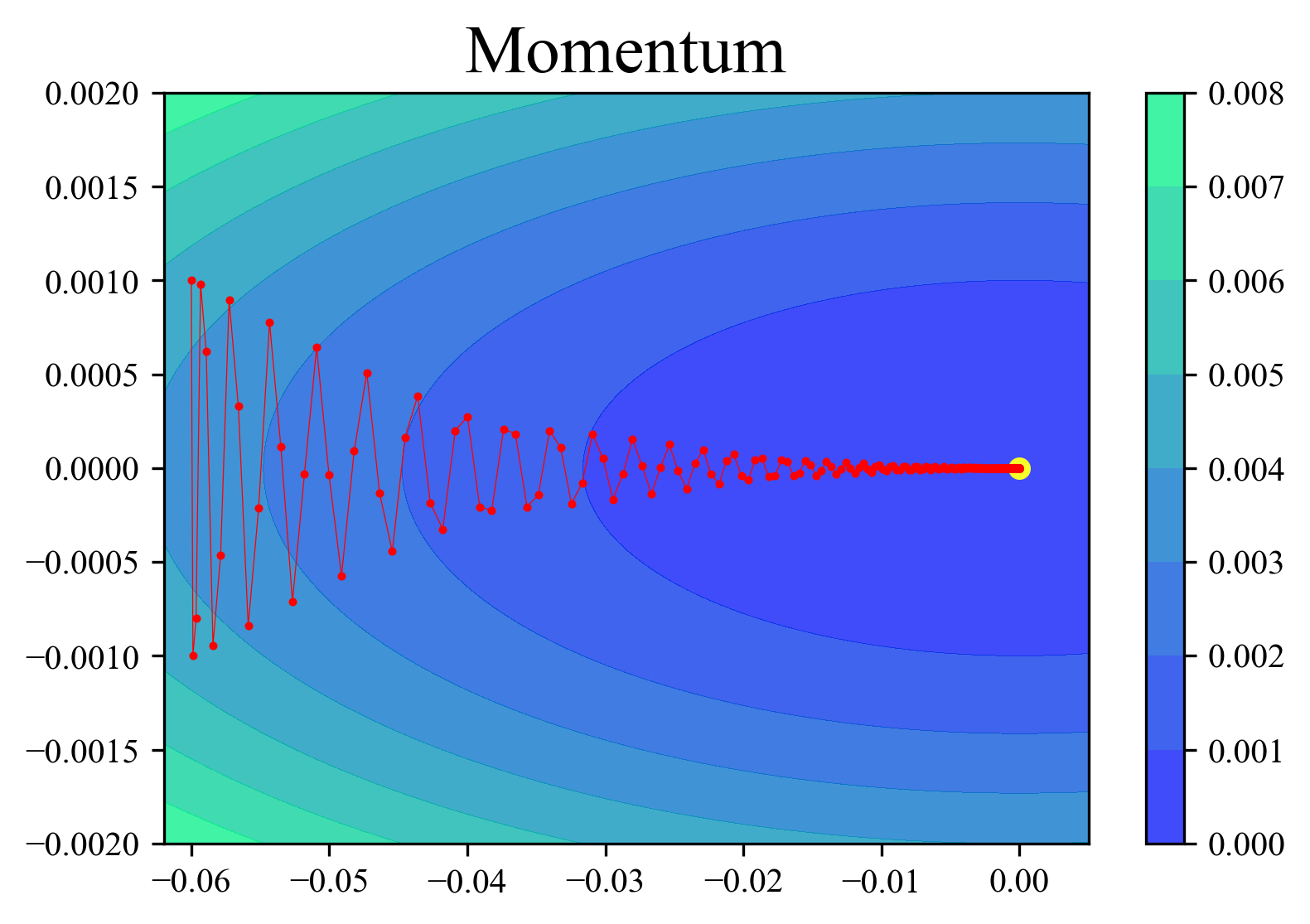}
    \end{minipage}
    \caption{Track visualization (start from (-0.06,0.001), $a = 1, b = 1000$)}
    \label{fig: case_track_1000}
\end{minipage}
\end{figure*}
\subsection{Experimental Results}
\label{sec: exp res}
In this section, we show the performances of DSA in different occasions. Experiments in this section consist of two groups totally.
Firstly, we apply DSA to fit convolution neural networks. ResNet is trained with all selected image datasets, while FMP and DNN are just evaluated on MNIST and SVHN because the performances are limited on the other two datasets. And the baselines used on FMP and DNN are ADAM family optimizers, because other optimizers would take ten thousands of iterations to train these two networks.
The second group is a series of experiments training MLP on small feature datasets.
\\\textbf{Experiments with Image Datasets}\\
Results on ResNet haven been collected in TABLE~\ref{tab: resnet mnist}, TABLE~\ref{tab: resnet svhn}, TABLE~\ref{tab: resnet cifar10} and TABLE~\ref{tab: resnet cifar100}. Results on DNN and FMP are shown in TABLE~\ref{tab: dnn mnist}, TABLE~\ref{tab: dnn svhn}, TABLE~\ref{tab: fmp mnist} and TABLE~\ref{tab: fmp svhn}. In tables, the baseline with $\star$ is the pretrained. DSA is trained for only 10 epochs after the pretraining, but it can get the peak of performance rapidly. That is, DSA can always enhance the pretrained model by $0.1\%\sim1\%$ on accuracy. Intrestingly, we find that the maximum of DSA's metric is usually smaller than that of pretrained baseline. For example, the precision's maximum of DSA on ResNet with SVHN is 97.70 and that of Momentum(pretrained baseline) is 97.80, i.e., DSA is smaller than Momentum. While the precision's minimum of DSA is 94.66 and that of Momentum is 94.07, i.e., DSA is larger than Momentum. And the accuracy of DSA is $0.3\%$ larger than that of Momentum. In fact, the trained model will favor the latest trained minibatch data more than others. So the neural network always do better on some part of a dataset. While in the formal training of DSA, which is in the form of batch training, DSA treated all training samples fairly. So the results of DSA is surely to be better totally. While such a training mode is only suitable for DSA, because only DSA can achieve fast adaptation for step size of model's parameters.
\\\textbf{Experiments with Feature Datasets}\\
In this section, we display the performance of DSA when processing MLP on feature datasets. As described in experiment settings, MLP is trained for 1000 epochs in a single session. While we do not show the result after all the epochs because nearly all the optimizer can get an accuracy of 100 percent on IRIS and AGAICUS. Instead, we will choose a distinguishable epoch for each dataset and capture the performance of each optimizer at that opint. Concretely, WINE, CAR, IRIS and AGAICUS are captured at 1000th epoch, 500th epoch, 30th epoch and 100th epoch, respectively.
We conclude the performance in TABLE~\ref{tab: mlp wine},~\ref{tab: mlp car},~\ref{tab: mlp iris} and~\ref{tab: mlp agaricus}. IRIS is the simplest of the four datasets and each optimizer can gain a perfect performance on it as long as training with enough iterations. So we abstract the information of first 30 from the total 1000 epochs to compare the convergence speed of optimizers. Obviously, RMSPprop, AdaGrad and DSA are the most remarkable by achieving an accuracy of 100 percent in only 30 epochs. Similar with AGAICUS, although there are thousands of training samples, it's just a \textit{yes or no} problem. So it's also a not complex task, where AdaGrad and DSA obtain full marks again with few epochs. Actually, DSA can accomplsh this within just 30 epochs and this is far more faster than any other optimizers. WINE is a little troublesome task for there are some classes with very little training samples. This is why there are still some classes with a RECALL of 0 even after all the 1000 epochs. Fortunately, AdaGrad, AdamW and DSA still finish the task effectively. As for CAR dataset, only DSA can do it all right within 500 epochs.
We also visualize the train loss in Fig.~\ref{fig: mlp_loss} and the validation accuracy in Fig.~\ref{fig: mlp_accu}. With help of DSA, tain loss always decreases to 0 rapidly surrounding all the other optimizers in the figures, especially on CAR and AGARICUS. There is no doubt that DSA is the most sensitive and efficient algorithm compared with these stat-of-arts.
\subsection{Case Study}
\label{sec: case study}
In this section, we conduct two case studies with DSA and baselines. The first is a simple regression problem to calculate the sum of four real numbers. The second is a convex optimization, i.e., minimization problem.
\\\textbf{Regression}\\
As the first case, we solve a simple regression problem calculating the sum of four real numbers with target of $\min_{\boldsymbol{w}}{||\boldsymbol{w}\cdot \boldsymbol{x}^\mathrm{T} - y||}$, where $\boldsymbol{w} = [w_1, w_2, w_3, w_4] \in \mathbb{R}^4$. Obviously, the optimal of each $w_i$ is 1. We set this case to observe the optimization detail of each optimizer from a more intuitive perspective of tracking the value of parameters. In the regression, each $w_i$ is initialized in a kaiming uniform distribution with $a = \sqrt{5}$~\cite{kaiming}. We randomly generate 10,000 pieces of $(x_1, x_2, x_3, x_4)$ from a uniform distribution [0, 1) and use the sum of $x_i$ as labels. In this case, we fine-tune the learning rate for each optimizer. Finally, learning rate of Adam family and AdaGrad is set to 0.1. DSA's initial learning rate is 0.05 and the step size of $\alpha$ is 0.5. The others stay the same.

Regression loss and track of parameters are visualized in Figure~\ref{fig: case_sum}. From the view of regression loss, DSA reaches $10^{-13}$ within about 100 epochs and this is at least twice as rapid as the other optimizers. From the track of parameters, we can see that every optimizer would have a fluctuation around 1. While DSA stabilized first through learning rate adaptation. HD also achieves stability with more iterations and adaptation, i.e., DSA is more sensitive.
\\\textbf{Convex Optimization}\\
The second case is a convex optimization with target of $\min_{w_1,w_2}{a*w_1^2 + b*w_2^2}$. We set this case to illustrate learning rate adaptation of DSA more intuitively. Obviously, the optimal is $w_1 = w_2 = 0$. In the experiment, we set different initial position of $(w_1, w_2)$ and different $(a,b)$ combination to observe each optimizer's performance. Practice consists of two groups. We set a trap for SGD in the first group, where SGD will fluctuate around extreme points by setting $(a,b)=(1,95)$. And the initial position for the first group is $(w_1^{(0)},w_2^{(0)})=(-1,1)$. Similarly, we set a trap for ADAM in the second group, where $(w_1^{(0)},w_2^{(0)})=(-1,1)$ and $(a,b)=(1,1000)$.
When conducting the first group, we set all the optimizer's learning rate or initial learning rate as 0.01 for fair. And to ensure the convergence under HD, we set its step size of learning rate as $10^{-7}$. As for the second group, initial learning rate is 0.001 uniformly and step size of HD is $10^{-4}$. Single test contains 1000 iterations.

Tracks of $(w_1, w_2)$ are visualized in Fig.~\ref{fig: case_track_95} and Fig.~\ref{fig: case_track_1000}. For the first group, we can see that $w_2$ of SGD get into the trap, i.e., fluctuating around 0 wildly. HD and Monmentum also stuck into the trap. But HD could get rid of it through its learning rate adaptation. The other optimizers all avoid the trap, especially the DSA. DSA speeds up rapidly in the beginning and slow down timely in the end. As a result, DSA could reach the optimal with extremely few iterations.
As for the second group, we set a so troublesome initial position on an extremely steep slope that nearly all the optimizers begin with fluctuations, more or less. It seems that only AdaGrad keeps away from fluctuations. Actually the learning rate of AdaGrad decreases rapidly because of the large gradient of $w_2$ in the first iteration. This makes AdaGrad survived from the fluctuations, while this also makes the learning rate decreases too much so that AdaGrad could not reach the optimal within limited iterations. Surprisingly, although DSA also get into the trap in the beginning, it quickly takes adaptation and rush towards the optimal within only 10 iterations, which is far beyond the capabilities of others.
\begin{figure}[h]
    \centering
    \includegraphics[width=\linewidth]{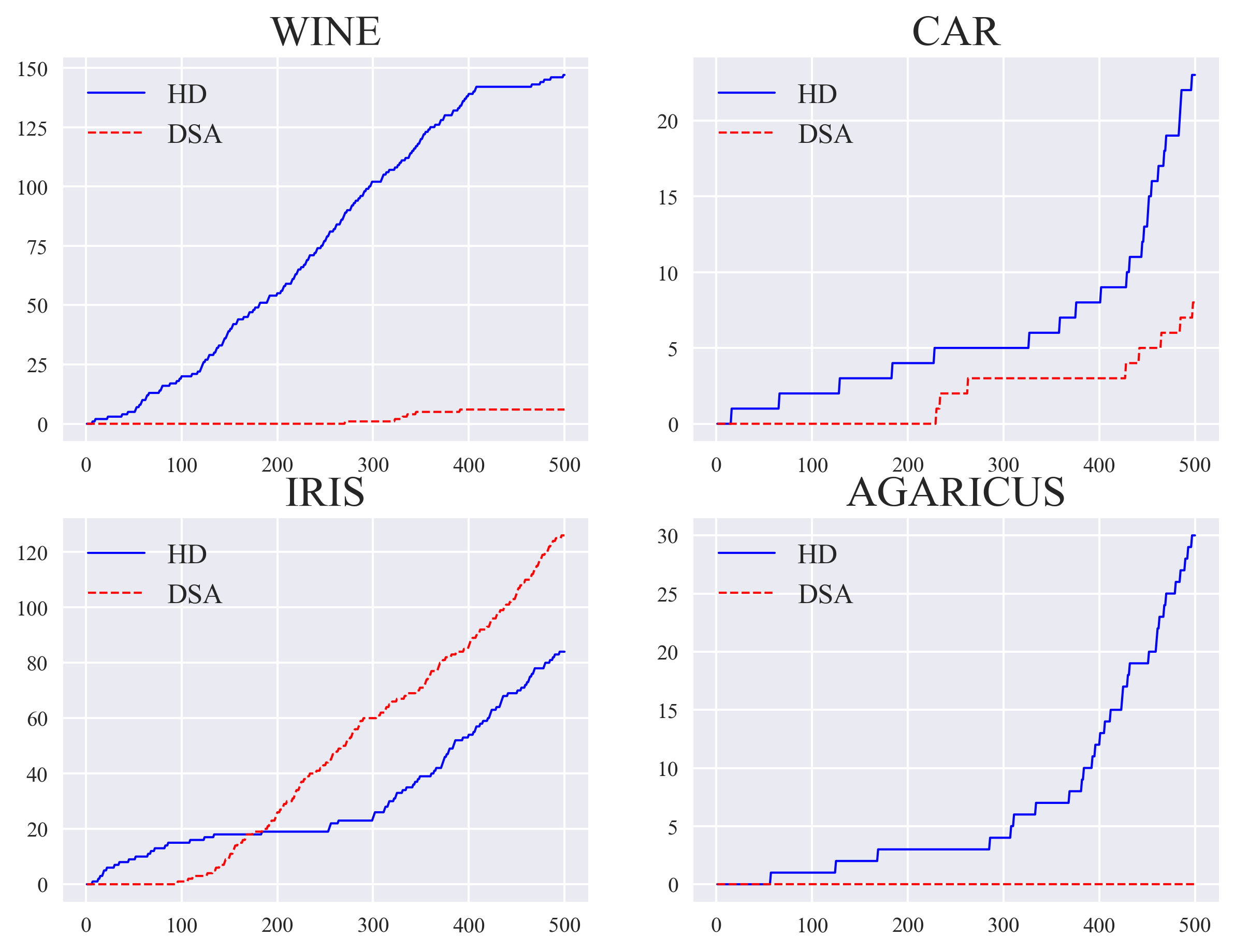}
    \caption{Miss frequency of HD and DSA.}
    \label{fig: conflict}
\end{figure}
\subsection{Ablation Experiment and Sensitivity Analysis}
\label{sec: sensitivity}
In this section, we take ablation experiment to evaluate the effect of techniques mentioned in Section~\ref{sec: trick}, including \textit{parameter specific} and \textit{step size equal learning rate}. And we also support evidence that \textit{detection} technique proposed in Section~\ref{sec: detect} is a better policy than adaptation of HD. Technique of \textit{internal structure} is not analysed here because it's necessary to ensure learning rate is positive. Additionally, sensitivity analysis for DSA's hyper-parameters is conducted then.
\\\textbf{Ablation Experiment}\\
Experiments are arranged as two groups. The first for \textit{detection} technique and the second for \textit{parameter specific} and \textit{step size equal learning rate}.

Firstly, we show that \textit{detection} technique of DSA can decrease miss rate greatly compared to HD. Miss frequency is computed in real time along with iterations as Fig.~\ref{fig: conflict}. DSA makes no mistakes in the beginning on each dataset. In the later of a training, miss rate of DSA increases because it steps around the extreme point, and so is HD.

Secondly, we show that \textit{parameter specific} and \textit{step size equal learning rate} are both important techniques for DSA. DSA will be tested without these techniques. We denote DSA without \textit{parameter specific} as DSA$\dagger$, and DSA$\dotplus$ for DSA without \textit{step size equal learning rate}.

\begin{figure}[h]
\begin{minipage}{0.5\linewidth}
    \centerline{DSA$\dagger$}
    \includegraphics[width=\linewidth]{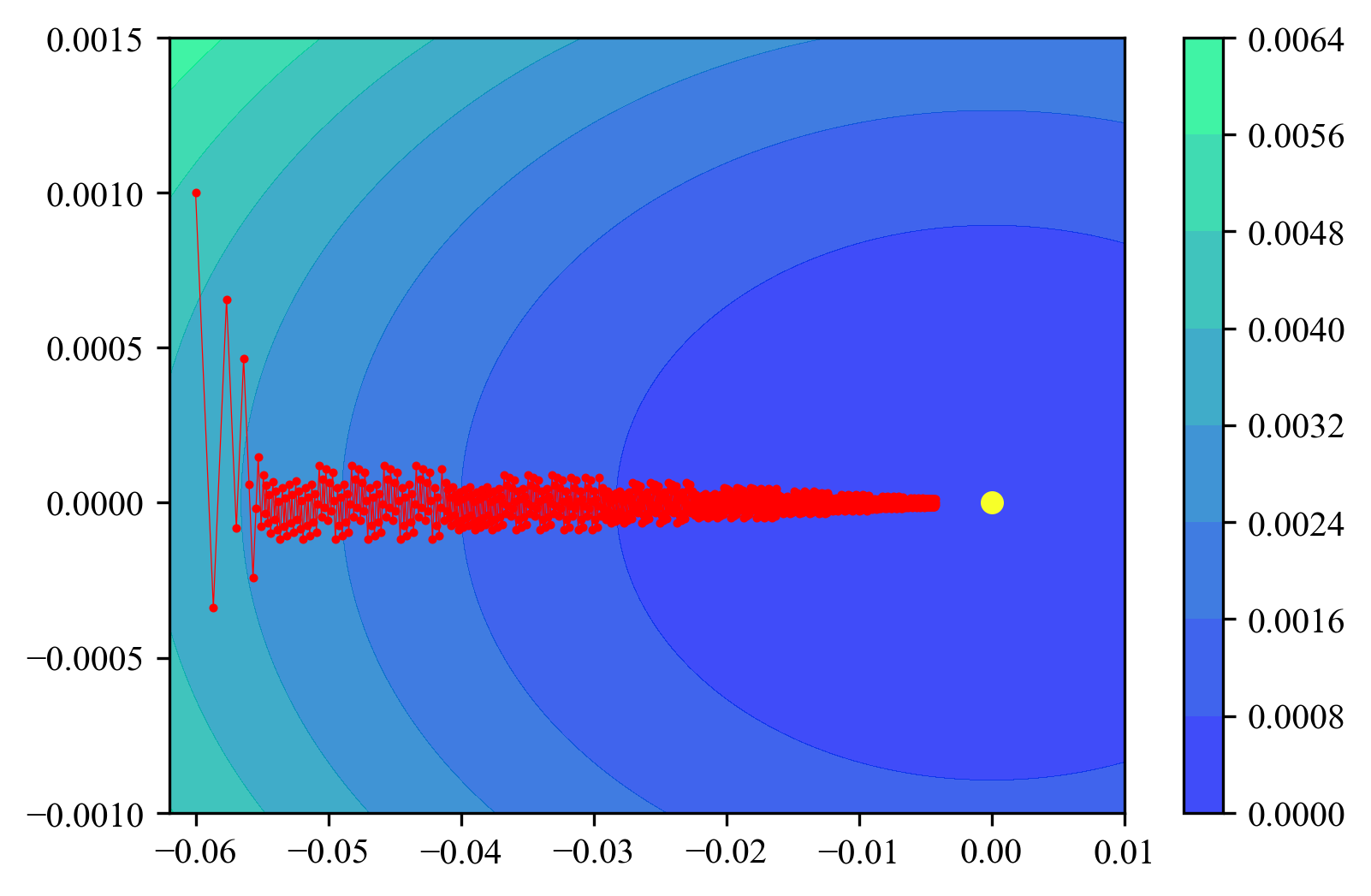}
\end{minipage}
\begin{minipage}{0.5\linewidth}
    \centerline{DSA$\dotplus$}
    \includegraphics[width=\linewidth]{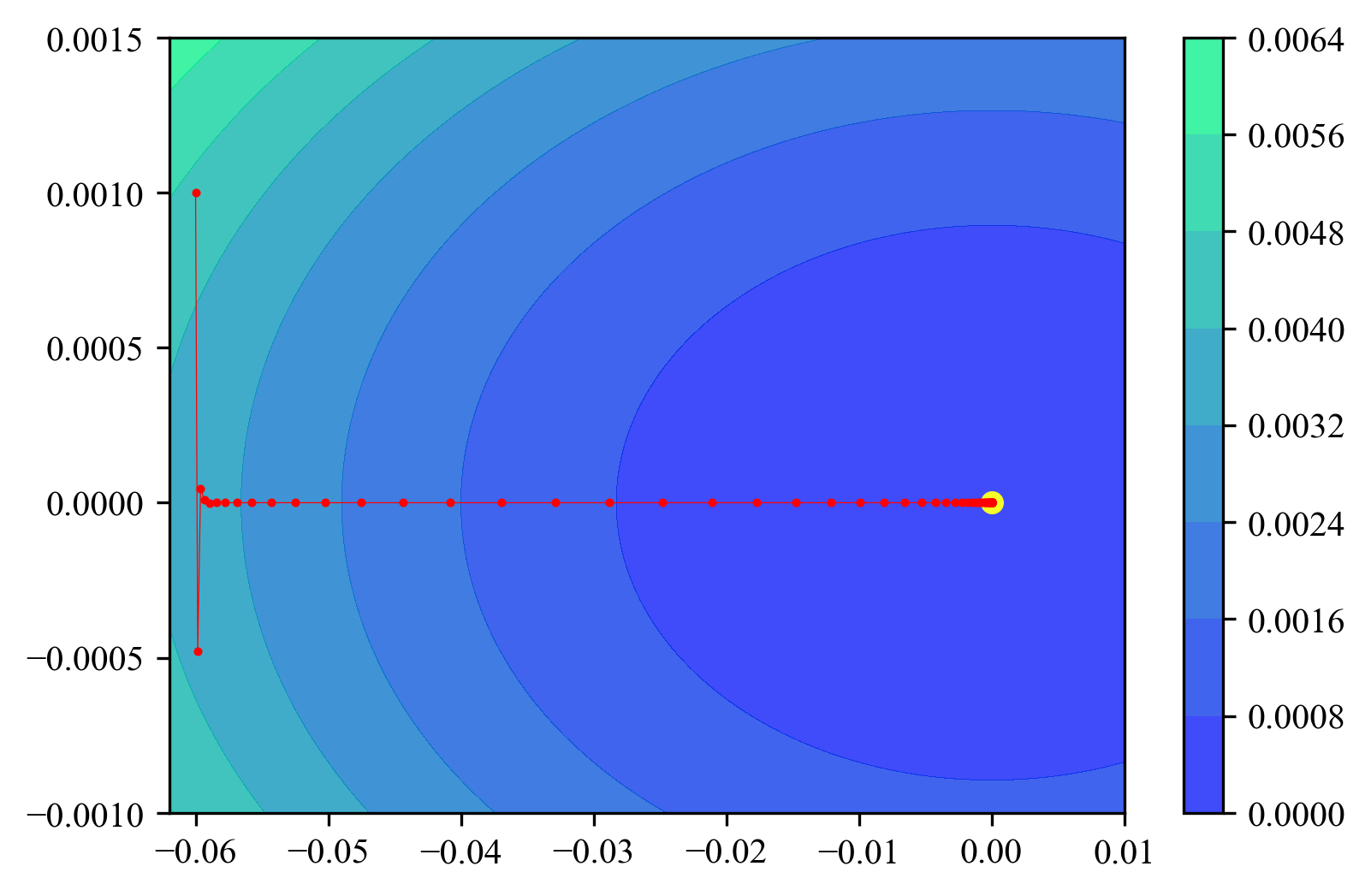}
\end{minipage}
\caption{Miss frequency of HD and DSA.}
\label{fig: dsa_dagger_dotplus}
\end{figure}
After retest the optimization task in case study with DSA$\dagger$ and DSA$\dotplus$, we draw the tracks of each in Fig.~\ref{fig: dsa_dagger_dotplus}. In the track of DSA$\dagger$, we find that learning rate is very small so that the convergence is very slow. $w_1$ needs a large learning rate and $w_2$ needs a small learning rate, while learning rate is globally unique and small because of $w_2$. So the steps of $w_1$ slow down passively. As for DSA$\dotplus$, the parameter is updated by basic gradient descent and we can get that it's not faster than DSA obviously because the step size of each step is smaller than that of DSA.

\begin{table}[h]
\caption{Performance(accuracy) of DSA$\dagger$ and DSA$\dotplus$}
\label{tab: dsa_dagger_dotplus}
\centering
\begin{tabular}{lrrrr}
        & WINE & CAR & MNIST & SVHN \\
    \toprule
    Adam & 97.22 & 99.42 & 99.26 & 96.30 \\
    SGD & 44.44 & 69.36 & 99.14 & 96.23 \\
    DSA$\dagger$ & 81.03 & 87.28 & 50.00$\sim$100.0 & 62.50$\sim$94.74 \\
    DSA$\dotplus$ & 87.93 & 69.36 & 100.0$\sim$100.0 & 100.0$\sim$100.0 \\
    DSA & 100.0 & 100.0 & 96.88$\sim$100.0 & 94.74$\sim$100.0 \\
\end{tabular}
\end{table}
We evaluate DSA$\dagger$ and DSA$\dotplus$ on WINE, CAR, MNIST and SVHN, compared with DSA and some other baselines as shown in TABLE~\ref{tab: dsa_dagger_dotplus}. The number of iterations on each dataset is 1000, 500, 200 and 200, respectively. The initial learning rates of DSA$\dagger$ and DSA$\dotplus$ are both 0.001 and the step size of $\alpha$ is 0.1 when dealing with feature datasets. And the initial learning rate will be $10^{-6}$ when applied on the pretrained model of large image datasets.
\begin{figure*}[h]
\begin{minipage}{\linewidth}
    \centering
    \begin{minipage}{0.33\linewidth}
        \centerline{$\gamma\sigma(\alpha) = 0.0001, \beta = 0.3$}
        \includegraphics[width=\linewidth]{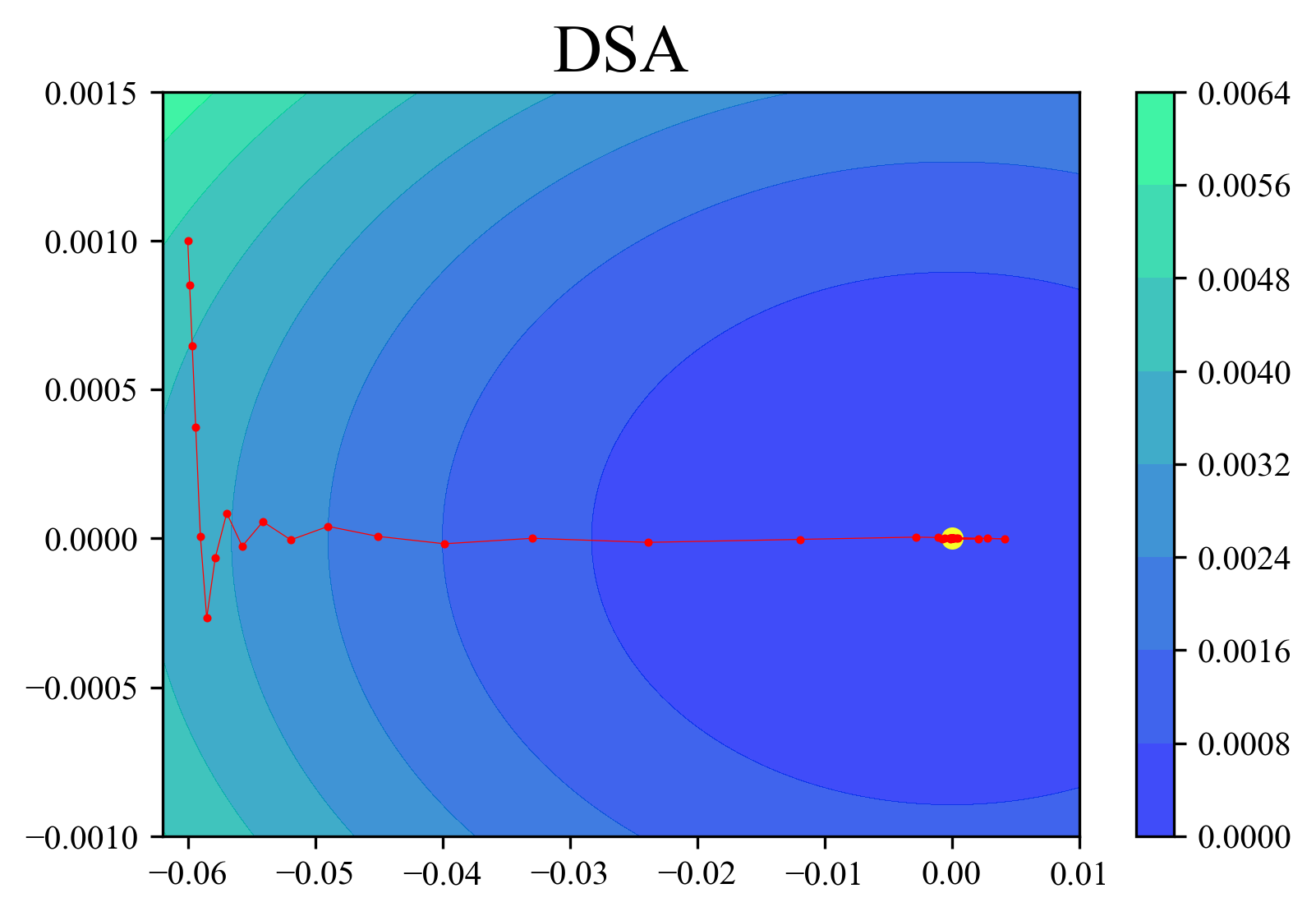}
    \end{minipage}
    \begin{minipage}{0.33\linewidth}
        \centerline{$\gamma\sigma(\alpha) = 0.001, \beta = 0.3$}
        \includegraphics[width=\linewidth]{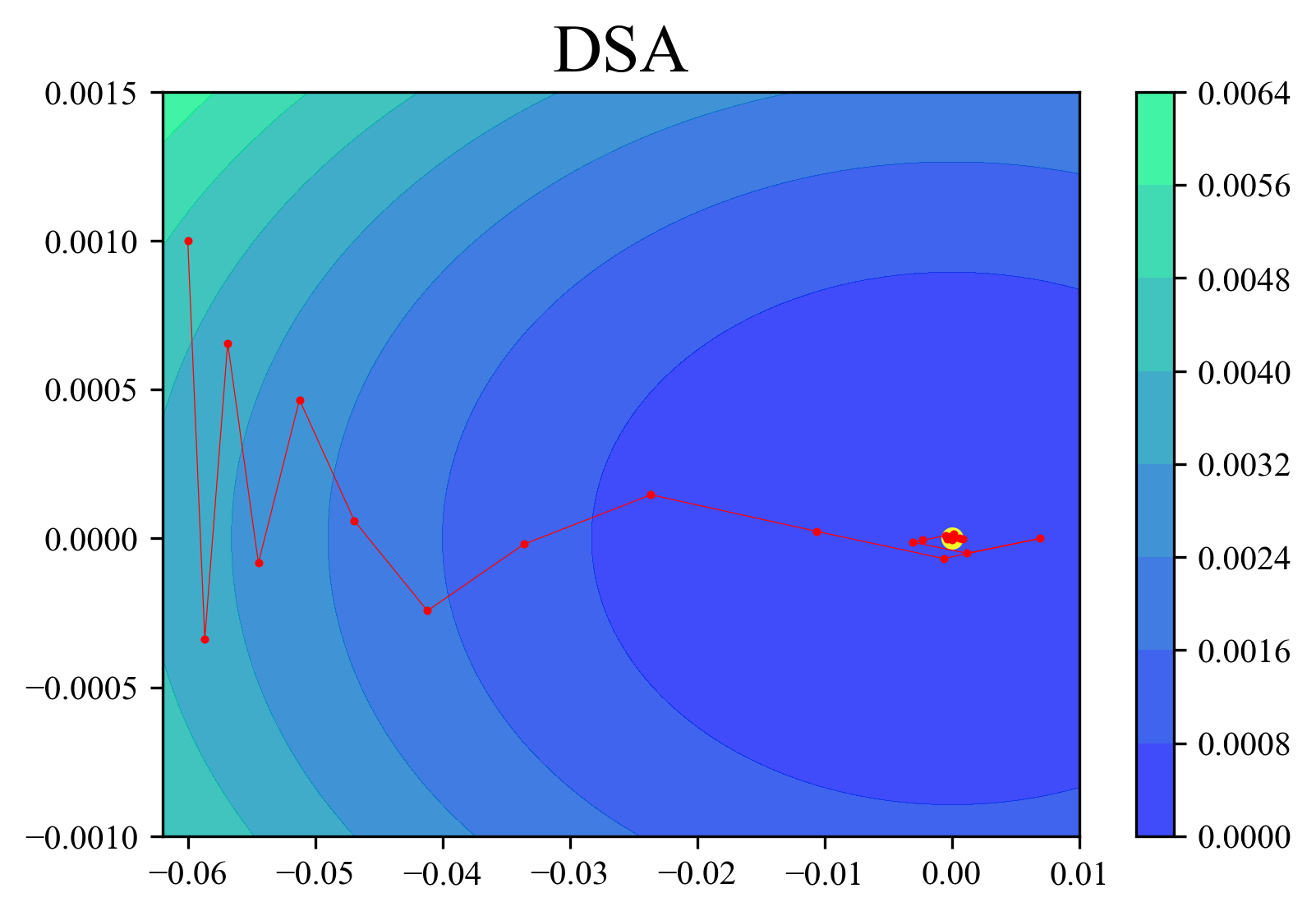}
    \end{minipage}
    \begin{minipage}{0.33\linewidth}
        \centerline{$\gamma\sigma(\alpha) = 0.01, \beta = 0.3$}
        \includegraphics[width=\linewidth]{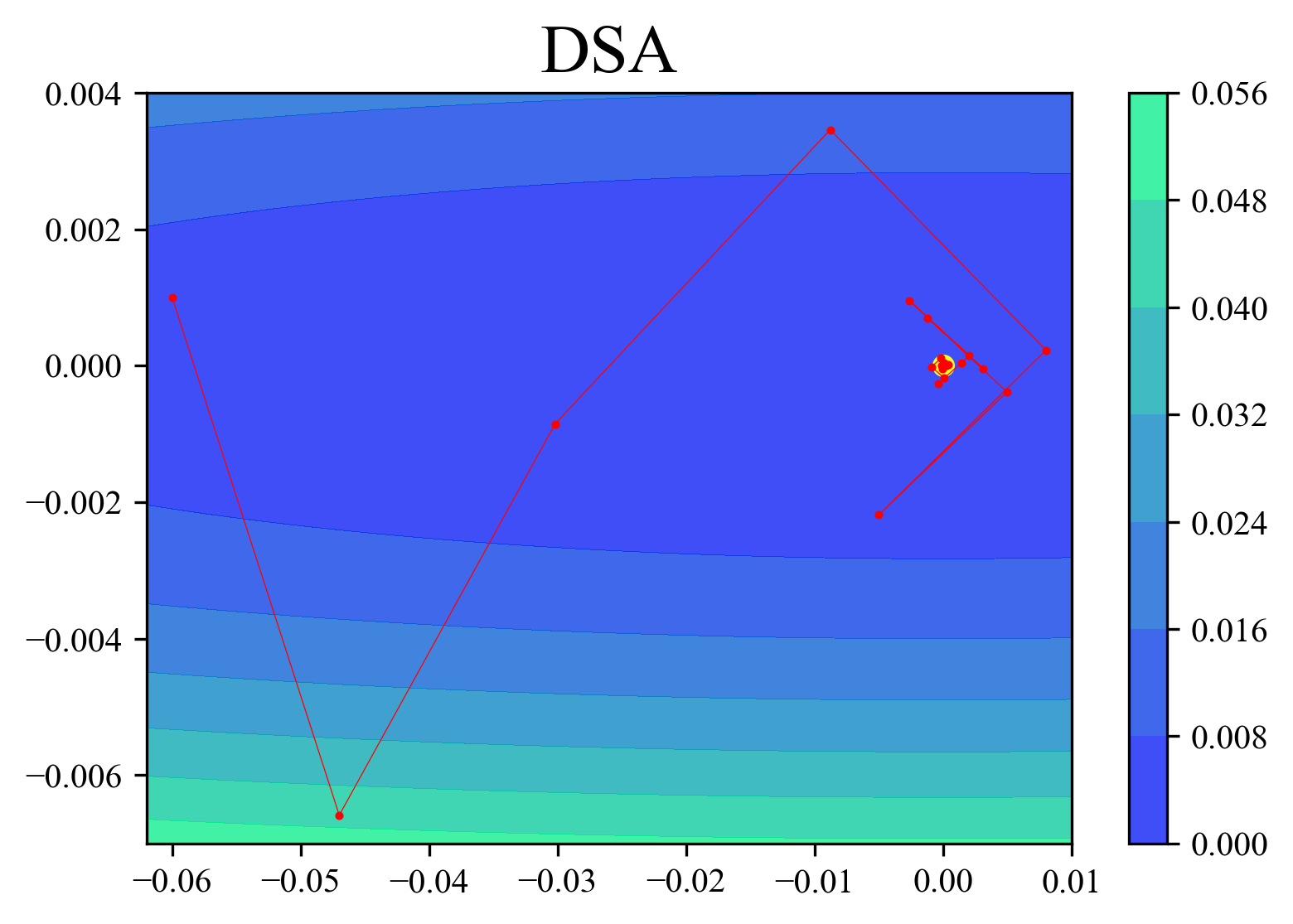}
    \end{minipage}
    \caption{Sensitivity of $\alpha$}
    \label{fig: sens_alpha}
\end{minipage}
\end{figure*}
\begin{figure*}[h]
\begin{minipage}{\linewidth}
    \centering
    \begin{minipage}{0.33\linewidth}
        \centerline{$\gamma\sigma(\alpha) = 0.001, \beta = 0.01$}
        \includegraphics[width=\linewidth]{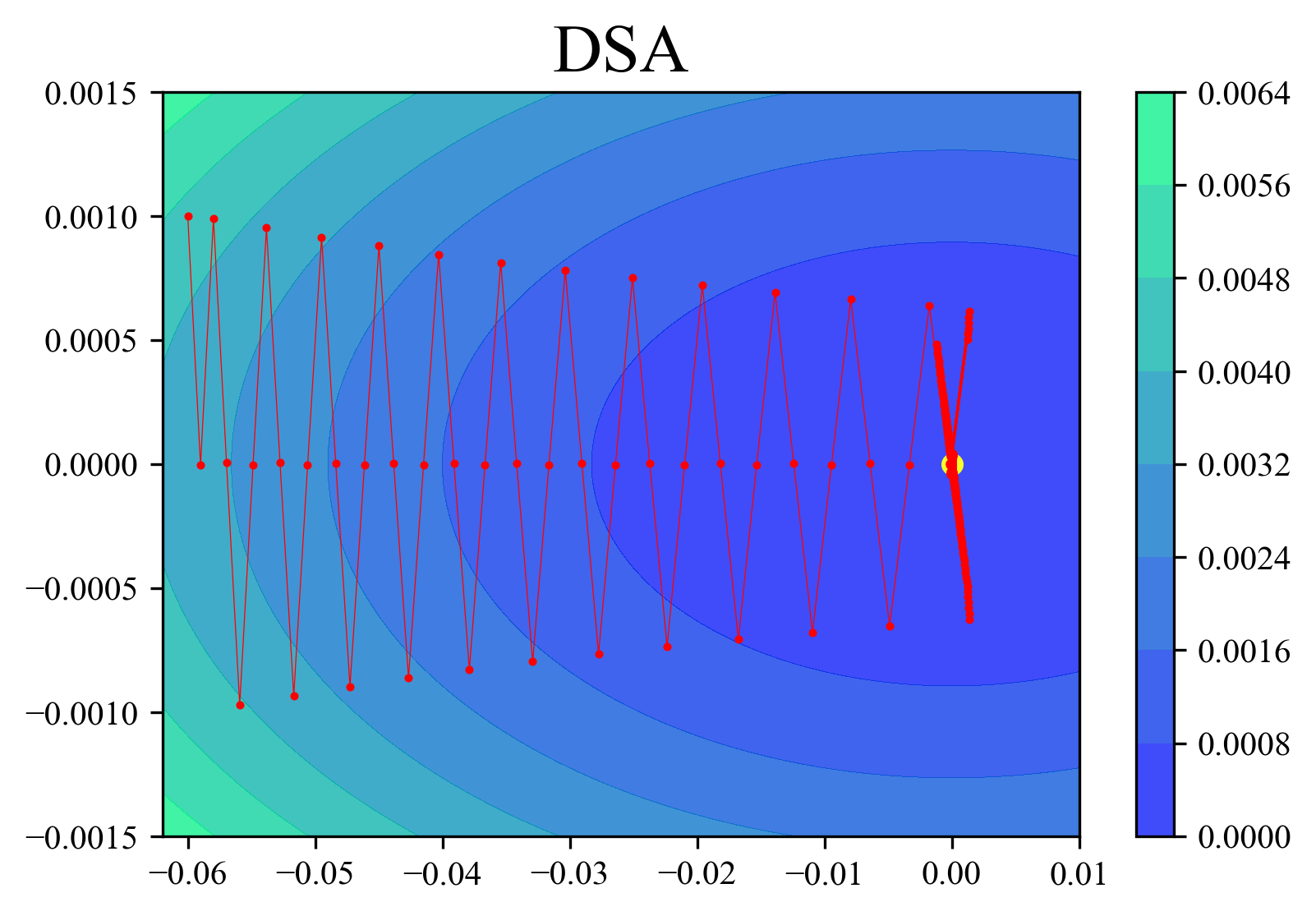}
    \end{minipage}
    \begin{minipage}{0.33\linewidth}
        \centerline{$\gamma\sigma(\alpha) = 0.001, \beta = 0.1$}
        \includegraphics[width=\linewidth]{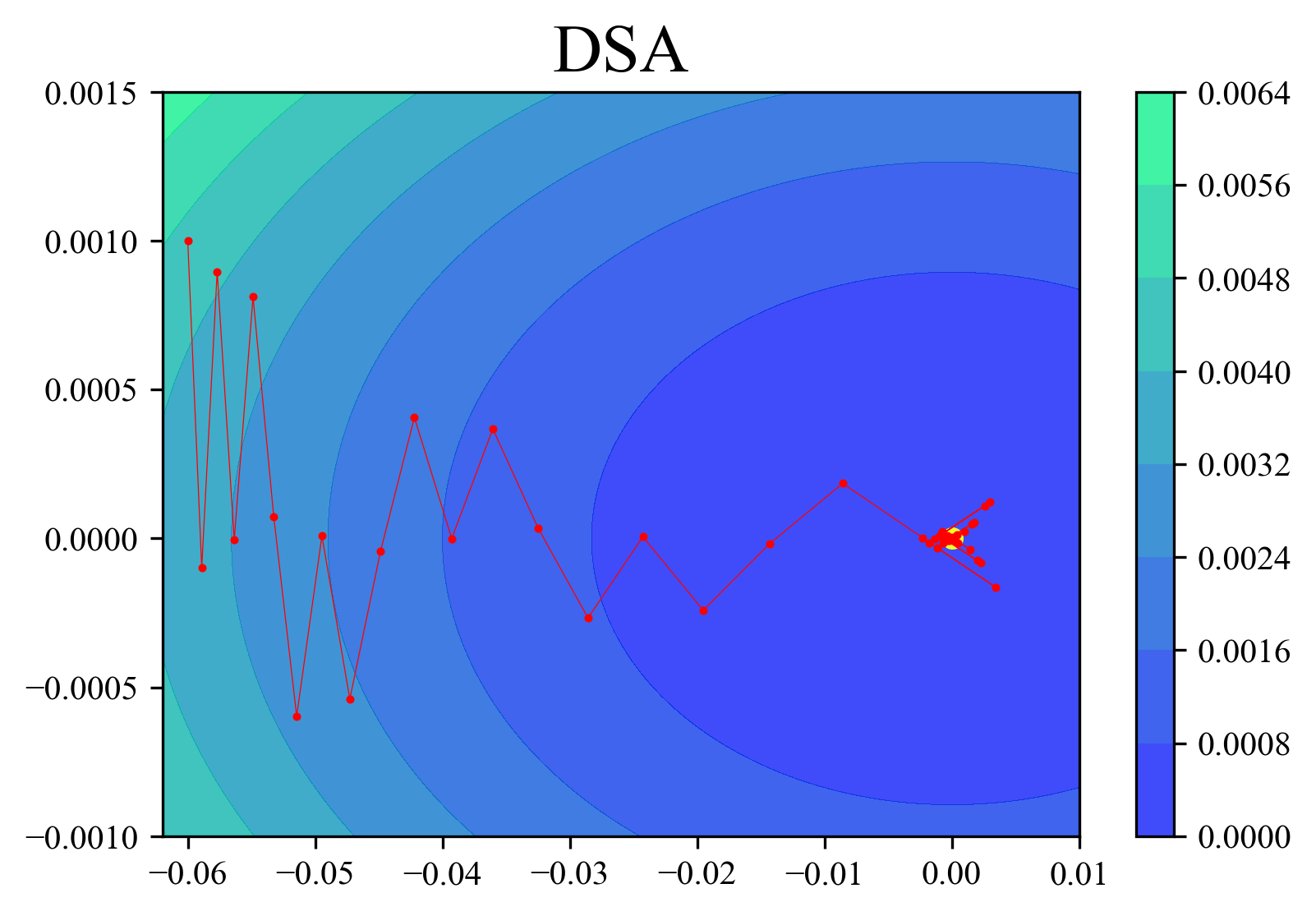}
    \end{minipage}
    \begin{minipage}{0.33\linewidth}
        \centerline{$\gamma\sigma(\alpha) = 0.001, \beta = 1$}
        \includegraphics[width=\linewidth]{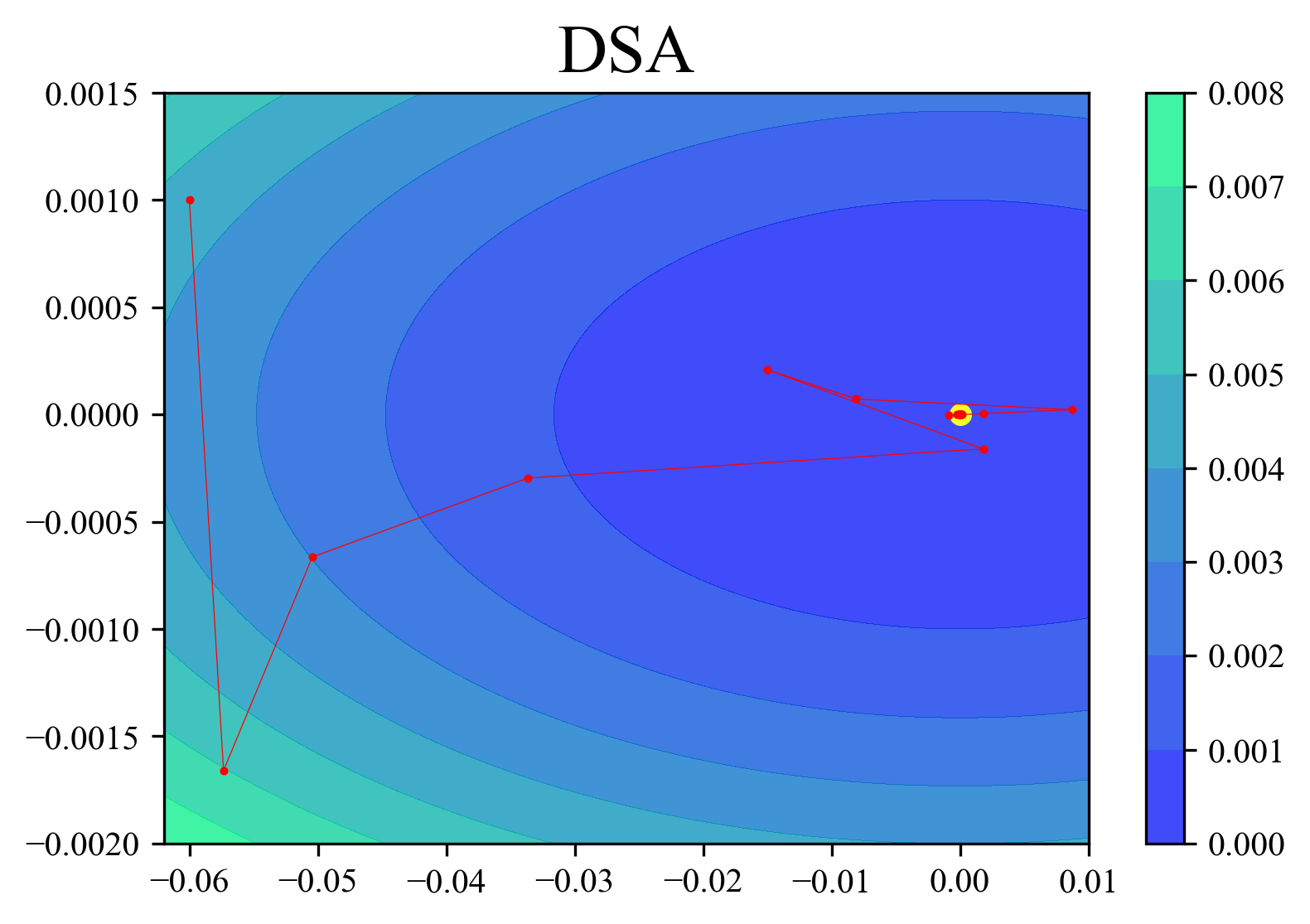}
    \end{minipage}
    \caption{Sensitivity of $\beta$}
    \label{fig: sens_beta}
\end{minipage}
\end{figure*}
\\\textbf{Sensitivity Analysis}\\
Now we take sensitivity analysis for DSA using the convex optimization case, where $(a, b) = (1, 1000)$ and $(w_1, w_2)$ starts from $(-0.06, 0.001)$. Two key hyper-parameters of DSA are initial learning rate which is essentially $\alpha$, and step size of $\alpha$, i.e., $\beta$. Firstly, $\beta$ is fixed as 0.3 and initial learning rate is observed on various scales including $(0.01, 0.001, 0.0001)$. The second group fixed initial learning rate as 0.001 and tuned $\beta$ from $(1, 0.1, 0.001)$.
Performance with different $\alpha$ and $\beta$ are demonstrated in Fig.~\ref{fig: sens_alpha} and Fig.~\ref{fig: sens_beta}. When the initial learning rate increases from 0.0001 to 0.01, DSA tend to approach the optimal with fewer iterations. However, it will need more steps to decrease learning rate around the optimal. DSA also approaches the optimal faster when $\beta$ increases from 0.01 to 1, because learning rate will adapt more sensitively. But it also takes more iterations to decrease learning rate in the end, because learning rate is adapted very large in the beginning. Totally, a sensitive and stable enough configuration on this task is $\gamma\sigma(\alpha) = 0.001, \beta = 0.3$.

\section{Conclusion $\And$ Future work}
\label{sec: conclusion}
In this paper, we propose the optimizer with sensitive self-adaptive learning rate for fast and high-qualified convergence.
Compared with existing optimizers, DSA has stronger adaptive capabilities and is competent to a variety of machine learning tasks. While this requires a reasonable initial learning rate and step size. In addition, in the later stage of training, the learning rate is still in an active state, i.e., pointless adaptation. Therefore, how to determine a reasonable initial value or eliminate the negative influence of the initial value and make the learning rate of DSA converge stably will be the main issue to be studied next.

\section{Related work}
\label{sec: related}
Learning rate is sure to be the most significant hyper-parameter for a machine learning model, which is the core of an optimizer. Current optimizers can be classified to three categories, that is troditional SGD~\cite{gradientdescent}, self-adaptive optimizer such as Adam~\cite{adam}, differentiable learning rate based optimizer known as hypergardient descent~\cite{hypergradient}.

Gradient descent~\cite{gradientdescent} is nearly the earliest optimization algorithm, the three primary types including stochastic gradient descent~(SGD), mini-batch type and batch type. Model parameters are updated according to their first-order gardient of the cost function, where backpropagation is applied to the calculation of gradients. However, this kind of naive optimization algorithm performs without any adaptation mechanisms for learnint rate, which causes a great dependence to gradient's magnitude and a trade-off between speed and accuracy.

According to gradient descent's fakes, one method is to adopt adaptive updating methods for learning rate. Those proposed methods include Momentum~\cite{momentum}, AdaGrad~\cite{adagrad}, RMSProp~\cite{rmsprop}, Adam~\cite{adam} and so on.
Gradient descent with momentum~\cite{momentum} accumulates an exponentially decaying moving average of past gradients and continues to move in their direction~\cite{deeplearning}. Momentum has the effect of dampening down the change in the gradient~\cite{montavon2012neural}, which makes it more stable than troditional gardient descent.
AdaGrad~\cite{adagrad} adaptively scaled the learning rate for each dimension with the sum of the squares of the gradients as denominator. That is, learning rate will keep decreasing in a training session.
RMSProp introduces second-order momentum to solve the problem that AdaGrad is greatly affected by historical gradients.
Root Mean Squared Propagation, or RMSProp~\cite{rmsprop}, is an extension of GD and AdaGrad that uses a decaying average of partial gradients in the adaptation of the step size for each parameter. RMSProp extends Adagrad to avoid the effect of a monotonically decreasing learning rate~\cite{kochenderfer2019algorithms}.
AdaDelta~\cite{adadelta} is an adaptive optimizer without learning rate, which is analogous to Newton's method to find a more accurate step size for each update.
Adam~\cite{adam} combines the best properties of the AdaGrad and RMSProp algorithms to provide an optimization algorithm that can handle sparse gradients on noisy problems. Variants of Adam gain new features, for example, AdamW~\cite{adamw} adds a regular term, Adamax~\cite{adam} extends the two norm to the infinite norm to obtain more stable and concise results, SparseAdam~\cite{adam} is designed to deal with sparse tensors.

Learning rate scheduler is a trick in the training which has similar effects with adaptive optimizer. For example, we usually set a milestone for learning rate in the half and the three quarters of the training session and in each milestone, the learning rate will be scaled down. Evidence shows SGD always does better than those adaptive optimizers with the help of scheduler.

However, scheduler is still a manual work. Hypergardient descent~(HD)~\cite{hypergradient} is the algorithm that best matches our understanding of adaptation, cause HD is the first to utilize the gradient of learning rate to the cost function to update learning rate. That is, HD makes learning rate differentiable. Similar related works like multi-level HD~\cite{jie2020adaptive} and so on are still on the way.


%

\ifCLASSOPTIONcompsoc
  \section*{Acknowledgments}
\else
  \section*{Acknowledgment}
\fi

This paper was supported by NSFC grant (U1866602 71773025). The National Key Research and Development Program of China (2020YFB1006104).

\ifCLASSOPTIONcaptionsoff
  \newpage
\fi



\bibliographystyle{IEEEtran}
\bibliography{IEEEabrv,ref.bib}

\begin{thebibliography}{10}
\providecommand{\url}[1]{#1}
\csname url@samestyle\endcsname
\providecommand{\newblock}{\relax}
\providecommand{\bibinfo}[2]{#2}
\providecommand{\BIBentrySTDinterwordspacing}{\spaceskip=0pt\relax}
\providecommand{\BIBentryALTinterwordstretchfactor}{4}
\providecommand{\BIBentryALTinterwordspacing}{\spaceskip=\fontdimen2\font plus
\BIBentryALTinterwordstretchfactor\fontdimen3\font minus
  \fontdimen4\font\relax}
\providecommand{\BIBforeignlanguage}[2]{{%
\expandafter\ifx\csname l@#1\endcsname\relax
\typeout{** WARNING: IEEEtran.bst: No hyphenation pattern has been}%
\typeout{** loaded for the language `#1'. Using the pattern for}%
\typeout{** the default language instead.}%
\else
\language=\csname l@#1\endcsname
\fi
#2}}
\providecommand{\BIBdecl}{\relax}
\BIBdecl

\bibitem{gradientdescent}
C.~Lemar{\'e}chal, ``Cauchy and the gradient method,'' \emph{Doc Math Extra},
  vol. 251, no. 254, p.~10, 2012.

\bibitem{bottou2012stochastic}
L.~Bottou, ``Stochastic gradient descent tricks,'' in \emph{Neural networks:
  Tricks of the trade}.\hskip 1em plus 0.5em minus 0.4em\relax Springer, 2012,
  pp. 421--436.

\bibitem{intability}
\BIBentryALTinterwordspacing
S.~Ruder, ``An overview of gradient descent optimization algorithms,''
  \emph{CoRR}, vol. abs/1609.04747, 2016. [Online]. Available:
  \url{http://arxiv.org/abs/1609.04747}
\BIBentrySTDinterwordspacing

\bibitem{momentum}
I.~Sutskever, J.~Martens, G.~Dahl, and G.~Hinton, ``On the importance of
  initialization and momentum in deep learning,'' in \emph{International
  conference on machine learning}.\hskip 1em plus 0.5em minus 0.4em\relax PMLR,
  2013, pp. 1139--1147.

\bibitem{adagrad}
J.~Duchi, E.~Hazan, and Y.~Singer, ``Adaptive subgradient methods for online
  learning and stochastic optimization.'' \emph{Journal of machine learning
  research}, vol.~12, no.~7, 2011.

\bibitem{rmsprop}
A.~Graves, ``Generating sequences with recurrent neural networks,'' \emph{arXiv
  preprint arXiv:1308.0850}, 2013.

\bibitem{adam}
D.~P. Kingma and J.~Ba, ``Adam: A method for stochastic optimization,'' in
  \emph{ICLR (Poster)}, 2015.

\bibitem{hypergradient}
A.~G. Baydin, R.~Cornish, D.~M. Rubio, M.~Schmidt, and F.~Wood, ``Online
  learning rate adaptation with hypergradient descent,'' in \emph{International
  Conference on Learning Representations}, 2018.

\bibitem{resnet}
K.~He, X.~Zhang, S.~Ren, and J.~Sun, ``Deep residual learning for image
  recognition,'' in \emph{Proceedings of the IEEE conference on computer vision
  and pattern recognition}, 2016, pp. 770--778.

\bibitem{fmp}
B.~Graham, ``Fractional max-pooling,'' \emph{arXiv preprint arXiv:1412.6071},
  2014.

\bibitem{dnn1}
Kunihiko and Fukushima, ``Neocognitron: A self-organizing neural network model
  for a mechanism of pattern recognition unaffected by shift in position,''
  \emph{Biological Cybernetics}, 1980.

\bibitem{dnn2}
Y.~Lecun and L.~Bottou, ``Gradient-based learning applied to document
  recognition,'' \emph{Proceedings of the IEEE}, vol.~86, no.~11, pp.
  2278--2324, 1998.

\bibitem{mlp}
M.~W. Gardner and S.~Dorling, ``Artificial neural networks (the multilayer
  perceptron)—a review of applications in the atmospheric sciences,''
  \emph{Atmospheric environment}, vol.~32, no. 14-15, pp. 2627--2636, 1998.

\bibitem{mnist}
Y.~LeCun and C.~Cortes, ``{MNIST} handwritten digit database,'' 2010.

\bibitem{svhn}
``Reading digits in natural images with unsupervised feature learning,''
  \emph{nips workshop on deep learning \& unsupervised feature learning}, 2011.

\bibitem{cifar}
A.~Krizhevsky, G.~Hinton \emph{et~al.}, ``Learning multiple layers of features
  from tiny images,'' 2009.

\bibitem{adadelta}
M.~D. Zeiler, ``Adadelta: an adaptive learning rate method,'' \emph{arXiv
  preprint arXiv:1212.5701}, 2012.

\bibitem{adamw}
I.~Loshchilov and F.~Hutter, ``Decoupled weight decay regularization,''
  \emph{arXiv preprint arXiv:1711.05101}, 2017.

\bibitem{scikit-learn}
F.~Pedregosa, G.~Varoquaux, A.~Gramfort, V.~Michel, B.~Thirion, O.~Grisel,
  M.~Blondel, P.~Prettenhofer, R.~Weiss, V.~Dubourg, J.~Vanderplas, A.~Passos,
  D.~Cournapeau, M.~Brucher, M.~Perrot, and E.~Duchesnay, ``Scikit-learn:
  Machine learning in {P}ython,'' \emph{Journal of Machine Learning Research},
  vol.~12, pp. 2825--2830, 2011.

\bibitem{3DSemanticSegmentationWithSubmanifoldSparseConvNet}
B.~Graham, M.~Engelcke, and L.~van~der Maaten, ``3d semantic segmentation with
  submanifold sparse convolutional networks,'' \emph{CVPR}, 2018.

\bibitem{SubmanifoldSparseConvNet}
B.~Graham and L.~van~der Maaten, ``Submanifold sparse convolutional networks,''
  \emph{arXiv preprint arXiv:1706.01307}, 2017.

\bibitem{dnn3}
Behnke and Sven, ``Hierarchical neural networks for image interpretation,''
  \emph{Springer,}, 2003.

\bibitem{prelu}
K.~He, X.~Zhang, S.~Ren, and J.~Sun, ``Delving deep into rectifiers: Surpassing
  human-level performance on imagenet classification,'' in \emph{Proceedings of
  the IEEE international conference on computer vision}, 2015, pp. 1026--1034.

\bibitem{NEURIPS2019_9015}
A.~Paszke, S.~Gross, F.~Massa, A.~Lerer, J.~Bradbury, G.~Chanan, T.~Killeen,
  Z.~Lin, N.~Gimelshein, L.~Antiga, A.~Desmaison, A.~Kopf, E.~Yang, Z.~DeVito,
  M.~Raison, A.~Tejani, S.~Chilamkurthy, B.~Steiner, L.~Fang, J.~Bai, and
  S.~Chintala, ``Pytorch: An imperative style, high-performance deep learning
  library,'' in \emph{Advances in Neural Information Processing Systems 32},
  H.~Wallach, H.~Larochelle, A.~Beygelzimer, F.~d\textquotesingle
  Alch\'{e}-Buc, E.~Fox, and R.~Garnett, Eds.\hskip 1em plus 0.5em minus
  0.4em\relax Curran Associates, Inc., 2019, pp. 8024--8035.

\bibitem{kaiming}
K.~He, X.~Zhang, S.~Ren, and J.~Sun, ``Delving deep into rectifiers: Surpassing
  human-level performance on imagenet classification,'' in \emph{Proceedings of
  the IEEE international conference on computer vision}, 2015, pp. 1026--1034.

\bibitem{deeplearning}
Y.~LeCun, Y.~Bengio, and G.~Hinton, ``Deep learning,'' \emph{nature}, vol. 521,
  no. 7553, pp. 436--444, 2015.

\bibitem{montavon2012neural}
G.~Montavon, G.~Orr, and K.-R. M{\"u}ller, \emph{Neural networks: tricks of the
  trade}.\hskip 1em plus 0.5em minus 0.4em\relax springer, 2012, vol. 7700.

\bibitem{kochenderfer2019algorithms}
M.~J. Kochenderfer and T.~A. Wheeler, \emph{Algorithms for optimization}.\hskip
  1em plus 0.5em minus 0.4em\relax Mit Press, 2019.

\bibitem{jie2020adaptive}
R.~Jie, J.~Gao, A.~Vasnev, and M.-N. Tran, ``Adaptive multi-level
  hyper-gradient descent,'' \emph{arXiv preprint arXiv:2008.07277}, 2020.

\end{thebibliography}
%



%
\begin{IEEEbiography}[{\includegraphics[width=1in,height=1.25in,clip,keepaspectratio]{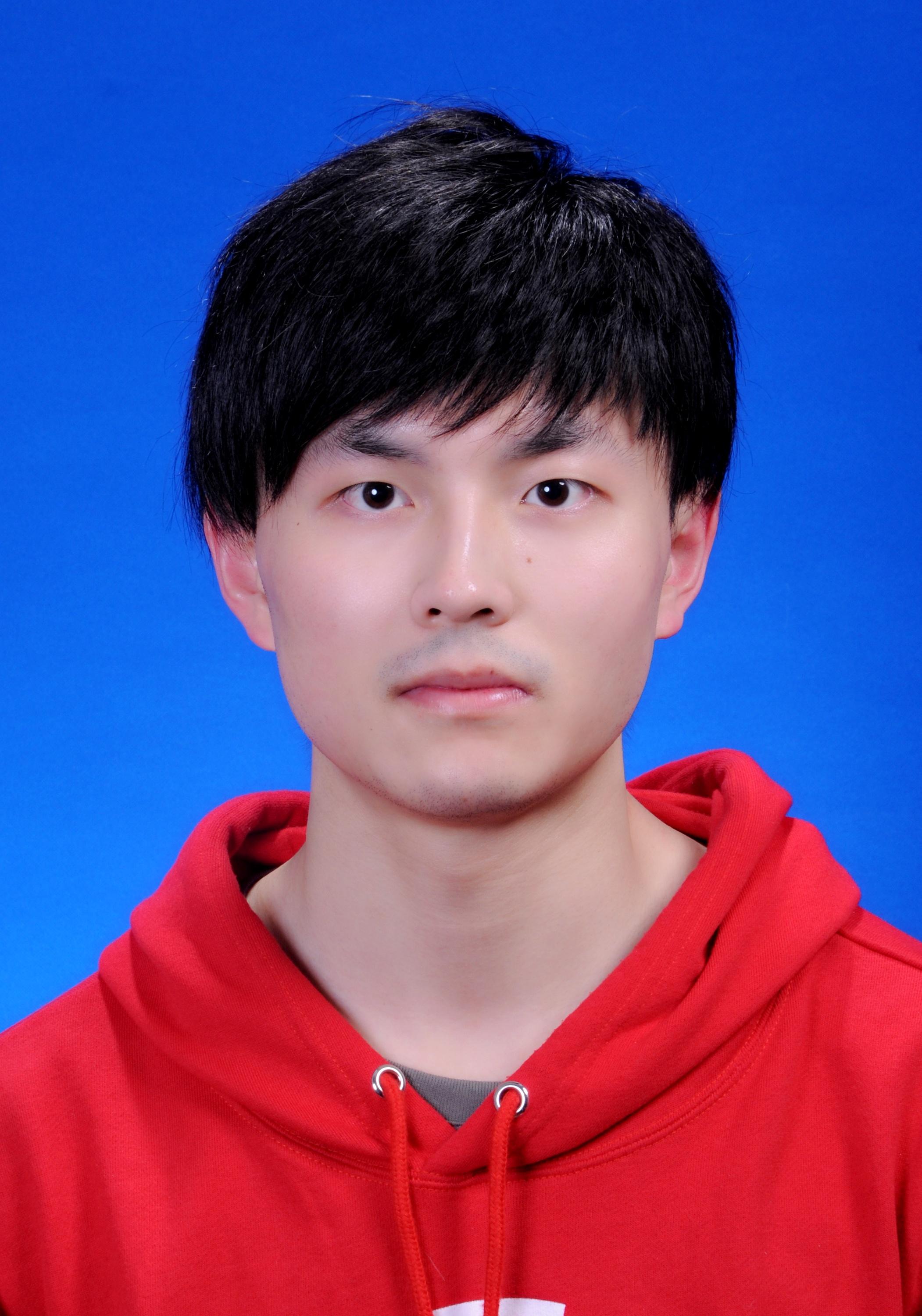}}]{Bozhou Chen}
received the BS degree from the Harbin Institute of Technology, China. He is currently studying for a master's degree of computer science at Harbin Institute of Technology, Harbin Institute of Technology, China. His research interests include knowledge completion, NAS and some other machine learning problems.
\end{IEEEbiography}
\begin{IEEEbiography}[{\includegraphics[width=1in,height=1.25in,clip,keepaspectratio]{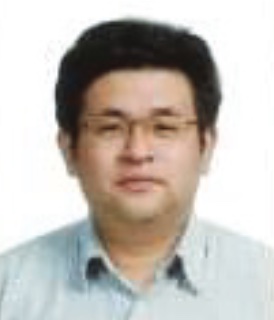}}]{Hongzhi Wang}
is a Professor and doctoral supervisor at Harbin Institute of Technology, ACM member. His research area is data management, includingdata quality and graph management. He is a recipient of the outstandingdissertation award of CCF and Microsoft Fellow.
\end{IEEEbiography}
\begin{IEEEbiography}[{\includegraphics[width=1in,height=1.25in,clip,keepaspectratio]{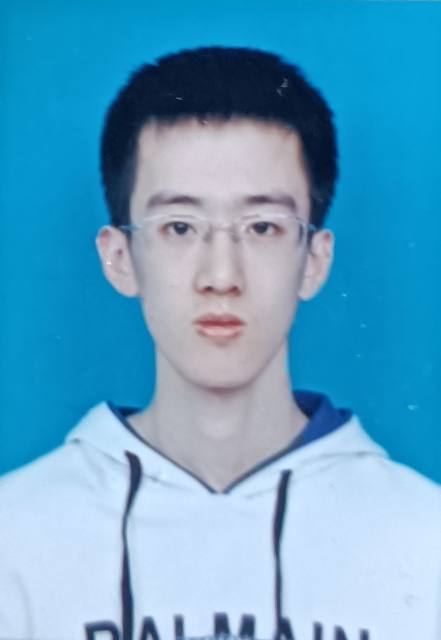}}]{Chenmin Ba}
received the bachelor's degree from Harbin Institute of technology, China, in 2021. He is currently studying for a master's degree of computer science at  Harbin Institute of Technology. His research interests include software vulnerability identification and software vulnerability location.
\end{IEEEbiography}








\end{document}